\documentclass[preprint,12pt,authoryear]{elsarticle}
\usepackage{graphicx,lmodern,amsfonts,mathrsfs,amsmath,amssymb,latexsym,xspace,epsfig,psfrag,color,tikz,tikzscale,pgfplots,colortbl}
\usepackage[colorlinks,bookmarksnumbered,citecolor=blue]{hyperref}
\usepackage[ruled]{algorithm2e}
\usepackage[utf8]{inputenc}  
\usepackage{scalefnt}
\usepackage{multirow}
\usepackage{pdflscape}
\usepackage{multicol}
\usepackage{natbib}
\usepackage{hyperref}
\usepackage{lineno}
\usepackage{longtable, lscape} 
\usepackage{bbm}
\usepackage{calrsfs}
\usepackage{txfonts}
\usepackage{rotating}
\usepackage{comment}

\usepackage[top=30mm,bottom=20mm,left=30mm,right=20mm,pdftex,includeheadfoot]{geometry}
\usepackage{graphicx}
\usepackage{subfig}
\usepackage{rotating}
\usepackage{hyperref}
\usepackage{tikz,pgfplots}
\pgfplotsset{compat=1.15}
\usepgfplotslibrary{statistics}
\usetikzlibrary{shapes,arrows}
\usetikzlibrary{arrows.meta}
\tikzset{%
  >={Latex[width=2mm,length=2mm]},
            base/.style = {rectangle, rounded corners, draw=black,
                           minimum width=2cm, minimum height=1cm,
                           text centered, font=\small},
  Start/.style = {base, fill=green!30},
       question/.style = {base, fill=red!30},
    end/.style = {base, fill=blue!30},
         process/.style = {base, minimum width=2.5cm, fill=orange!10,
                           font=\small},
}

\definecolor{Gray}{gray}{0.95}

\journal{Expert Systems with Applications}

\begin{document}

\begin{frontmatter}

\title{Meteorological and human mobility data on predicting COVID-19 cases by a novel hybrid decomposition method with anomaly detection analysis: a case study in the capitals of Brazil}

\author{Tiago Tiburcio da Silva\corref{cor1}}
\ead{ttsilva@gmail.com}
\cortext[cor1]{Corresponding author}

\author{Rodrigo Francisquini}
\ead{rodrigo@francisquini.com}

\author{Mariá C. V. Nascimento}
\ead{mcv.nascimento@unifesp.br}

\address{Instituto de Ciência e Tecnologia, Universidade Federal de São Paulo (UNIFESP)\\ Av. Cesare M. G. Lattes, 1201, Eugênio de Mello, São José dos Campos-SP, CEP: 12247-014, Brasil
}

\begin{abstract}
In 2020, Brazil was the leading country in  COVID-19 cases in Latin America, and capital cities were the most severely affected by the outbreak. Climates vary in Brazil due to the territorial extension of the country, its relief, geography, and other factors. Since the most common COVID-19 symptoms are related to the respiratory system, many researchers have studied the correlation between the number of COVID-19 cases with meteorological variables like temperature, humidity, rainfall, etc. Also, due to its high transmission rate, some researchers have analyzed the impact of human mobility on the dynamics of COVID-19 transmission. There is a dearth of literature that considers these two variables when predicting the spread of  COVID-19 cases.  In this paper, we analyzed the correlation between the number of COVID-19 cases and human mobility, and meteorological data in Brazilian capitals. We found that the correlation between such variables depends on the regions where the cities are located. We employed the variables with a significant correlation with   COVID-19 cases to predict the number of COVID-19 infections in all Brazilian capitals and proposed a prediction method combining the Ensemble Empirical Mode Decomposition (EEMD)  method with the Autoregressive Integrated Moving Average Exogenous inputs (ARIMAX) method, which we called EEMD-ARIMAX. After analyzing the results poor predictions were further investigated using a signal processing-based anomaly detection method. Computational tests showed that EEMD-ARIMAX achieved a forecast  26.73\%  better than ARIMAX. Moreover, an improvement of 30.69\% in the average root mean squared error (RMSE) was noticed when applying the EEMD-ARIMAX method to the data normalized after the anomaly detection. \end{abstract}


\end{frontmatter}

\section{Introduction}

According to the Centers for Disease Control and Prevention, a pandemic ``refers to an increase, often sudden, in the number of cases of a disease above what is normally expected'' ``over several countries or continents, usually affecting a large number of people'' \citep{Dicker2006}. Several pandemic outbreaks have befallen humanity over the centuries. One of the first recorded pandemics occurred between 165 A.D. and 180 A.D. in the reign of Marcus Aurelius, when Antonine Plague wiped out a third of the population in some areas of the Roman empire and decimated the Roman army \citep{Ligon2006}.

 Almost 500 years later, during the mid-sixth century, the Justinian plague hit the Byzantine empire. During this epidemic, 40\% of Constantinople's population was wiped out. One of the greatest pandemics in human history, the Black Death, or Bubonic plague, occurred between 1347 and 1352, and killed between 75 and 200 million people. Several other pandemics have occurred, such as New World Smallpox (1520-unknown), The Third plague (1855), and The 1918 Flu (1918-1920). 
 
In 2002, Severe Acute Respiratory Syndrome (SARS), caused by SARS Coronavirus (SARS-CoV), emerged in the province of Guangdong, southern  China, infecting thousands of people and causing the death of approximately one thousand humans \citep{Zhong2003}. \cite{Cheng2007} stated that ``the presence of a large reservoir of SARS-CoV viruses in horseshoe bats, together with the culture of eating exotic mammals in southern China'' was ``a time bomb''. The authors warned about the possibility of a resurgence of SARS-Cov and other new viruses in animals or laboratories and that everyone should be prepared for a new pandemic. Eight years later, a new coronavirus variant was discovered in the Middle East, the  Middle East Respiratory Syndrome Coronavirus (MERS-CoV), which is still a reality. Four years after MERS-CoV, another coronavirus emerged in Wuhan, China (December 2019). Because of its similarity to SARS-CoV, this new coronavirus was called SARS-CoV-2, and the disease, COVID-19 (Coronavirus Disease 2019) \citep{Huang2020}.

COVID-19 is a highly contagious virus that spread rapidly around the world, causing  worldwide travel restrictions as well as mandatory lockdown in many cities. On April 29, 2021, the World Health Organization (WHO) reported that the virus was in 223 Countries, with 148,999,876 confirmed cases, and 3,140,115 deaths \citep{whocovid}. In Brazil, until April 29, 2021, there had been confirmed 14,441,563 COVID-19 cases and 395,022 deaths caused by the virus \citep{who}. For this reason, scientists around the world from the most diverse areas have  focused their studies on understanding COVID-19 transmission dynamics \citep{Fang2020}, prevention \citep{Ali2020,Li2020,vaccine}, detection \citep{ISMAEL2021,VIDAL2021}, control measures \citep{Meo2020}, and prediction analysis \citep{HERNANDEZMATAMOROS2020,KATRIS2021,PETROPOULOS2020}.

Historically, viral respiratory tract infections,  such as the ones caused by the coronaviruses from past epidemics, H1N1 influenza and syncytial virus, were related to meteorological factors which possibly influenced the transmission and stability of the virus \citep{Baker2019,Barreca2012,Chanetal2011,Lowen2014,Paynter2015}. Several authors studied the correlation between climatic variables and the number of COVID-19 cases in the world:  absolute humidity and temperature in the USA \citep{GUPTA2020}; UV index, wind speed, absolute humidity, among others, in 206 countries/regions \citep{ISLAM2020}; average air humidity and temperature in Brazil \citep{Neto2020}; temperature, absolute humidity, dew point, among others, in Singapore \citep{PANI2020}; and wind speed and temperature in Turkey \citep{SAHIN2020}, for example. However, only a few authors addressed the prediction of COVID-19 cases using models that consider climatic variables, as can be observed in \cite{DASILVA2020,MAKADE2020,Mousavi2020}.


Some studies also investigate the impact of human mobility on COVID-19 transmission. In these cases,  mobility can be measured by passenger traffic in airports \citep{OZTIG2020}, for example, or by changes in commuting patterns \citep{Badr2020,SHAO2021,Wang2020,ZHU2020}. All these studies show that there is a strong correlation between human mobility and the number of people infected by COVID-19.

To our knowledge, no study has yet investigated the impact of both human mobility and meteorological variables on COVID-19 transmission rates. Both these factors should both be considered in such studies as there is clear evidence that climate affects human mobility \citep{BRUMBASTOS2018}. This statement likely holds since, on warmer days, people lean toward performing outdoor activities and attending open-air events; and on colder days, the opposite holds is true.

The daily number of COVID-19 cases can be modeled and studied through the time series theory. In a general way, a time series can be thought of as a combination of other time series, each explaining the original data at different frequencies \citep{Buyuksahin2018}. In this way, the frequency range of each subdivision is formed and creates more linear structures within them, making the prediction of this original time series more accurate. Several techniques can be used to obtain the decomposition, such as Principal Component Analysis (PCA) \citep{Jolliffe2002}, Variational Mode Decomposition (VMD) \citep{Dragomiretskiy2014}, Fourier Transform (FT) \citep{Graps1995}, Empirical Mode Decomposition (EMD) \citep{Huang1998}, Ensemble Empirical Mode Decomposition (EEMD) \citep{HuangWu2008}, and Singular Spectrum Analysis (SSA) \citep{Golyandina2001}. Since PCA is limited to linear time series; FT is limited to linear, periodic, or stationary time series \citep{Huang1998}; SSA is an application of PCA  in the time domain \citep{Hsieh2002}; the VMD application has to solve a variational optimization problem  which requires predetermining an appropriate number of variational modes; and since EMD presents a mode-mixing problem; EEMD has been considered one of the most useful tools to decompose time series, either because of its simplicity or because it is not limited to linear nor stationary time series. 

According to \cite{Dong2019}, EMD‐based methods, like EEMD, substantially enhance prediction accuracy and have been successfully used in several types of datasets, such as IoT systems \citep{Yu2019}, bitcoin \citep{khaldi2018}, geology \citep{LIU2019}, economy \citep{WU2019}, finance \citep{LIN2021}, medicine \citep{Guangda2019,ZHA2018}, machine fault diagnosis \citep{AMIRAT2018}, and water resource management\citep{NIU2019}. It has also been applied for meteorological data, such as temperature \citep{LIU2019}, precipitation \citep{Alizadeh2019}, and wind speed \citep{SANTHOSH2019}. In this paper, we propose an adaptation of the EEMD method to decompose several time series, and to use these new decomposed time series in the forecast of another time series also decomposed by the same adaptation. The time series that will be predicted corresponds to the number of daily cases of COVID-19, and the other series, used as independent variables, correspond to the meteorological and human mobility time series. Only the time series that showed a reasonable correlation with the daily cases of COVID-19, in each city, are considered in the prediction. The prediction method employed in this paper is the Autoregressive Integrated Moving Average Exogenous inputs (ARIMAX) method  \citep{BoxJenkins1990}, a well-established method previously mentioned in the literature.




In this paper, we also aim at understanding how, together, meteorological conditions and human mobility  affect the transmission of COVID-19.   For such, we analyze both meteorological variables (rainfall, maximum temperature, minimum temperature, and humidity) and human mobility variables (movement trends over time by geography, across different categories of places: retail and recreation areas, grocery stores and pharmacies, parks, transit stations, workplaces, and residential areas).
The main contributions of this paper can be summarized as follows:

\begin{itemize}
    \item It provides a thorough analysis on the correlation of meteorological  and human mobility variables in Brazilian capitals;
    \item It uses meteorological variables and human mobility in the prediction of daily cases of COVID-19 in Brazilian capitals;
    \item It adapts EEMD to decompose time series with independent variables;
    \item It proposes a novel method that combines the introduced EEMD-based method  with ARIMAX to predict time series with independent variables, called EEMD-ARIMAX; 
    \item It develops an oriented-case anomaly detection algorithm to better investigate the significant errors in prediction and thus adjust the prediction;
    \item It improves the ARIMAX forecast by 26.73\% using the new EEMD-ARIMAX method;
    \item It refines the method by using the introduced  anomaly detection strategy, thus improving the prediction by 30.69\%.
\end{itemize}

The rest of the paper is organized as follows. Section~\ref{sec:related} presents a general literature review on the prediction of COVID-19 cases using human mobility and meteorological data. Moreover, it also shows a brief discussion on prediction methods, giving special attention to decomposition-based methods introduced to predict COVID-19 cases.  Section~\ref{sec:data} shows the main features of the data used in the case study and introduces the proposed EEMD-ARIMAX method. Section~\ref{sec:results} presents the results obtained by the proposed strategy EEMD-ARIMAX after a thorough correlation data analysis is carried out. Section~\ref{sec:anomaly} shows the performed  data anomaly detection and the results of the EEMD-ARIMAX and ARIMAX  in normalized data. Section~\ref{sec:final} wraps up the paper drawing some conclusions and giving directions for future works. A list of symbols referring to all the notations used throughout the paper is presented in \ref{sec:appendix}.

\section{Related Work}
\label{sec:related}

This section presents a brief literature review on predicting COVID-19 cases considering either meteorological or mobility variables. It also presents a short overview of methods for COVID-19 prediction, in particular, methods more closely related to the performed study. 

\subsection{Human mobility in the prediction of COVID-19 cases}
According to \cite{Nayak2021}, one of the primary impacts on predicting the COVID-19 cases consists of the variations in engagement, i.e. how committed people are to taking measures to reduce the number of COVID-19 cases. These measures include washing hands, wearing face masks, and maintaining social distancing. Concerning social distancing, one way to estimate the level of commitment is by analyzing the rates of human mobility. In line with this, \cite{OZTIG2020} considered the flow of people at airports as a human mobility measure, and observed that the greater the number of airports in a country, the more likely it is for the country to have a higher number of COVID-19 cases. This conclusion was drawn by the use of negative binomial regression analysis.
 
 \cite{Badr2020}  studied the correlation between social distancing and COVID-19 cases, where social distancing was quantified by mobility patterns. To model the mobility data, the authors considered changes in commuting patterns between and within counties in the USA. The data to model the mobility patterns were obtained by \href{https://www.teralytics.net/}{Teralytics} (Zürich, Switzerland). \cite{Wang2020} coupled the data of confirmed COVID-19 cases with the Google mobility data in Australia. The authors concluded that the social restriction policies imposed in the country at the emergence of the first COVID-19 case were effective in curbing the spread of the virus. Moreover, they observed that the correlation between human mobility and the spread of COVID-19 varies according to the type of mobility.

\cite{SHAO2021} used human mobility data from 47 countries in 6 continents collected from Mobility Trends Reports (from  Apple Inc.), and showed that human mobility is strongly related to the COVID-19 transmission rate. \cite{ZHU2020} demonstrated a positive link between human mobility and the number of people infected by COVID-19, considering data from 120 cities in China. These studies show a clear influence of human mobility on the spread of COVID-19. However, cities present different human mobility patterns depending on factors such as how technological the cities are, the conditions of public and private transportation systems, among others. Therefore, the relationship/correlation between human mobility and dissemination of COVID-19 must be evaluated considering the cities' particularities. Bearing this in mind, in this study we focus on the relationship between human mobility and the spread of COVID-19 cases in each of the 27 Brazilian capitals.

\subsection{Meteorological variables in the prediction of COVID-19 cases}
\cite{SAHIN2020} and \cite{Sharma2021} state that meteorological features should be used to improve the accuracy of  COVID-19 predictions. Such variables are crucial factors affecting infectious diseases, whether in terms of changes in the transmission dynamics, regarding host susceptibility, or the survival of the virus in the environment \citep{McClymont2021}. In line with this, \cite{GUPTA2020} studied the relationship among new COVID-19 cases, absolute humidity, and temperatures in the USA. The authors observed that the spread of COVID-19 was majorly influenced by the absolute humidity in a narrow range of 4 to 6 g/m$^3$.

\cite{ISLAM2020} investigated the link between some environmental factors and COVID-19 cases in  206 countries/regions (until April 20, 2020). The relationship between the spread of COVID-19 and humidity, and UV index were inconclusive. Their investigation suggested a negative relationship between wind speed and COVID-19 cases. Moreover, a higher rate of COVID-19 cases was observed in environments with an absolute humidity between 5 and 10 g/m$^3$. In Singapore, \cite{PANI2020} revealed that temperature, absolute humidity, and dew point have a positive correlation with the number of daily COVID-19 cases. Wind speed, atmospheric boundary layer height, and ventilation coefficient, on the other hand, showed a negative correlation with the number of COVID-19 cases. In Turkey,  \cite{SAHIN2020} showed that wind speed has a positive correlation with COVID-19 cases, and that temperature and COVID-19 cases are negatively correlated.

In Brazil, \cite{Neto2020} concluded that only the average air humidity was significantly correlated with the number of COVID-19 cases (considering data from Brazilian capitals, and data available from April 2020 to May 2020). The study revealed a positive correlation, in contrast with the results obtained by others studies performed in cities in China, Spain, and the United States. The authors also demonstrated that population density presented a strong positive correlation with the number of COVID-19 cases in the Brazilian capitals. They emphasize that population density, which is linked with higher human mobility, and poorer social-economic environments that have deficient sanitary conditions contribute to the spread of the virus.

Although some studies have addressed the studies involving the relationship between meteorological variables and COVID-19, some results appear to be inconsistent. On the one hand, temperature and humidity, for example, were reported as having a significant impact in the majority of the studies. On the other hand, the correlation was positive in some cases and negative in others. These observations suggest that the link between meteorological features and the number of COVID-19 cases is complex and hard to generalize. There is evidence that meteorological variables contribute to the increase in the transmission of COVID-19, but the effect of these relationships should be studied locally, since other factors such as human mobility and public health measures (lockdown, for example) also have a strong influence on the number of COVID-19  cases.

\subsection{COVID-19 and forecasting}

From a methodological point of view, several studies attempt to understand the spread of COVID-19 using artificial intelligence. \cite{Albahri2020} provided an exhaustive overview of integrated artificial intelligence based on data mining and machine learning algorithms. The authors pointed to a need for integrated sensor technologies for outdoor scenarios to control the spread of the coronavirus. This process is only possible when there is an interconnection with IoT technologies. \cite{Nayak2021} present an overview of the applicability of intelligent systems such as machine learning and deep learning to solve COVID-19 outbreak-related issues. \cite{Sharma2021} reported and summarized the research performed on COVID-19 with machine learning and big data.

The literature presents few studies that address the problem of predicting new COVID-19 cases through decomposition methods. To our knowledge, only \cite{DASILVA2020} and \cite{Mousavi2020} used decomposition methods in their predictions considering independent variables. Both proposed strategies were based on the variational mode decomposition (VMD). \cite{DASILVA2020} used VMD and some prediction techniques including deep learning and machine learning, to predict COVID-19 cumulative confirmed cases in five Brazilian states and five American states with high daily incidences. The authors used temperature and precipitation as exogenous variables. They pointed that the VMD coupled with cubist regression achieved the best results among the tested techniques. \cite{Mousavi2020} proposed a model based on the combination of VMD with Long Short Term Memory considering the daily temperature, humidity, and transmission rates in the prediction of new COVID-19 daily cases in Maharashtra, Tamil Nadu, and Gujarat, India. Among these works, only \cite{Mousavi2020} addressed the prediction of daily COVID-19 cases, since such prediction is more difficult because of the accumulated cases. In this study, we address the prediction of new daily cases of COVID-19.

\section{Case study: predictive analysis of Brazilian data}
\label{sec:data}

In 2020,  Brazil had an estimated population of 212,622,578 inhabitants \citep{ibge}. Brazil was the country with the greatest number of COVID-19 cases in 2020 in Latin America, ranking third in the world. Capitals are the most affected cities, and some experience health system collapse, such as Manaus-AM \citep{ferrante2020brazil}.  Since 23.86\% of the Brazilian population lives in capital cities, the spatial units of analysis in this study were the 27 capitals in Brazil \citep{ibge}.

Brazil is a country with continental dimensions and the 5th largest country in the world in territorial extension occupying an area of  8,510,295.91 km$^2$. The Brazilian climate has great variations, with 3 climate zones and 12 climate types \citep{alvares2013}. We want to analyze the correlation among COVID-19 cases with meteorological and human mobility parameters, and if there are differences in these correlations within the same country. 


\subsection{Data}

COVID-19 data were obtained from Brasil.io \citep{dadoscovid}, which compiles newsletters from the State Health Secretariats of Brazil. Meteorological data were obtained from the \textit{Centro de Previsão de Tempo e Estudos Climáticos} located at the \textit{Instituto Nacional de Pesquisas Espaciais} \citep{cptec}. The meteorological data considered in this study are:

\begin{itemize}
    \item Minimum Temperature (Min Temp): refers to the daily minimum temperature in degrees Celsius;
    \item Maximum Temperature (Max Temp):  refers to the daily maximum temperature in degrees Celsius;
    \item Humidity (Hum): refers to the daily air humidity in percentage;
    \item Rainfall (Rain): refers to the daily total precipitation in millimeters.
\end{itemize}

Human mobility data were obtained from the COVID-19 Community Mobility Reports \citep{google} prepared by Google. These reports point to geographical movement trends over time, across different categories of places. The place categories are:

\begin{itemize}
    \item Retail and recreation (RR): refers to mobility trends to places like restaurants, shopping centers, theme parks, etc;
    \item Grocery and pharmacy (GP): refers to mobility trends to places like grocery markets, farmers markets, pharmacies, etc;
    \item Parks (PA): refers to mobility trends to places like local parks, public beaches, public gardens, etc;
    \item Transit stations (TS): refers to mobility trends to places like subway, bus, train stations, etc;
    \item Workplaces (WO): refers to mobility to places of work;
    \item Residential (RE): refers to mobility to places of residence.
\end{itemize}

The Residential category shows a change in the permanence of people in their homes, while the other categories measure changes in the total number of visitors. Changes in mobility patterns each day were compared with a baseline corresponding to the same day of the week. This baseline corresponds to the median of the corresponding day of the week, during the five weeks from January 3 to February 6, 2020.

The number of observations in human mobility data and meteorological data varies according to the number of data in the variable relative to daily COVID-19 cases in each city. In each city, all reported data start at the day they confirmed the first COVID-19 case in the city (column ``first case'' in Table \ref{tab:average-weather}  of the \ref{appendix-data}) and end at the final compiled day: November 6, 2020.  A more descriptive analysis of the data is presented in \ref{appendix-data}.


\subsection{Ensemble Empirical Mode Decomposition}
\label{sec:eemd}

Time series decomposition techniques have the goal of extracting simple periodic signals from the original time series, which can be used as inputs to machine learning approaches or other statistical models. Our study focuses on the use of the Ensemble Empirical Mode Decomposition (EEMD) technique \citep{HuangWu2008}, an adaptive data analysis method based on local characteristics of the data. EEMD catches nonlinear, non-stationary oscillations effectively. EEMD has been  successfully used in several types of datasets \citep{LIN2021,NIU2019}, mainly in meteorological data, such as temperature \citep{LIU2019}, precipitation \citep{Alizadeh2019}, and wind speed \citep{SANTHOSH2019}.

EEMD is an improvement of the  empirical mode decomposition (EMD) method \citep{Huang1998,HuangWu2008}.  It aims at  decomposing the original data into a series of modes, called finite intrinsic mode functions (IMFs) and a residual, identifying the oscillatory modes that coexist. EEMD overcomes the so-called mode-mixing problem found in  EMD. The mode-mixing occurs when different oscillation components coexist in a single IMF and very similar oscillations reside in different IMFs \citep{HuangWu2008}.

 EEMD uses an ensemble of IMFs obtained by applying EMD to several different series of the original time series obtained by adding white Gaussian noise. Adding a white Gaussian noise reduces the mode-mixing problem by occupying the whole time-frequency space \citep{HuangWu2008}. In summary, EEMD has the following steps. 

\begin{enumerate}
    \item Let $W_t$, $m$ and $s$ be the input data corresponding to, respectively,  the original time series that will be decomposed, the number of ensembles, and the number  of IMFs to be extracted from $W_t$.
    \item Make $k=1$, a control variable that indicates the ensemble to be generated in the iteration.
    \item Generate a new time series $Z_t$, obtained from $W_t$ for the ensemble $k$, adding to it a white noise with a standard deviation $\sigma_ {noise}$ proportional to the standard deviation  of $W_t$, called $\sigma_{original}$.  Therefore, $\sigma_ {noise} = \mu \sigma_{original}$, where $\mu$ is a relatively small number which must be empirically determined.  \label{generate}
    Make $j=1$,  a control variable related to the index of the IMF of the $k$-th ensemble to be defined in the following steps, referred to as IMF$^k_j$.
    \item     Identify all the local extreme values of $Z_t$ -- a combination of high and low values of the series. After that, interpolate all this values by a cubic spline interpolation as the upper (high values) and lower envelopes (low values), respectively $e_{max}^k$ and $e_{min}^k$. \label{identify1}
    \item Calculate the point-to-point  arithmetic mean between the envelopes  -- $m_t^k = (e_{min}^k + e_{max}^k)/2$ -- and subtract this ``average time series'' from  time series $Z_t$, obtaining the  time series $d_t^k$ -- $d_t^k = Z_t
    - m_t^k$. \label{identify2}
    \item If $j \leq s$, then IMF$^k_j=d_t^k$, $j=j+1$, $Z_t=Z_t-d_t^k$, and repeat steps~\ref{identify1} and \ref{identify2}. If $j>s$, assign $Z_t$ to the residual time series, called Res$_{W_t}$. 
    \label{end}
    \item Make $k=k+1$ and repeat  Steps~\ref{generate} to \ref{end}  until $k>m$, i.e. until the method obtains the $m$ ensembles.
\end{enumerate}

\begin{figure} 
\centering
 \subfloat{ 
    \includegraphics[trim = 20mm 19mm 5mm 5mm,scale=0.6]{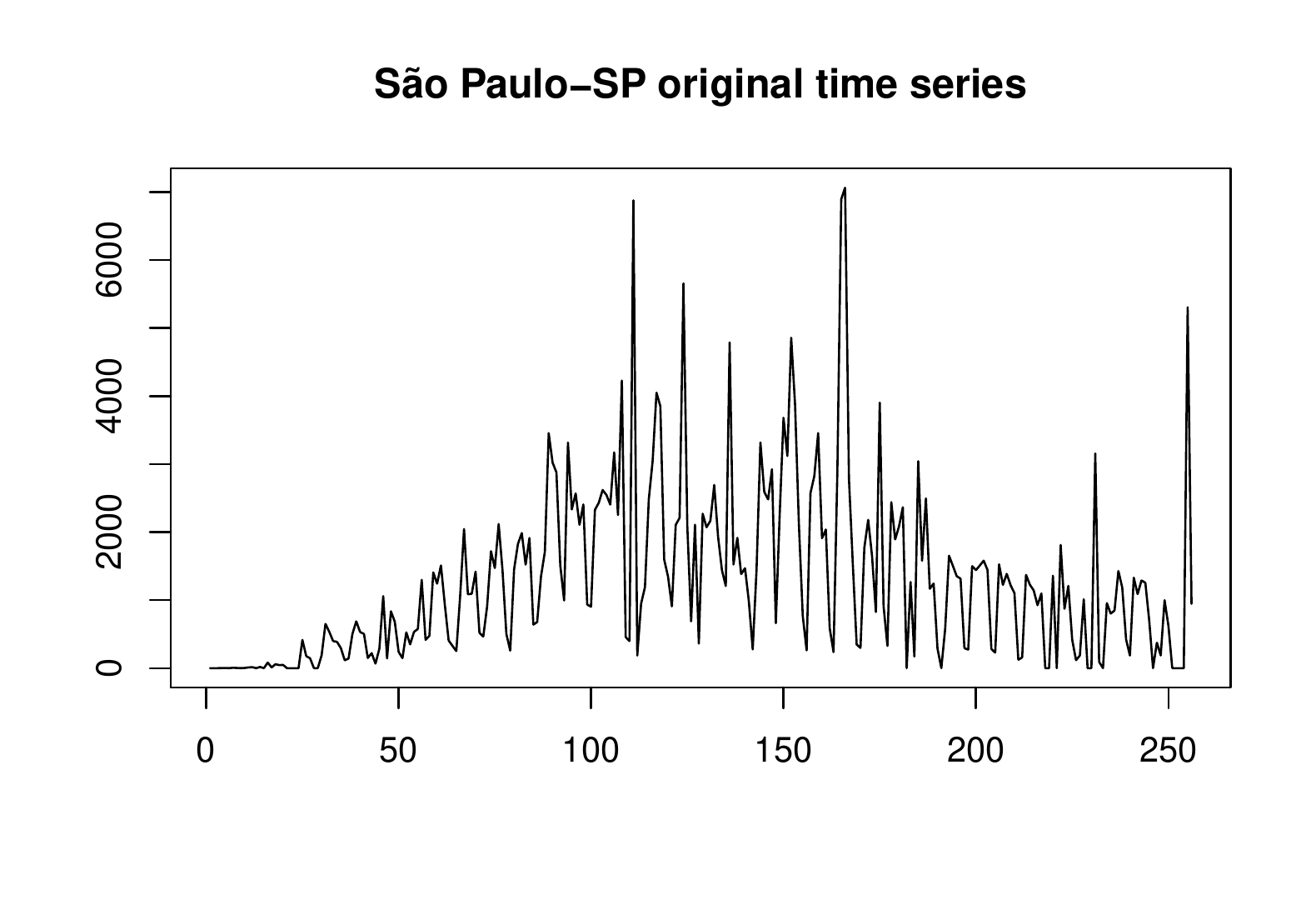}
  } 
  \subfloat{ 
    \includegraphics[trim = 20mm 19mm 7mm 5mm,scale=0.6]{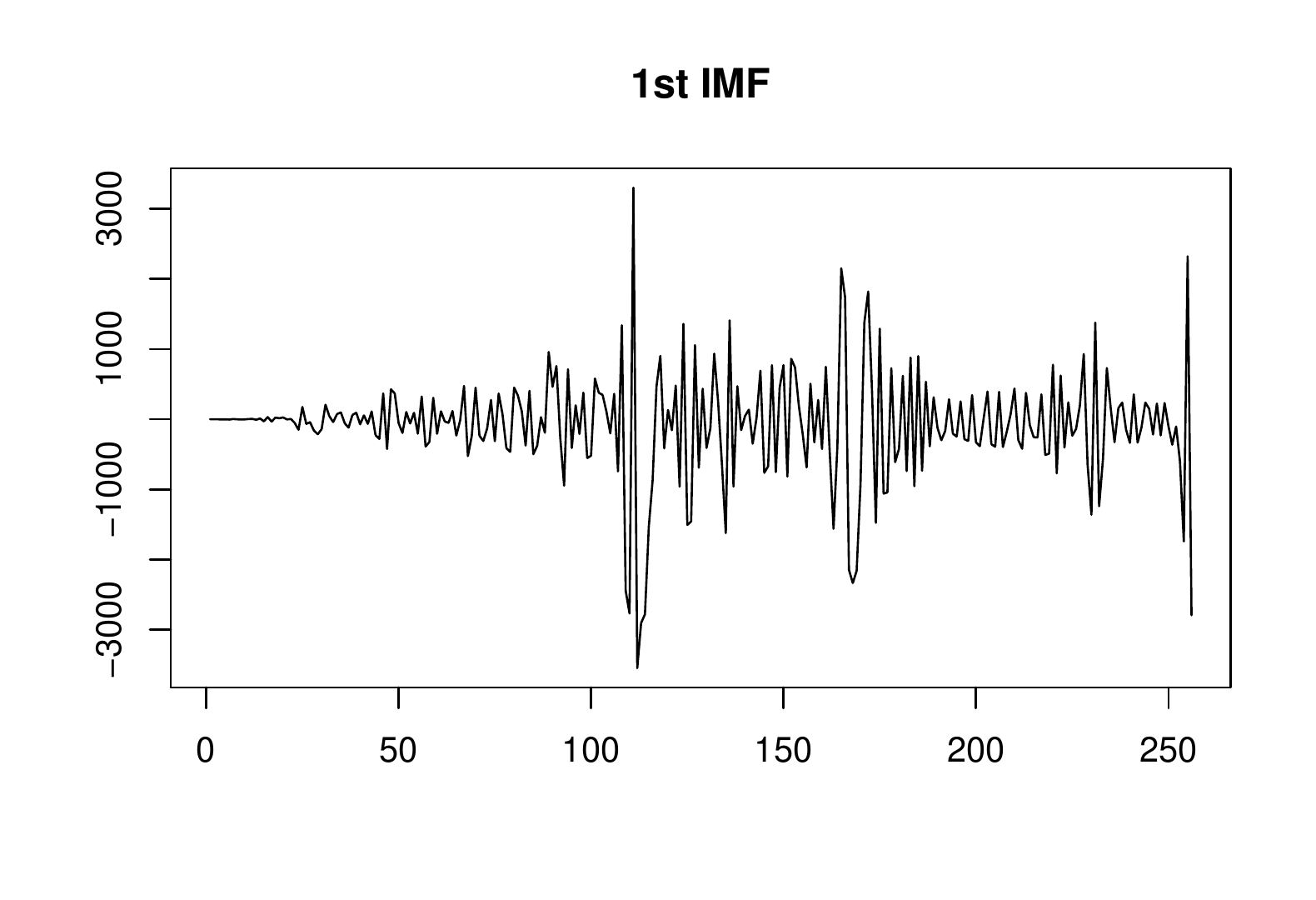}
  } 
  \\ 
  \subfloat{ 
     \includegraphics[trim = 20mm 19mm 5mm 5mm,scale=0.6]{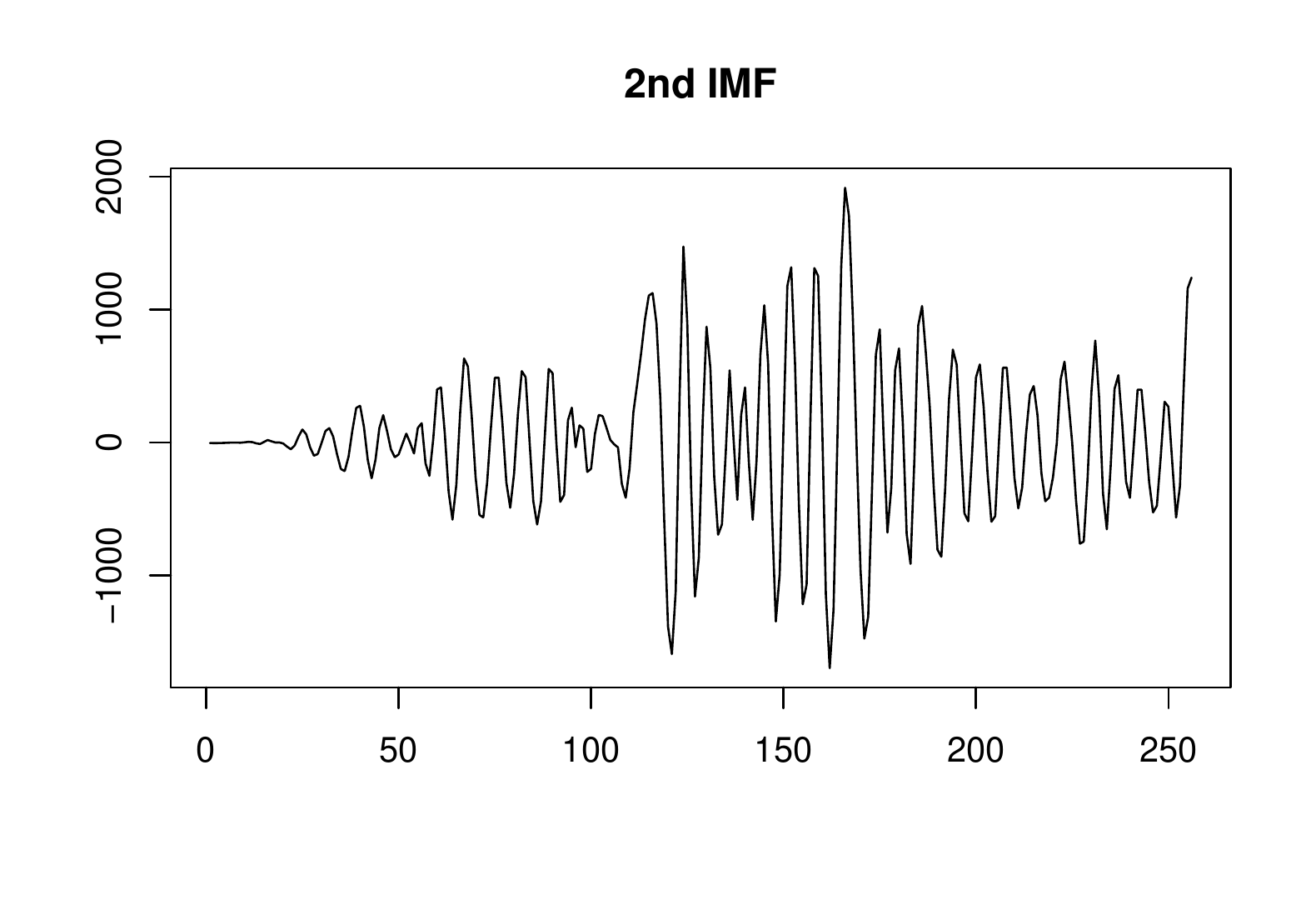}
  } 
    \subfloat{ 
    \includegraphics[trim = 20mm 19mm 7mm 5mm,scale=0.6]{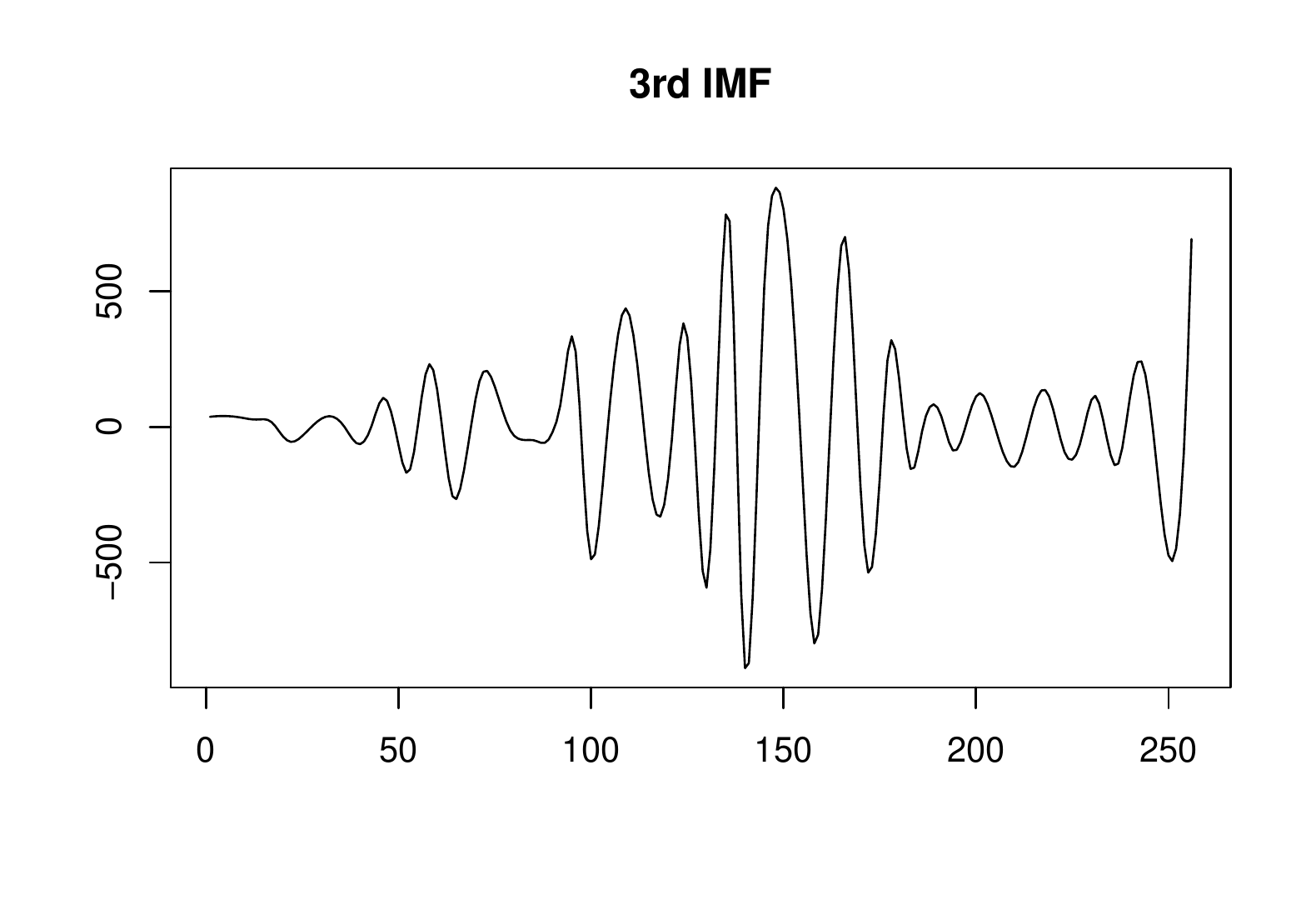}
  }
    \\ 
  \subfloat{ 
     \includegraphics[trim = 20mm 19mm 5mm 5mm,scale=0.6]{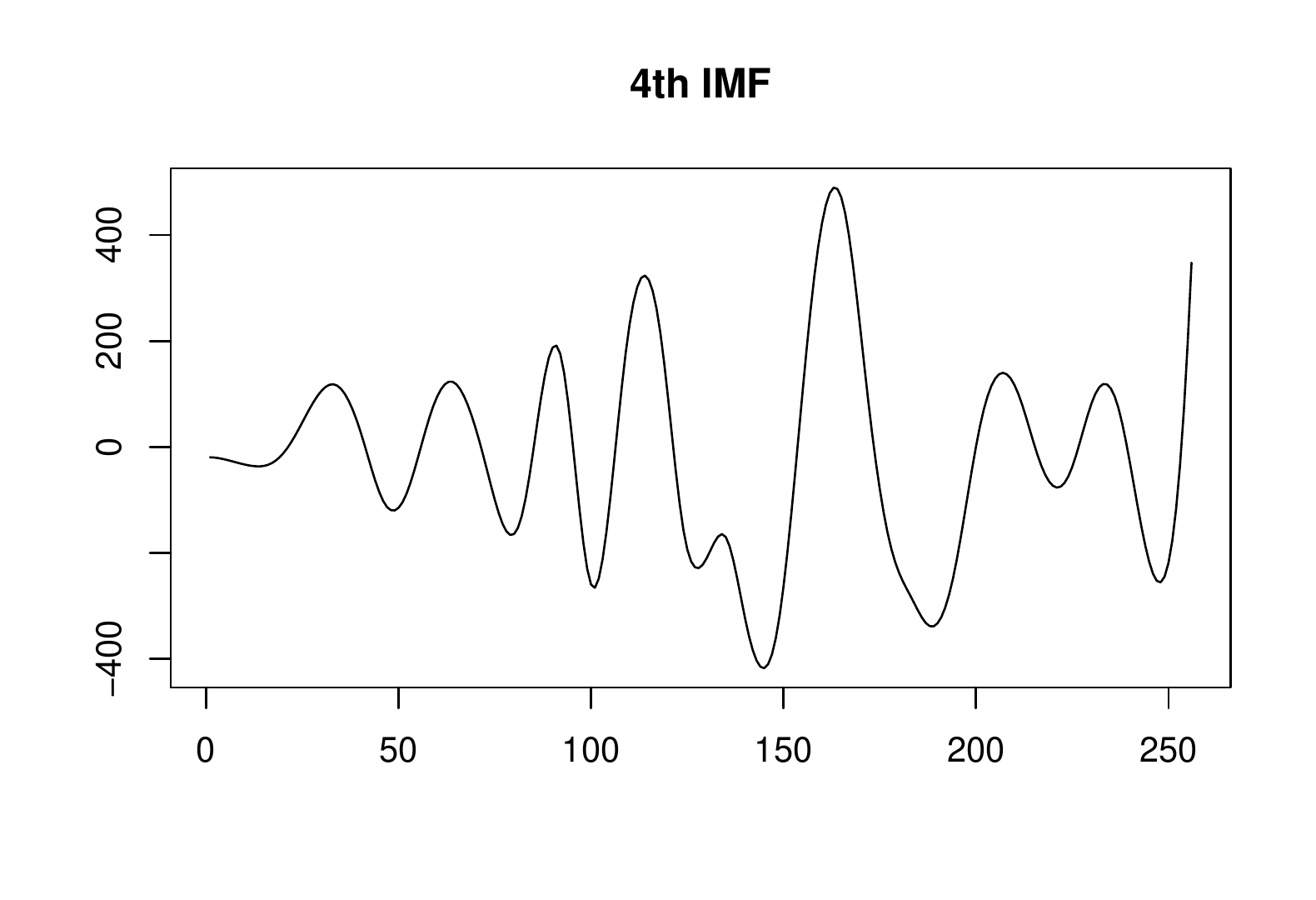}
  } 
    \subfloat{ 
    \includegraphics[trim = 20mm 19mm 7mm 5mm,scale=0.6]{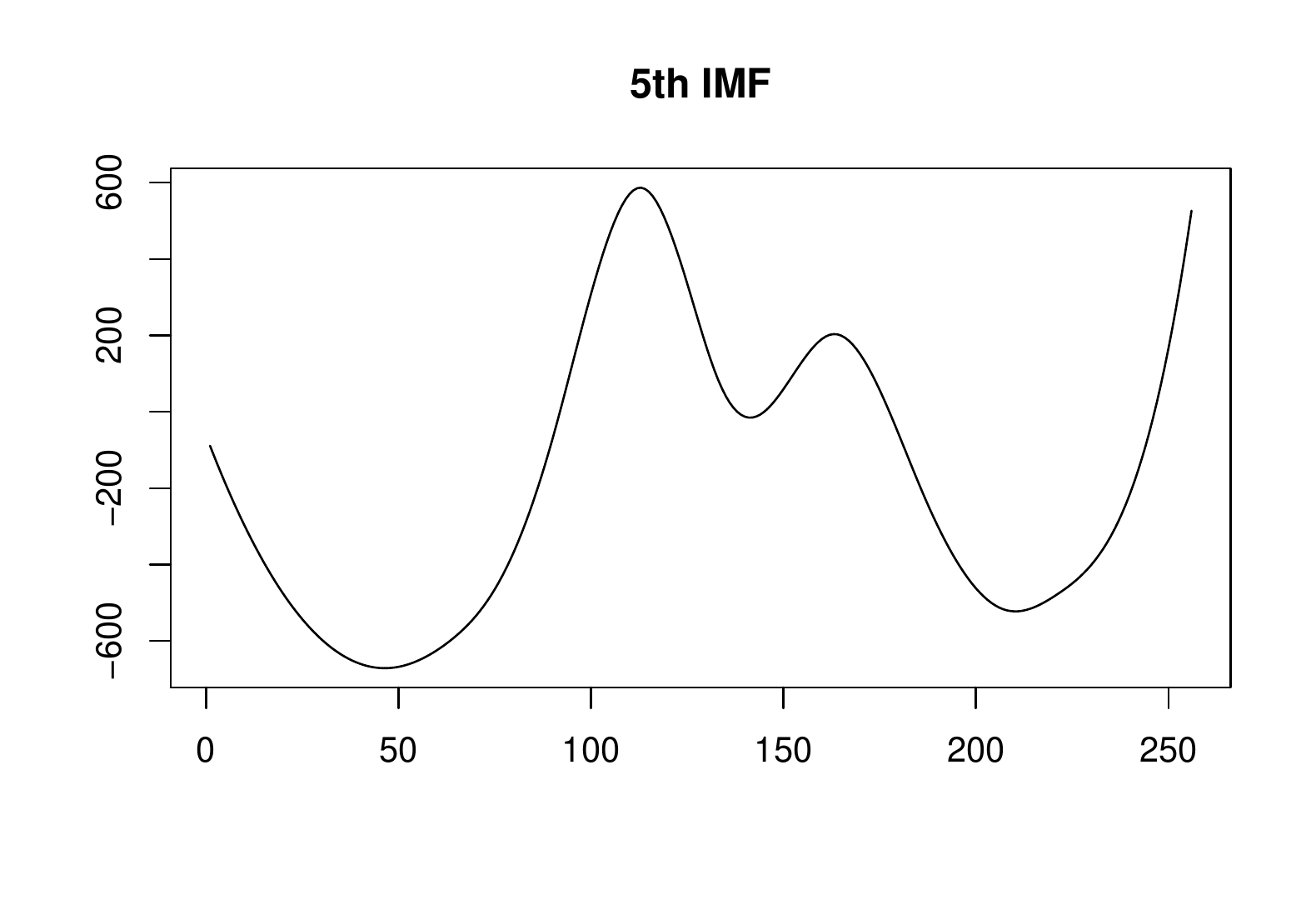}
  }
  \\
  \subfloat{
  \includegraphics[trim = 20mm 19mm 0mm 5mm,scale=0.6]{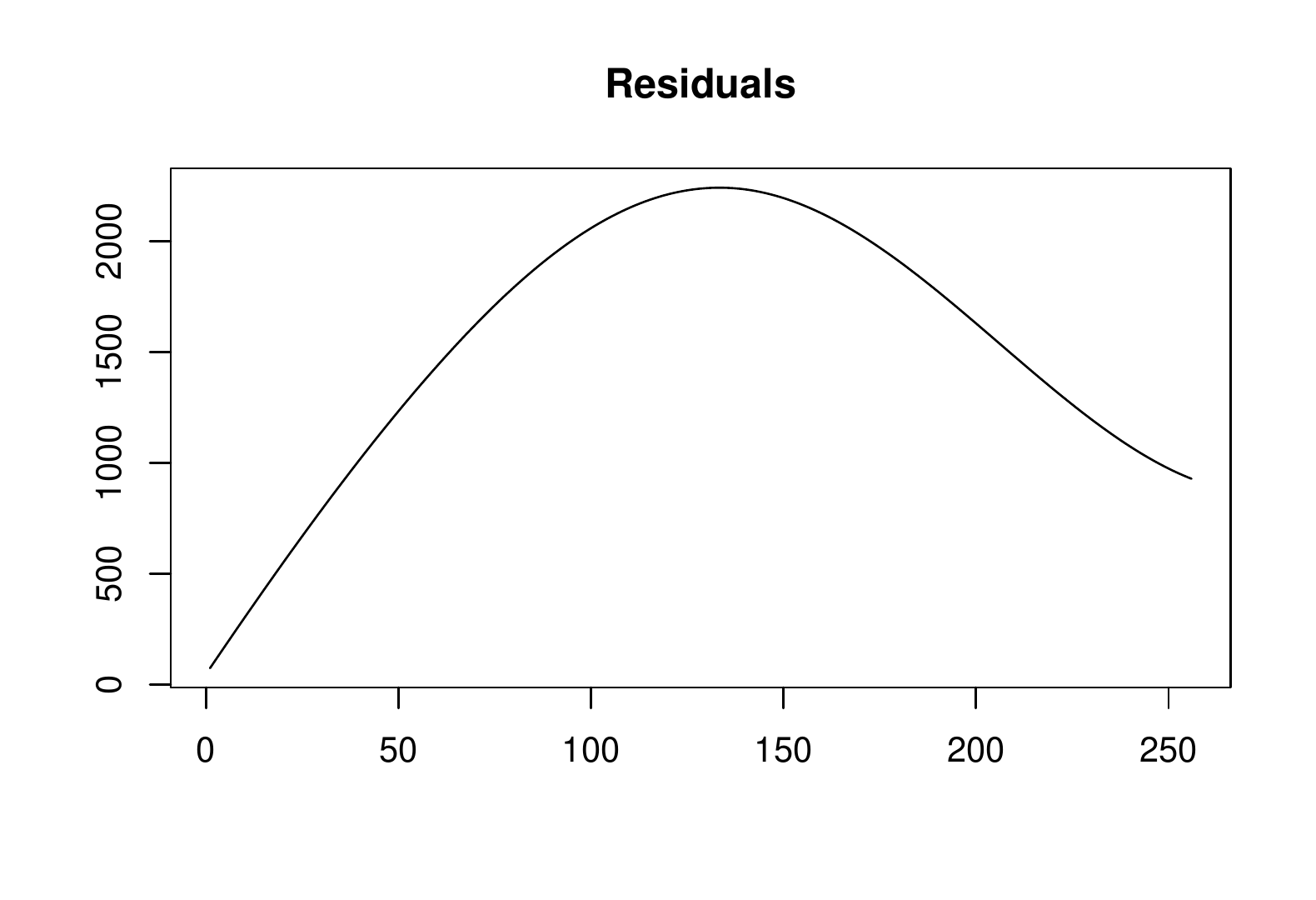}
  }
  \caption{Decomposed IMFs and residual obtained by EEMD considering the number of COVID-19 cases in São Paulo.}
    \label{fig:saopaulo-eemd}
\end{figure}

The values of $\mu$ and $m$ were empirically chosen after several computational tests. These tests indicated that an ensemble number $m = 125$ and the $\mu$ value equals $0.01$ presented better outcomes. Furthermore, because of the proposed EEMD-ARIMAX method in Section \ref{sec:eemd-arima}, the number of IMFs into which the time series is decomposed was fixed in advance and, after tests, we found that a decomposition into 5 IMFs plus a residual was the most appropriate, i.e., $s=5$. 
 Figure \ref{fig:saopaulo-eemd} shows the IMFs extracted from the data of São Paulo COVID-19 cases, by applying the EEMD algorithm. The IMFs were plotted from the first to the last component  extracted from the series, where the last plot corresponds to the residual. The x-axis indicates the days, whereas  the y-axis represents the values of the decomposed time series. 


\subsection{Autoregressive Integrated Moving Average Exogenous inputs (ARIMAX)}
\label{sec:arima}

The Autoregressive Integrated Moving Average (ARIMA) model proposed by \cite{BoxJenkins1990} is the most general class of models for forecasting time series due to its simplicity of application and capability of handling non-stationary data. The AR part of ARIMA indicates that the variable of interest is regressed on its own lagged values. The MA part indicates that the regression error is a linear combination of error values that occurred in the past. Finally, the I (for ``integrated'') part represents the order of differencing to turn the time series into a stationary series (if necessary). Differencing means replacing the original series by the difference between their values and the previous values \citep{BoxJenkins1990}. The ARIMA model that includes other time series as input variables (exogenous variables) is referred to as Autoregressive Integrated Moving Average Exogenous inputs (ARIMAX) model.

The parameters of ARIMAX($p,d,q,n$) model are: $p$, the number of autoregressive terms; $d$, the number of nonseasonal differences needed for stationary; $q$, the number of lagged forecast errors in the prediction equation; $n$, the number of exogenous variables; $\eta$, a constant; and, $\phi_i$, for $i=1,\ldots, p$, $\theta_j$, for $j=1, \ldots q$, and $\zeta_l$, for $l=1,\ldots,n$, the model parameters. Mathematically, this model can be formulated as in Equation~\eqref{eq:arima}.

\begin{equation}
\label{eq:arima} W_t = \eta + \displaystyle\sum_{i=1}^{p}\phi_i W_{t-i} - \displaystyle\sum_{j=1}^{q}\theta_j e_{t-j} + \displaystyle\sum_{l=1}^{n} \zeta_l Y_l, 
\end{equation}

\noindent where $W_t$ and $W_{t-i}$, for $i=1,\ldots,p$, are the predicted values of the time series; $Y_l$, for $l=1,\ldots,n$, are the exogenous variables; and $e_{t-j}$, for $j=1,\ldots,q$, represent the error terms.

\subsection{EEMD-ARIMAX}
\label{sec:eemd-arima}

To our knowledge, EEMD has not yet been used to predict time series with independent variables. The main idea behind EEMD-ARIMAX is to predict time series of independent and dependent variables. For this, we first decompose each time series of the independent variables ($Y_1, Y_2, \ldots, Y_n$) and dependent variables by applying the EEMD method, creating $s$ levels of decomposition for each variable. Then, in each level of the decomposition, we use the ARIMAX method to predict the IMFs related to the dependent variables, by considering the IMFs of the variables $Y_1, Y_2, \ldots, Y_n$ as the exogenous variables. We employ the same procedure to predict the time series of the residual values. Finally, by summing the predicted time series, we obtain the prediction for the original time series of the daily number of COVID-19 cases. The algorithm of the proposed EEMD-ARIMAX method can be described by steps  \ref{step1}-\ref{laststep}:

\begin{enumerate}
    \item Let $X_t$ be dependent variable under study,  $Y_1$, $Y_2,\ldots, Y_n$  the independent/predictor variables, $m$ the number of ensembles, and $s$ the number of IMFs that will be extracted of each time series; \label{step1}
    \item Apply EEMD to decompose the time series of the dependent and independent variables individually, to obtain a set of $s$ IMFs and a  time series Res, in each decomposition; 
    \item Fit IMFs of the same ``levels'' using ARIMAX -- meaning that the $j$-th IMF of the time series represented by the ``Daily number of COVID-19 cases'', denoted here by IMF$_{X_t}^{j}$, will be fitted by the $j$-th IMFs of the same level $j$ of the time series related to meteorological/mobility variables $Y_l$, denoted by IMF$_{Y_l}^{j}$, for all $l=1,\ldots,n$. The estimated $j$-th IMF is denoted  by $\hat{\mbox{IMF}}^{j}$;\label{imf1}
    \item Denote  the residual values obtained by applying EEMD in $Y_1,\ldots Y_n$ by Res$_{Y_1}$,$\ldots$,Res$_{Y_n}$, respectively. Denote the residual value found by applying EEMD in $X_t$ by Res$_{X_t}$. Let $\hat{\mbox{Res}}$ be the estimated time series of  Res$_{X_t}$ through ARIMAX using Res$_{Y_1}$,$\ldots$,Res$_{Y_n}$ as exogenous variables;\label{imf2}
    \item Denote the fitted values of variable $X_t$ by $\hat{X}_t$. Thereby,  $\hat{X}_t = \hat{\mbox{IMF}}^1 + \ldots + \hat{\mbox{IMF}}^s + \hat{\mbox{Res}}$. \label{laststep}
\end{enumerate}

\begin{figure}
    \centering
\begin{tikzpicture}[node distance=1.5cm,
    every node/.style={fill=white}, align=center]
  \tikzstyle{every node}=[font=\small]
  \tikzstyle{decision} = [diamond, minimum width=3cm, minimum height=1cm, text centered, draw=black, fill=green!30]
  \tikzstyle{io} = [trapezium, trapezium left angle=70, trapezium right angle=110, minimum width=3cm, minimum height=1cm, text centered, draw=black, fill=blue!30]
  \node (start)             [Start]              {Start};
  \node (data)     [io, right of=start, xshift=6cm]          {Data: Daily number of COVID-19 cases, $X_t$; \\ Time series (meteorological and mobility), $Y_l, l=1,\ldots,n$ \\ Number of ensembles, $m$; Number of IMFs, $s$};
  
  \node (EEMD)      [process, below of=data, yshift=-1cm]   {Decompose all time series by EEMD to obtain: \\ IMF$^{j}_{Y_l}$, Res$_{l}$, $j=1,\ldots,s, \ l=1, \ldots,n$};

   \node (initialj)      [process, below of=EEMD, yshift=-1cm]   { $j=1$};

  \node (fit)     [process, below of=initialj, yshift=-1cm]   {
  Obtain $\hat{\mbox{IMF}}^j$: the fitted IMF$_{X_t}^{j}$ by  ARIMAX \\ with IMF$_{Y_1}^{j}$,\ldots,IMF$_{Y_n}^{j}$ as the exogenous variables};
  \node (question)      [decision, below of=fit, yshift=-1cm]
                                                      {$j > s$?};
  \node (residual)      [process, below of=question, yshift=-1cm]
                                                                {
                                                                Obtain $\hat{\mbox{Res}}$: the fitted Res$_{X_t}$ by using ARIMAX \\ with Res$_{Y_1}$,\ldots,Res$_{Y_n}$ being the exogenous variables};
  \node (predict)      [process, below of=residual, yshift=-1cm]
                                                        {Obtain the predict $\hat{X}_t$ of $X_t$: \\ $\hat{X}_t = \displaystyle\sum_{j=1}^{s} \hat{\mbox{IMF}}^j + \hat{\mbox{Res}}$};

   \node (no) [process, right of=question, xshift=3cm] {$j = j+1$};                                                       
    \node (end) [end, right of=predict, xshift=3cm]
                                                    {End};     
\draw[->] (start) -- (data);
\draw[->] (data) -- (EEMD);
\draw[->] (EEMD) --  (initialj);
\draw[->] (initialj) --  (fit);
\draw[->] (fit) -- (question);
\draw[->] (question) -- node[xshift=0.5cm] {Yes} (residual);
\draw[->] (residual) -- (predict);
\draw[->] (question) -- node[anchor=south] {No} (no);
\draw[->] (no) |- (fit);
\draw[->] (predict) -- (end);
\end{tikzpicture}
\caption{Flowchart  of the proposed EEMD-ARIMAX method.}
    \label{fig:floweemd-arima}
\end{figure}
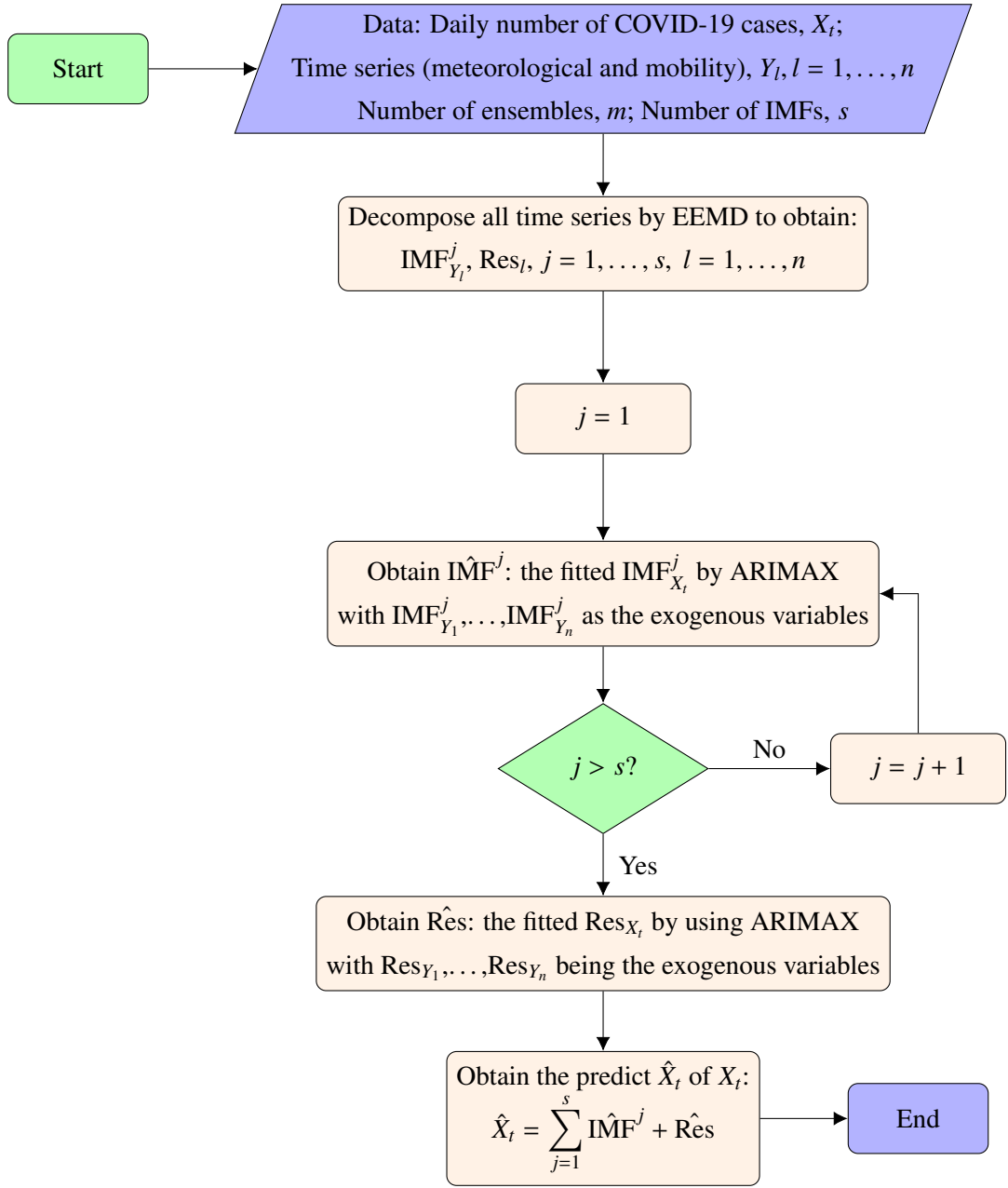

A flowchart of the proposed EEMD-ARIMAX method is presented in Figure  \ref{fig:floweemd-arima}.

\section{Results and discussion}
\label{sec:results}

We apply a lag of 5 days in the number of new confirmed COVID-19 daily cases, since symptoms start five days after someone is infected, and the patients seek medical advice \citep{He2020}. All studies were performed considering the database with this lag. We used R statistical software \citep{softwareR} in all tests carried out for this paper.

\subsection{Correlation analysis}

We evaluate the pairwise correlation between the number of COVID-19 cases and meteorological/mobility variables using Spearman correlation. For more details about this measure, see \ref{appendix-spearman}.

Tables \ref{tab:Spearman-weather} and \ref{tab:Spearman-mobility} show the correlation values  -- columns ``$\rho$'' -- between the number of COVID-19 cases and the meteorological and human mobility variables, respectively, for all Brazilian capitals, in the period considered. In addition, these tables present the $p$-value regarding the statistical significance of the corresponding variables at a significance level of $\alpha=0.01$. Therefore, if the $p$-value of the indicated correlation is less than or equal to $0.01$, the correlation is said to be statistically significant.

\begin{table}[!htpb]
    \centering
    \caption{Spearman correlation between the number of COVID-19 cases and the meteorological data.}    \label{tab:Spearman-weather}
    \scriptsize{
\begin{tabular}{|c|l|c|c|c|c|c|c|c|c|c|c|c|}
  \hline
  \multirow{2}{*}{Region} & \multirow{2}{*}{City-Federative unit} & \multicolumn{2}{c|}{Rain (mm)} & \multicolumn{2}{c|}{Max Temp (ºC)} & \multicolumn{2}{c|}{Min Temp (ºC)} & \multicolumn{2}{c|}{Hum (\%)}\\ \cline{3-10}
  & & $\rho$ & p-value & $\rho$ & p-value & $\rho$ & p-value & $\rho$ & p-value \\ \hline
  \multirow{7}{*}{North} & Belém-PA & 0.005 & 0.945 & 0.137 & 0.037 & -0.013 & 0.842 & -0.137 & 0.037\\
  & Boa Vista-RR & 0.154 & 0.019 & -0.201 & 0.002 & -0.290 & 0.000 & -0.071 & 0.285  \\
  & Macapá-AP &  -0.015 & 0.817 & 0.133 & 0.043 & 0.052 & 0.433 & -0.021 & 0.749\\
  & Manaus-AM & 0.009 & 0.885 &-0.051 & 0.429 & 0.028 & 0.672 &-0.013 & 0.843 \\
  & Palmas-TO &  \cellcolor{gray!20} -0.527 & 0.000 & \cellcolor{gray!60} 0.315 & 0.000 & \cellcolor{gray!20} -0.489 & 0.000 & \cellcolor{gray!20} -0.614 & 0.000  \\
  & Porto Velho-RO & -0.293 & 0.000 & 0.135 & 0.040 & \cellcolor{gray!20} -0.349 & 0.000 &-0.261 & 0.000  \\
  & Rio Branco-AC & -0.170 & 0.009 &-0.191 & 0.003 &-0.241 & 0.000 & 0.067 & 0.307   \\ \hline
  \multirow{9}{*}{Northeast} & Aracaju-SE & 0.064 & 0.324 & \cellcolor{gray!20} -0.527 & 0.000 & \cellcolor{gray!20} -0.433 & 0.000 & -0.003 & 0.959\\
  & Fortaleza-CE & 0.219 & 0.001 & \cellcolor{gray!20} -0.325 & 0.000 & 0.013 & 0.837 & \cellcolor{gray!60} 0.329 & 0.000 \\
  & João Pessoa-PB & 0.206 & 0.002 & \cellcolor{gray!20} -0.517 & 0.000 & \cellcolor{gray!20} -0.305 & 0.000 & 0.118 & 0.071 \\
  & Maceió-AL & \cellcolor{gray!60} 0.355 & 0.000 & \cellcolor{gray!20} -0.570 & 0.000 & \cellcolor{gray!20} -0.372 & 0.000 & 0.153 & 0.016  \\
  & Natal-RN & -0.012 & 0.857 & \cellcolor{gray!20} -0.321 & 0.000 & -0.293 & 0.000 & -0.171 & 0.008 \\
  & Recife-PE & 0.267 & 0.000 & \cellcolor{gray!20} -0.367 & 0.000 & -0.139 & 0.031 & \cellcolor{gray!60} 0.316 & 0.000  \\
  & Salvador-BA & 0.004 & 0.953 & \cellcolor{gray!20} -0.482 & 0.000 & \cellcolor{gray!20} -0.511 & 0.000 & -0.058 & 0.368  \\
  & São Luis-MA & \cellcolor{gray!60} 0.339 & 0.000 & \cellcolor{gray!20} -0.402 & 0.000 & -0.017 & 0.801 & 0.289 & 0.000  \\
  & Teresina-PI & \cellcolor{gray!20} -0.520 & 0.000 & \cellcolor{gray!60} 0.393 & 0.000 & \cellcolor{gray!20} -0.361 & 0.000 & \cellcolor{gray!20} -0.585 & 0.000  \\ \hline
  \multirow{4}{*}{Midwest} & Brasilia-DF & \cellcolor{gray!20} -0.609 & 0.000 & -0.111 & 0.083 & \cellcolor{gray!20} -0.677 & 0.000 & \cellcolor{gray!20} -0.597 & 0.000  \\
  & Campo Grande-MS & -0.086 & 0.188 & 0.098 & 0.130 & -0.079 & 0.224 & -0.209 & 0.001  \\
  & Cuiabá-MT & -0.127 & 0.053 & \cellcolor{gray!20} -0.482 & 0.000 & \cellcolor{gray!20} -0.590 & 0.000 & 0.263 & 0.000 \\
  & Goiânia-GO & \cellcolor{gray!20} -0.427 & 0.000 & 0.232 & 0.000 & -0.166 & 0.010 & \cellcolor{gray!20} -0.551 & 0.000 \\ \hline
  \multirow{4}{*}{Southeast} & Belo Horizonte-MG & -0.053 & 0.421 & -0.169 & 0.009 & -0.292 & 0.000 & -0.066 & 0.311\\
  & Rio de Janeiro-RJ & -0.077 & 0.227 & -0.194 & 0.002 & \cellcolor{gray!20} -0.336 & 0.000 & -0.078 & 0.225  \\
  & São Paulo-SP & -0.082 & 0.192 & -0.207 & 0.001 & \cellcolor{gray!20} -0.369 & 0.000 & -0.203 & 0.001  \\
  & Vitória-ES & 0.012 & 0.861 & -0.100 & 0.127 & -0.225 & 0.001 & -0.072 & 0.274  \\ \hline
  \multirow{4}{*}{South} & Curitiba-PR & 0.003 & 0.959 & -0.155 & 0.016 & -0.254 & 0.000 & -0.099 & 0.127  \\
  & Florianópolis-SC & 0.127 & 0.050 & -0.247 & 0.000 & -0.192 & 0.003 & 0.167 & 0.009 \\
  & Porto Alegre-RS & 0.214 & 0.001 & -0.227 & 0.000 & -0.118 & 0.068 & 0.154 & 0.016 \\
  \hline
\end{tabular}
}
\end{table}

\begin{table}[!htpb]
    \centering
    \caption{Spearman correlation between the number of COVID-19 cases and the human mobility data.}
    \label{tab:Spearman-mobility}
    \tiny{
\begin{tabular}{|c|l|c|c|c|c|c|c|c|c|c|c|c|c|}
  \hline
  \multirow{2}{*}{Region} &  \multirow{2}{*}{City-Federative unit}  & \multicolumn{2}{c|}{RR} & \multicolumn{2}{c|}{GP} & \multicolumn{2}{c|}{PA} & \multicolumn{2}{c|}{TS} & \multicolumn{2}{c|}{WO} & \multicolumn{2}{c|}{RE}\\ \cline{3-14}
  & & $\rho$ & p-value & $\rho$ & p-value & $\rho$ & p-value & $\rho$ & p-value & $\rho$ & p-value & $\rho$ & p-value\\ \hline
  \multirow{7}{*}{North} & Belém-PA & 0.041 & 0.534 & 0.102 & 0.121 & 0.082 & 0.209 & 0.020 & 0.766 & 0.127 & 0.052 & 0.010 & 0.882\\
  & Boa Vista-RR & \cellcolor{gray!50} 0.308 & 0.000 & \cellcolor{gray!50} 0.420 & 0.000 & 0.247 & 0.000 & \cellcolor{gray!50} 0.332 & 0.000 & \cellcolor{gray!50} 0.427 & 0.000 & -0.238 & 0.000 \\
  & Macapá-AP & -0.039 & 0.557 & 0.004 & 0.956 & -0.041 & 0.537 & -0.060 & 0.359 & 0.057 & 0.385 & 0.021 & 0.747 \\
  & Manaus-AM & 0.109 & 0.094 & 0.146 & 0.024 & 0.069 & 0.291 & 0.132 & 0.042 & 0.080 & 0.218 & 0.054 & 0.403 \\
  & Palmas-TO & \cellcolor{gray!50} 0.380 & 0.000 & \cellcolor{gray!50} 0.410 & 0.000 & \cellcolor{gray!50} 0.461 & 0.000 & \cellcolor{gray!50} 0.365 & 0.000 & \cellcolor{gray!50} 0.332 & 0.000 & \cellcolor{gray!20} -0.312 & 0.000  \\
  & Porto Velho-RO & 0.103 & 0.119 & 0.175 & 0.008 & 0.052 & 0.433 & 0.045 & 0.499 & 0.166 & 0.012 &-0.004 & 0.957 \\
  & Rio Branco-AC & -0.122 & 0.062 &-0.051 & 0.437 &-0.106 & 0.106 &-0.097 & 0.138 &-0.045 & 0.488 & 0.227 & 0.000 \\ \hline
  \multirow{9}{*}{Northeast} & Aracaju-SE & -0.021 & 0.741 & 0.075 & 0.248 & -0.145 & 0.025 & -0.068 & 0.299 & 0.056 & 0.389 & 0.093 & 0.151\\
  & Fortaleza-CE & \cellcolor{gray!20} -0.348 & 0.000 & -0.256 & 0.000 & \cellcolor{gray!20} -0.406 & 0.000 & \cellcolor{gray!20} -0.322 & 0.000 & -0.207 & 0.001 & \cellcolor{gray!50} 0.350 & 0.000\\
  & João Pessoa-PB & 0.023 & 0.723 & 0.162 & 0.013 & -0.092 & 0.160 & -0.058 & 0.378 & 0.169 & 0.010 & -0.029 & 0.661 \\
  & Maceió-AL & -0.267 & 0.000 &-0.185 & 0.004 &-0.290 & 0.000 &-0.243 & 0.000 &-0.233 & 0.000 & \cellcolor{gray!50} 0.345 & 0.000  \\
  & Natal-RN & -0.041 & 0.527 & 0.076 & 0.238 & -0.076 & 0.244 & 0.005 & 0.943 & 0.030 & 0.647 & 0.090 & 0.163 \\
  & Recife-PE & -0.286 & 0.000 & -0.175 & 0.007 & \cellcolor{gray!20} -0.313 & 0.000 & -0.279 & 0.000 & -0.177 & 0.006 & \cellcolor{gray!50} 0.336 & 0.000  \\
  & Salvador-BA & -0.015 & 0.820 & 0.126 & 0.052 & 0.003 & 0.969 & 0.008 & 0.901 & 0.087 & 0.178 & 0.024 & 0.716  \\
  & São Luis-MA & -0.297 & 0.000 & -0.192 & 0.003 & -0.290 & 0.000 & \cellcolor{gray!20} -0.339 & 0.000 & -0.222 & 0.001 & \cellcolor{gray!50} 0.316 & 0.000  \\
  & Teresina-PI &  \cellcolor{gray!50} 0.333 & 0.000 & 0.279 & 0.000 & \cellcolor{gray!50} 0.530 & 0.000 & \cellcolor{gray!50} 0.406 & 0.000 & \cellcolor{gray!50} 0.484 & 0.000 & \cellcolor{gray!20} -0.398 & 0.000 \\ \hline
  \multirow{4}{*}{Midwest} & Brasilia-DF & 0.258 & 0.000 & \cellcolor{gray!50} 0.341 & 0.000 & 0.294 & 0.000 & 0.228 & 0.000 & 0.251 & 0.000 & -0.153 & 0.016 \\
  & Campo Grande-MS &  \cellcolor{gray!50} 0.437 & 0.000 & \cellcolor{gray!50} 0.562 & 0.000 & \cellcolor{gray!50} 0.343 & 0.000 & \cellcolor{gray!50} 0.382 & 0.000 & \cellcolor{gray!50} 0.392 & 0.000 & -0.161 & 0.013 \\
  & Cuiabá-MT & \cellcolor{gray!20} -0.534 & 0.000 & \cellcolor{gray!20} -0.369 & 0.000 & \cellcolor{gray!20} -0.597 & 0.000 & \cellcolor{gray!20} -0.544 & 0.000 & \cellcolor{gray!20} -0.352 & 0.000 & \cellcolor{gray!50} 0.519 & 0.000\\
  & Goiânia-GO &  \cellcolor{gray!50} 0.491 & 0.000 & \cellcolor{gray!50} 0.608 & 0.000 & \cellcolor{gray!50} 0.431 & 0.000 & \cellcolor{gray!50} 0.500 & 0.000 & \cellcolor{gray!50} 0.482 & 0.000 & \cellcolor{gray!20} -0.472 & 0.000 \\ \hline
  \multirow{4}{*}{Southeast} & Belo Horizonte-MG & 0.099 & 0.128 & 0.195 & 0.003 & 0.128 & 0.050 & 0.162 & 0.012 & \cellcolor{gray!50} 0.311 & 0.000 & -0.235 & 0.000\\
  & Rio de Janeiro-RJ & -0.106 & 0.099 & -0.011 & 0.862 & -0.145 & 0.023 & -0.087 & 0.176 & -0.009 & 0.888 & 0.098 & 0.126 \\
  & São Paulo-SP & -0.197 & 0.002 & -0.006 & 0.929 & -0.118 & 0.059 & -0.163 & 0.009 & -0.059 & 0.344 & 0.137 & 0.029 \\
  & Vitória-ES & 0.280 & 0.000 & \cellcolor{gray!50} 0.339 & 0.000 & 0.027 & 0.683 & 0.253 & 0.000 & \cellcolor{gray!50} 0.369 & 0.000 & -0.245 & 0.000 \\ \hline
  \multirow{4}{*}{South} & Curitiba-PR & 0.155 & 0.016 & 0.137 & 0.034 & 0.149 & 0.021 & 0.127 & 0.049 & 0.237 & 0.000 & -0.185 & 0.004 \\
  & Florianópolis-SC & \cellcolor{gray!50} 0.548 & 0.000 & \cellcolor{gray!50} 0.532 & 0.000 & \cellcolor{gray!50} 0.367 & 0.000 & \cellcolor{gray!50} 0.608 & 0.000 & \cellcolor{gray!50} 0.500 & 0.000 & \cellcolor{gray!20} -0.563 & 0.000\\
  & Porto Alegre-RS & \cellcolor{gray!50} 0.379 & 0.000 & \cellcolor{gray!50} 0.453 & 0.000 & 0.146 & 0.023 & \cellcolor{gray!50} 0.421 & 0.000 & \cellcolor{gray!50} 0.410 & 0.000 & \cellcolor{gray!20} -0.372 & 0.000 \\
  \hline
\end{tabular}
}
\end{table}

We consider that two variables are correlated if $\rho \geq 0.3$ or $\rho \leq -0.3$. As stated before,  if $\rho$ is positive,  the variables are directly proportional,  otherwise, they are inversely proportional. Therefore, on the one hand, we say that there is a positive correlation between a  pair of  variables when $\rho \geq 0.3$, meaning that there is evidence that the variables grow together.  On the other, when the correlation is negative, i.e.,  $ \rho \leq -0.3$, it means that  the analyzed pair of variables has an opposite behavior: the greater the values of one variable, the smaller  the values of the other variable. For better visualization, we highlighted the positive correlations in dark gray, and negative correlations in light gray.

According to the results, the number of COVID-19 cases and meteorological variables were correlated in 16 cities. In 11 of them, the correlated meteorological variable was the minimum temperature.  The number of COVID-19 cases and meteorological variables were not correlated in any of the cities in the South region. The maximum temperature and the number of COVID-19 cases were correlated in all cities in the Midwest region.

The correlations between the number of COVID-19 cases and minimum temperature were negative, indicating that the number of cases increases when the minimum temperature decreases. The same behavior was observed between the number of COVID-19 cases and maximum temperature, except in Teresina-PI and Palmas-TO. In these two cities, the relationship between the number of COVID-19 cases and maximum temperature was inversely proportional.

Humidity and the number of daily COVID-19 cases are correlated  in the following cities: Palmas-TO, Fortaleza-CE, Recife-PE, Teresina-PI, Brasília-DF, and Goiânia-GO. Particularly in Palmas-TO and Teresina-PI, the humidity and number of COVID-19 cases showed a strong correlation. The average humidity of Palmas-TO was the lowest among the capitals of the North region. The average humidity of Teresina-PI was the second lowest average of the capitals of the Northeast region. Since humidity is directly linked to temperature, these facts could explain the inversely proportional correlations between the number of COVID-19 cases and the maximum temperature in both cities.

Among the 6 capitals that showed a correlation between the number of COVID-19 cases and humidity,  Fortaleza-CE and Recife-PE presented correlations of 0.329 and 0.316 respectively. The other four capitals showed negative correlations. One can observe that the rainfall variable and the number of COVID-19 cases are not correlated in Fortaleza-CE and Recife-PE. In Palmas-TO, Teresina-PI, Brasília-DF, and Goiânia-GO, on the other hand, it is possible to see that they were negatively correlated. 

It is known that meteorological data regarding temperature, humidity, and rainfall are related and, therefore, influence one another. In this study, however, we will only consider the meteorological variables of each capital that had a correlation with the number of COVID-19 cases greater than 0.3, in absolute value. These values are summarized in  Table \ref{tab:correlation}.

\begin{table}[!htpb]
    \centering
    \caption{Variables per Brazilian capital which showed some level of correlation with the number of COVID-19 cases and  were considered in the proposed models.}
    \label{tab:correlation}
    \scriptsize{
\begin{tabular}{|c|l|l|l|}
  \hline
  Region & City-Federative unit  & Meteorological variables & Mobility variables\\ \hline
  \multirow{7}{*}{North} & Belém-PA & - & -\\
  & Boa Vista-RR & - & RR, GP, TS, WO\\
  & Macapá-AP & - & -\\
  & Manaus-AM & - & -\\
  & Palmas-TO & Rain, Max Temp,  Min Temp,  Hum & RR, GP, PA, TS, WO, RE \\
  & Porto Velho-RO & Min Temp & -\\
  & Rio Branco-AC & - & - \\ \hline
  \multirow{9}{*}{Northeast} & Aracaju-SE & Max Temp, Min Temp & -\\
  & Fortaleza-CE & Max Temp,  Hum & RR, PA, TS, RE\\
  & João Pessoa-PB & Max Temp,  Min Temp & - \\
  & Maceió-AL & Rain,  Max Temp,  Min Temp & RE \\
  & Natal-RN & Max Temp & -\\
  & Recife-PE & Max Temp,  Hum & PA, RE \\
  & Salvador-BA & Max Temp, Min Temp & -\\
  & São Luis-MA & Rain,  Max Temp & TS, RE \\
  & Teresina-PI &   Rain,  Max Temp,  Min Temp,  Hum & RR, PA, TS, WO, RE \\ \hline
  \multirow{4}{*}{Midwest} & Brasilia-DF & Rain,  Min Temp,  Hum & GP\\
  & Campo Grande-MS & - & RR, GP, PA, TS, WO\\
  & Cuiabá-MT & Max Temp,  Min Temp & RR, GP, PA, TS, WO, RE\\
  & Goiânia-GO & Rain,  Hum & RR, GP, PA, TS, WO, RE \\ \hline
  \multirow{4}{*}{Southeast} & Belo Horizonte-MG & - & WO\\
  & Rio de Janeiro-RJ &  Min Temp & -\\
  & São Paulo-SP & Min Temp & -\\
  & Vitória-ES & - & GP, WO \\ \hline
  \multirow{4}{*}{South} & Curitiba-PR & - & -\\
  & Florianópolis-SC & - & RR, GP, PA, TS, WO, RE\\
  & Porto Alegre-RS & - & RR, GP, TS, WO, RE \\
  \hline
\end{tabular}
}
\end{table}



As mentioned before, Table \ref{tab:Spearman-mobility} shows the correlation between the mobility variables and the number of COVID-19 cases.  The mobility variables and the corresponding cities with which the number of COVID-19 cases have a positive correlation are:

\begin{itemize}
    \item Retail and recreation: Boa Vista-RR, Palmas-TO, Teresina-PI, Campo Grande-MS, Goiânia-GO, Florianópolis-SC, Porto Alegre-RS;
    \item Grocery and pharmacy: Boa Vista-RR, Palmas-TO, Teresina-PI, Brasília-DF, Campo Grande-MS, Goiânia-GO, Vitória-ES, Florianópolis-SC, Porto Alegre-RS; 
    \item Parks: Palmas-TO, Teresina-PI, Campo Grande-MS, Goiânia-GO, Florianópolis-SC;
    \item Transit stations: Boa Vista-RR, Palmas-TO, Teresina-PI, Campo Grande-MS, Goiânia-GO, Florianópolis-SC, Porto Alegre-RS;
    \item Workplaces: Boa Vista-RR, Palmas-TO, Teresina-PI, Campo Grande-MS, Goiânia-GO, Belo Horizonte-MG, Vitória-ES, Florianópolis-SC, Porto Alegre-RS;
    \item Residential: Fortaleza-CE, Maceió-AL, Recife-PE, São Luis-MA, Cuiabá-MT.
\end{itemize}

On the one hand, the positive correlation between the mobility parameters and the number of COVID-19 cases, except for the Residential variable, shows that the increase in the number of COVID-19 cases is directly proportional to the rise in the populations' mobility trends in traffic, pharmacies, work, parks, and retail. This means that the higher the mobility rate, the greater the number of cases. On the other hand, some cities showed a negative correlation between mobility variables and the number of COVID-19 cases. They are:

\begin{itemize}
    \item Retail and recreation: Fortaleza-CE, Cuiabá-MT;
    \item Grocery and pharmacy: Cuiabá-MT;
    \item Parks: Fortaleza-CE, Recife-PE, Cuiabá-MT;
    \item Transit stations: Fortaleza-CE, São Luis-MA, Cuiabá-MT;
    \item Workplaces: Cuiabá-MT;
    \item Residential: Palmas-TO, Teresina-PI, Goiânia-GO, Florianópolis-SC, Porto Alegre-RS.
\end{itemize}

Therefore, for example,  the negative correlation between Residential and daily COVID-19 cases means that the fewer people stayed at home, the greater the number of COVID-19  cases.

The negative correlations between COVID-19 cases and meteorological parameters in Cuiabá-MT are due to a sequence of null values at the end of the series describing the number of COVID-19 cases. If these values were excluded, the correlation coefficient between these variables would be positive.



\subsection{Analysis of the number of predicted cases}

The EEMD-ARIMAX method was implemented in the R software using the ``\textit{Rlibeemd}'' and ``\textit{forecast}''. We generated 125 new time series for each variable considering that the standard deviation of Gaussian noise was 1\% of the standard deviation of the corresponding original time series.

Table \ref{tab:resul-eemd} shows the results of the EEMD-ARIMAX method for all Brazilian capitals. We compared  EEMD-ARIMAX  with the ARIMAX method. The objective was to analyze the effect of  EEMD  on the prediction. In both methods and for each city, we present the widely employed mean error (ME), root-mean-square deviation (RMSE), and mean absolute error (MAE) measures to describe the results of the predictions. For details about these measures, see Section \ref{appendix-spearman}. Column ``City-Federative unit''  shows the pair city-federative unit and the parameters used by ARIMAX to forecast the number of COVID-19 cases in this capital. These parameters were calibrated for each city using  \textit{auto.arima()} function in R. The independent variables that were considered to predict the number of cases of COVID-19 in each corresponding city are shown in Table \ref{tab:correlation} and follow the Spearman correlation coefficients shown in Tables \ref{tab:Spearman-weather} and \ref{tab:Spearman-mobility}.  

In all cities, the proposed decomposition method improved the predictions of the time series in terms of RMSE values. The average RMSE of the predictions considering only the ARIMAX method was  211.987 with a standard deviation of 186.335. Using the EEMD-ARIMAX method, the average RMSE was 155.330 with a standard deviation of 145.645. EEMD-ARIMAX showed an improvement of 26.73\% over ARIMAX. \ref{appendix-graphics} presents some graphics comparing the original time series with the predicted values by EEMD-ARIMAX in all Brazilian regions.


\begin{table}[!htpb]
      \caption{Results achieved by ARIMAX and EEMD-ARIMAX methods.}
    \label{tab:resul-eemd}  
    \centering
    \scriptsize{
\begin{tabular}{|c|l|c|c|c|c|c|c|}
  \hline
  \multirow{2}{*}{Region} & \multirow{2}{*}{City-Federative unit}  & \multicolumn{3}{c|}{ARIMAX} & \multicolumn{3}{c|}{EEMD-ARIMAX}\\ \cline{3-8}
  & & ME & RMSE & MAE & ME & RMSE & MAE \\ \hline
  \multirow{7}{*}{North} & Belém-PA (1,0,1) & 3.213 & 139.121 & 100.688 & -5.891 & 89.189 & 61.905\\
  & Boa Vista-RR (0,1,1) & 5.092 & 235.997 & 123.145 & -2.248 & 159.335 & 97.518\\
  & Macapá-AP (3,0,2) & 0.229 & 185.697 & 79.845 & -7.852 & 148.587 & 69.708 \\
  & Manaus-AM (2,1,3) & 9.106 & 213.732 & 152.327 & -12.994 & 142.140 & 102.373 \\
  & Palmas-TO (2,0,2) &-0.223 & 62.519 & 37.374 & -0.168 & 53.240 & 34.604 \\
  & Porto Velho-RO (2,1,3) & 6.074 & 186.198 & 104.206 & 0.266 & 162.095 & 98.116 \\
  & Rio Branco-AC (0,1,2) & 1.088 & 46.482 & 28.736 & -1.487 & 30.006 & 18.009 \\ \hline
  \multirow{9}{*}{Northeast} & Aracaju-SE (0,1,1) & 1.413 & 163.284 & 91.061 & -0.183 & 107.398 & 58.659\\
  & Fortaleza-CE (1,0,1) & 7.275 & 255.173 & 139.918 & 0.023 & 150.085 & 98.026 \\
  & João Pessoa-PB (3,0,2) & 6.512 & 104.106 & 73.007 & -0.629 & 61.758 & 42.498 \\
  & Maceió-AL (2,1,2) & 0.646 & 89.629 & 54.596 & 0.010 & 61.337 & 41.793 \\
  & Natal-RN (1,0,3) & 4.485 & 209.895 & 104.098 & -12.078 & 183.497 & 99.912 \\
  & Recife-PE (3,0,2) & 3.028 & 144.336 & 82.521 & -6.969 & 114.020 & 69.529 \\
  & Salvador-BA (0,1,3) & 3.456 & 329.488 & 197.839 & -2.474 & 214.520 & 134.791\\
  & São Luis-MA (2,0,3) & 2.091 & 54.378 & 32.419 & -2.704 & 32.342 & 19.566 \\
  & Teresina-PI (2,0,3) & 0.703 & 78.607 & 59.522 & -4.655 & 56.288 & 43.008 \\ \hline
  \multirow{4}{*}{Midwest} & Brasilia-DF (0,1,4) & 5.992 & 272.017 & 173.067 & 3.756 & 176.154 & 111.602\\
  & Campo Grande-MS (0,1,4) & 4.305 & 130.909 & 65.215 & -7.399 & 87.561 & 50.282\\
  & Cuiabá-MT (1,0,4) & 1.487 & 58.115 & 28.884 & 1.521 & 35.695 & 19.566 \\
  & Goiânia-GO (4,1,1) & 7.194 & 262.247 & 165.356 & -11.852 & 209.587 & 150.245 \\ \hline
  \multirow{4}{*}{Southeast} & Belo Horizonte-MG (2,1,3) & 6.596 & 273.975 & 183.085 & -1.762 & 186.489 & 123.697\\
  & Rio de Janeiro-RJ (2,1,3) & 12.532 & 444.872 & 294.399 & -15.907 & 318.717 & 227.619 \\
  & São Paulo-SP (0,1,5) & 18.479 & 988.765 & 660.780 & 4.203 & 775.817 & 494.870 \\
  & Vitória-ES (1,0,2) & 3.598 & 62.274 & 42.943 & 1.440 & 41.234 & 27.729 \\ \hline
  \multirow{4}{*}{South} & Curitiba-PR (5,1,0) & 0.843 & 127.775 & 76.223 & -9.406 & 103.601 & 61.407 \\
  & Florianópolis-SC (0,1,3) & 26.907 & 230.129 & 80.612 & 15.483 & 210.716 & 86.626 \\
  & Porto Alegre-RS (0,1,1) & 25.704 & 373.925 & 140.315 & -23.187 & 282.510 & 140.701 \\
  \hline
\end{tabular}
}

\end{table}

\section{Anomaly analysis}\label{sec:anomaly}

The data used in the case study have several registration errors that may affect the accuracy of the prediction model. We used an anomaly detection strategy to identify whether there is a relationship between data errors and significant errors in the values predicted by  EEMD-ARIMAX. The employed anomaly detection method uses the Fourier transform in graphs as a tool to analyze the daily variation in the number of COVID-19 cases in each region. Thus, it identifies days with potentially anomalous numbers of COVID-19 cases.

Section \ref{subsec:errors} presents a discussion about the strategy adopted to define and quantify the model errors. Section \ref{subsec:anomaly} shows the concept of anomaly adopted and a tool to highlight anomalies. Section \ref{subsec:compare} addresses the methodology employed to compare the errors of the model with the detected anomalies. Section \ref{subsec:normalized} presents the strategy adopted to correct the anomalies and run the model again.

In summary, the anomaly analysis shows that there is a direct relationship between the days when the EEMD-ARIMAX significantly missed the prediction and the days when the anomaly detection strategy pointed to an abnormality. This indicates that the data errors affected the models' effectiveness. After normalizing and correcting the data, EEMD-ARIMAX's accuracy showed a significant increase.

\subsection{Analyzing Model Errors}\label{subsec:errors}

We analyzed the days for which the model significantly missed the predicted number of cases for each city. The error made by the model was quantified by the difference between the observed and predicted number of cases, as shown in Equation \eqref{eq:error}.

\begin{equation}\label{eq:error}
    e_i^t = \left| 1 - \displaystyle\frac{c_i^t}{\hat{c}_i^t} \right|
\end{equation}

\noindent where $c_i^t$ and $\hat{c}_i^t$ are, respectively, the observed and predicted number of COVID-19 cases on day $t$ in city $i$. An error is considered significant when $e_i^t > TD(E_i)$, where  $E_i$ is the vector formed by the elements $e_i^t$ $\forall t$, and $TD$ is defined by Equation~\eqref{eq:threshold}, where $Mean(E_i)$ and $STD(E_i)$ are the arithmetic mean and standard deviation of $E_i$, respectively.

\begin{equation}
    TD(E_i) = Mean(E_i) + (1.5 \times STD(E_i))
    \label{eq:threshold}
\end{equation}

\begin{figure}[htbp]
\centerline{\includegraphics[width=11cm,height=8cm,keepaspectratio]{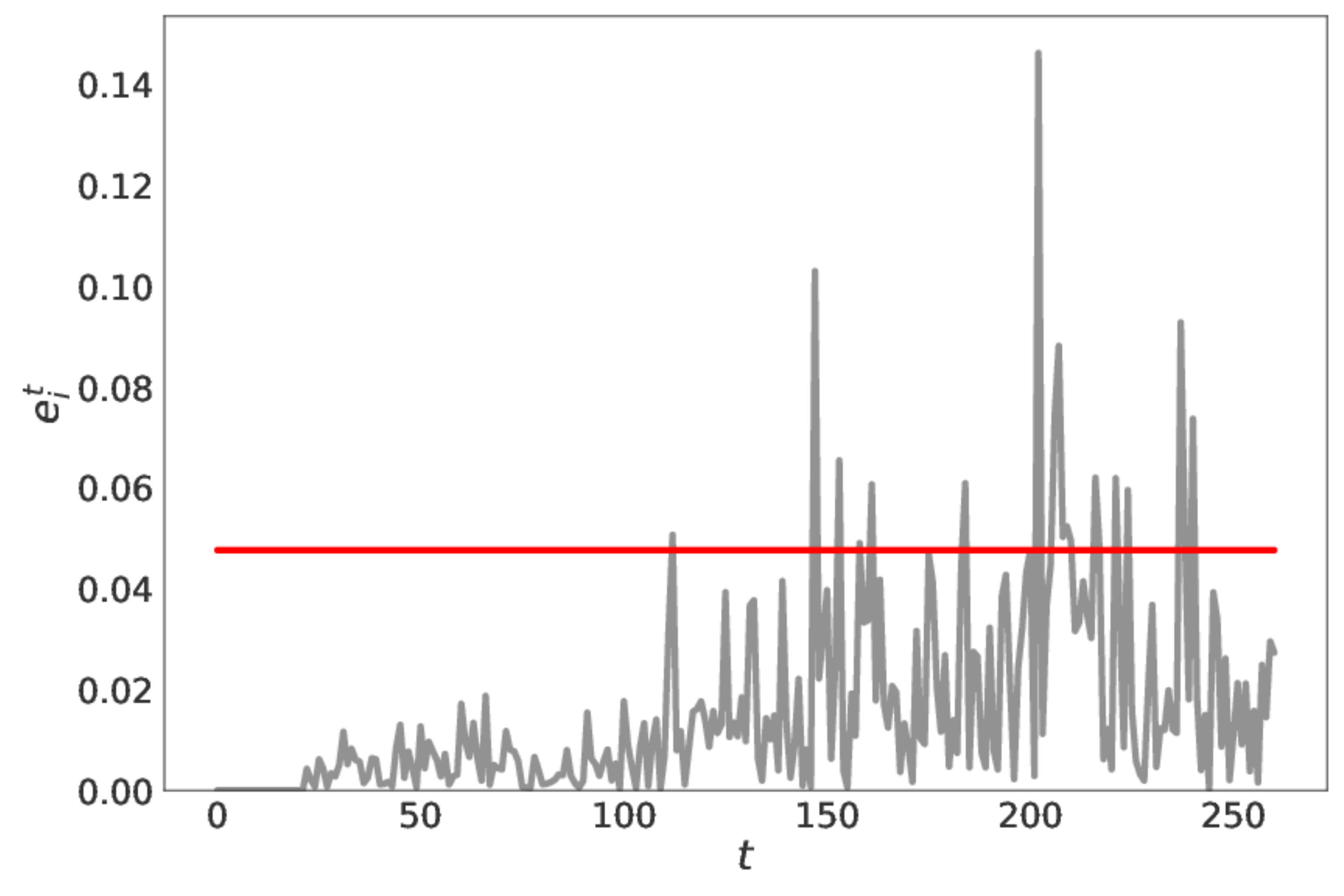}}
\caption{Model errors of Goiânia - GO and the significance threshold.}
\label{fig:threshold_error}
\end{figure}

 Figure \ref{fig:threshold_error} illustrates the values of $e_i^t$ considering the city of  Goiânia - GO. The threshold value $TD(E_i)$ is highlighted in red. Therefore,  every day $t$ whose $e_i^t$ is above the red line corresponds to a significantly mispredicted day by the model.

\subsection{Analyzing Data Anomalies}\label{subsec:anomaly}

A spectral anomaly detection strategy was adopted to detect days when the recorded number of daily COVID-19 cases was potentially anomalous. While the model errors are identified by comparing the predicted values with the observed values, the anomaly detection strategy analyzes the daily variation in the number of cases considering the distance between cities to identify potentially anomalous variations.

For example, if a city has a slight variation over two days in the number of COVID-19 cases, we expect nearby cities to have a similar variation. Similarly, if the number of cases in a city suffers a significant increase from one day to the other, we expect nearby cities also to have a relative increase in the number of cases.

To perform this analysis, we model a complete and weighted network where a node $v_i \in V$ represents a city, and the weight of the edge $w_{ij}$ is the Euclidean distance between cities $i$ and $j$. Each node $v_i$ carries the daily variation in the number of cases in city $i$, with the daily variation $s_i^t$ defined by Equation~\eqref{eq:sit},  $\forall t > 1$.

\begin{equation}
    s_i^t = \left| 1 - \frac{c_i^t}{c_i^{(t-1)}} \right| \label{eq:sit}
\end{equation}

A signal $S^t$ contains the values $s_i^t$ for every city $i$ in the dataset. We calculate the spectra $\hat{S^t}$ of $S^t$ signal, $ \forall t > 1$, using the Fourier transform for graphs \citep{AliakseiSandryhaila2012DISCRETEMoura}. In graph Fourier analysis, the graph Laplacian eigenvectors associated with small eigenvalues $\lambda_l$ vary slowly across the graph, whereas eigenvectors associated with larger eigenvalues oscillate more rapidly \citep{Ortega2018GraphApplications}. It means that if two vertices are connected by an edge with a large weight, the values of the eigenvector at those locations are likely to be similar. This concept is then used to define low  and high frequencies for signals indexed by graphs. 

\begin{figure}[htbp]
\centerline{\includegraphics[width=11cm,height=8cm,keepaspectratio]{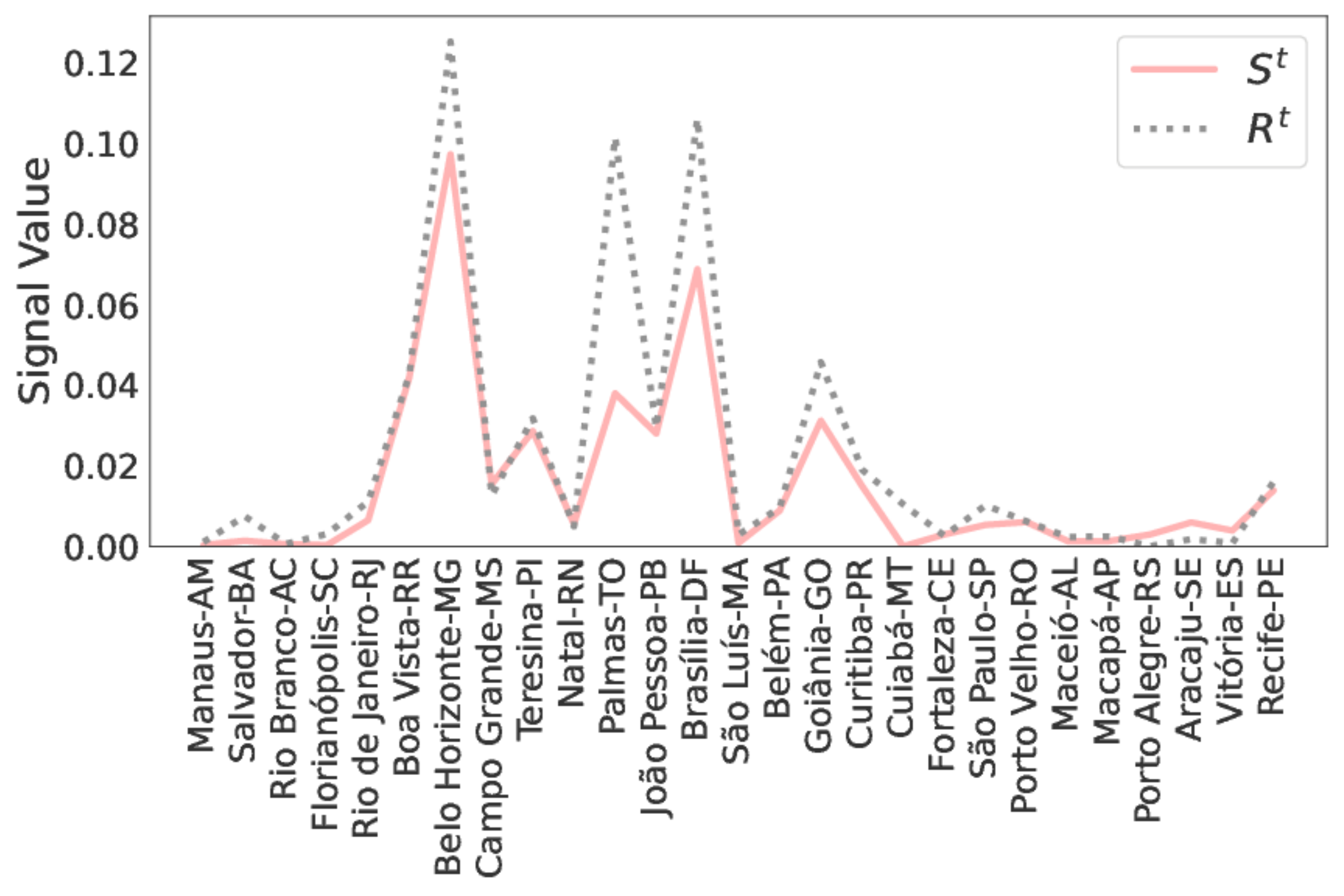}}
\caption{Observed variation versus accentuated variation in all 27 capitals of Brazil on August 17, 2020.}
\label{fig:atenuated_difference}
\end{figure}

According to this definition, abrupt oscillations are concentrated at the high frequencies of the signal spectrum. To highlight abrupt variations and expose anomalies, we accentuate the magnitude of the high frequencies of $\hat{S}^t$ spectrum to make anomalous variations more evident, generating a new spectrum $\hat{R}^{t}$. We apply the inverse Fourier transform to $\hat{R}^{t}$ to get a new $R^{t}$ signal, that contains the accentuated variation in the number of cases in each city. The intuition behind this operation is that, if the variation in the number of cases in a city $i$ is normal, then $r^{t}_i < s^t_i$ probably holds, where $r^{t}_i$ is the $i$-th element in vector $R^{t}$. On the other hand, if the city has an anomalous variation,  $r^{t}_i > s^t_i$ probably holds. Figure \ref{fig:atenuated_difference} shows a graphic visualization of the normal variation and the accentuated variation.

\begin{figure}[htbp]
\centerline{\includegraphics[width=11cm,height=7cm,keepaspectratio]{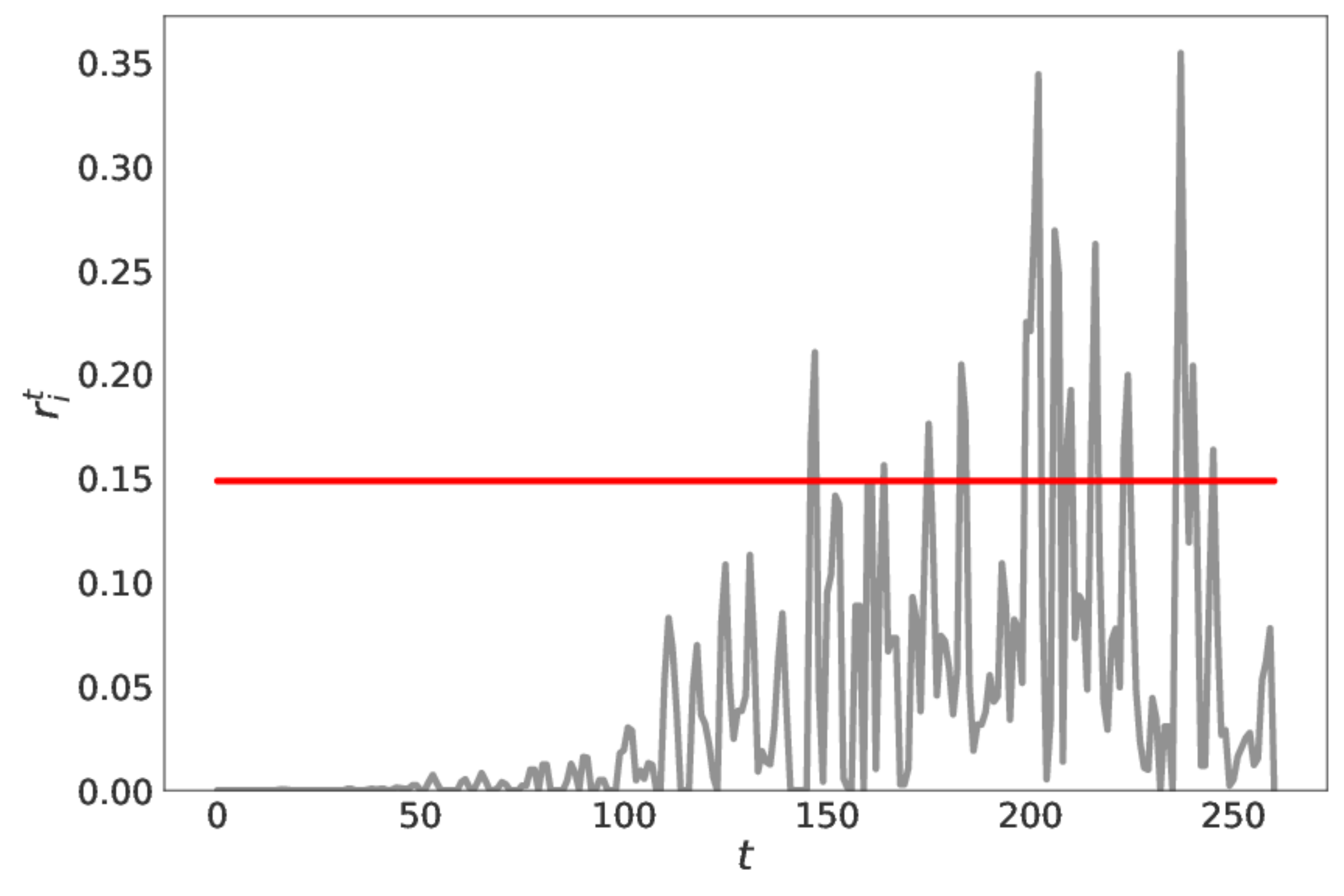}}
\caption{Accentuated daily variations in the number of COVID-19 cases in Goiânia - GO and the anomaly threshold.}
\label{fig:threshold_anomaly}
\end{figure}

The threshold used to determine whether a variation is anomalous or not is calculated in the same way for errors, as defined in Equation \eqref{eq:threshold}, for $R^{t}$. Figure \ref{fig:threshold_anomaly} illustrates the values of $R_i$, which is a vector with the $r_i^t$ of a given city $i$, and the threshold $TD(R_i)$. It is worth noting the similarity between Figures \ref{fig:threshold_error} and \ref{fig:threshold_anomaly}, which points out that there is a direct relation between the cases in which EEMD-ARIMAX significantly missed the prediction and the days when the attenuator pointed out potentially anomalous variations.

\subsection{Comparing Errors and Anomalies}\label{subsec:compare}

As presented in Section \ref{subsec:anomaly}, Figures \ref{fig:threshold_error} and \ref{fig:threshold_anomaly} indicate that there is a direct relationship between the days when EEMD-ARIMAX made a significant error and the days whose variation in the number of cases was interpreted as potentially anomalous. We compared the model's errors with anomalous variations to establish a quantifier that indicates whether there is, in fact, a direct relationship between them.

For each city, two sets were defined: set $CE$, containing the days on which the model made a significant error; and set $CA$, with the days whose variation was detected as potentially anomalous. To quantify the relationship between $CE$ and $CA$, we adopted the following criterion: if a day $t \in CE$, $1<t<nc$
and $\{t-1, t, t+1\} \cap CA = \emptyset$,
then the error made by the model on day $t$ and the anomalous variation that occurred on the days adjacent to $t$ are directly related.

\begin{figure}[htbp]
\centerline{\includegraphics[width=12cm,height=8cm,keepaspectratio]{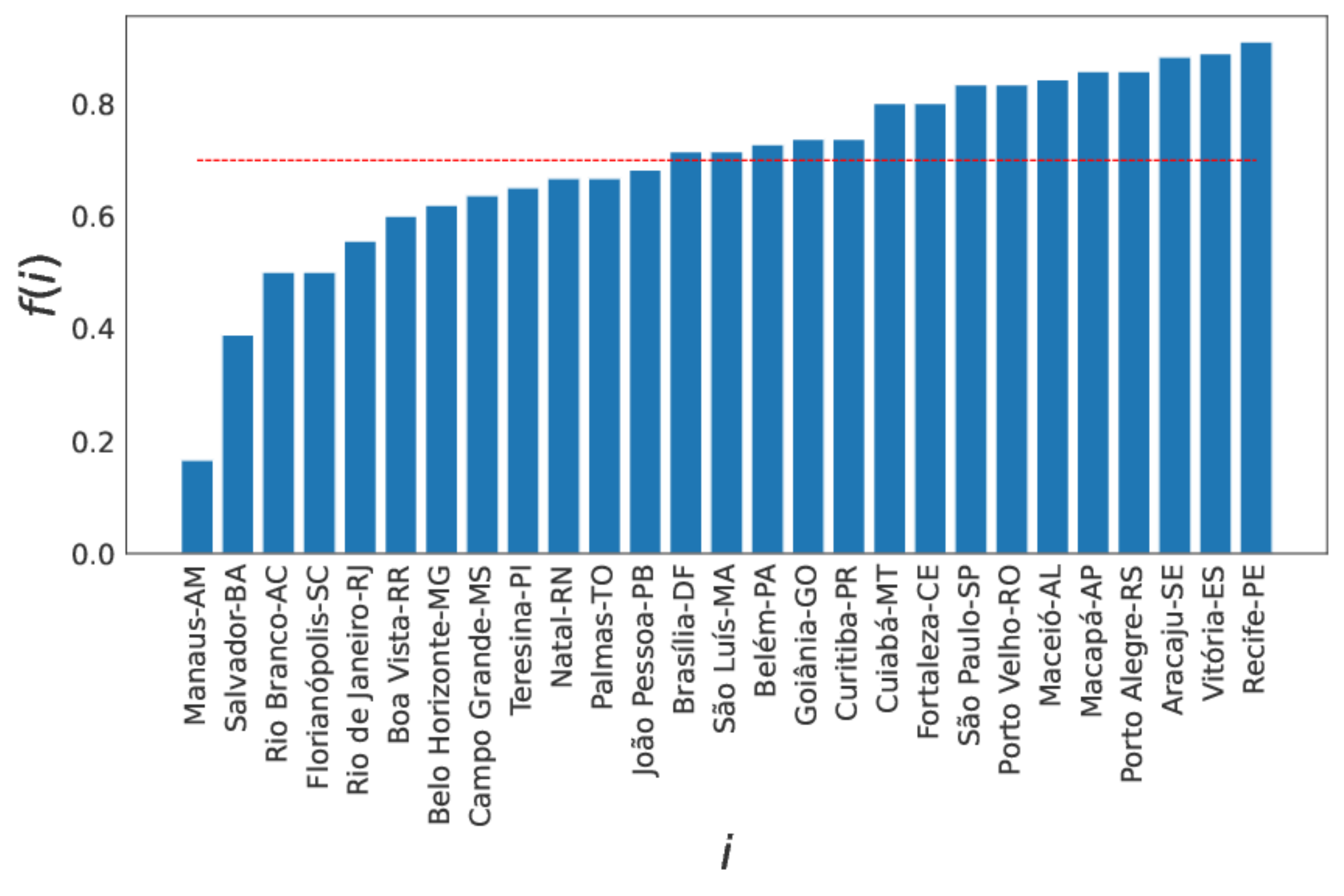}}
\caption{Percentage of days which EEMD-ARIMAX significantly mispredicted and corresponded to an anomaly.}
\label{fig:hits_percent}
\end{figure}

Figure \ref{fig:hits_percent} shows the percentage of days the model made significant mispredictions and which are directly related to a day with a potentially anomalous variation. On average, more than 60\% of the days when the model was wrong were detected as anomalous, as indicated by the red line, which represents the average. 

This result indicates that EEMD-ARIMAX was affected by errors in the data and the results point that the model's errors are directly related to anomalous variations. To overcome this problem and correct the anomalies, we adopted a spectral strategy for removing anomalies, also based on the Fourier transform. Section \ref{subsec:normalized} presents the methodology employed to correct the data anomalies.

\subsection{Normalizing Data}\label{subsec:normalized}

As discussed before, the abrupt variations are concentrated in the high frequencies. To detect the anomalies, the presented anomaly detection strategy accentuated the magnitude of the high frequencies. Then, to correct the anomalies, we adopted a strategy that does the opposite, using a low-pass filter that attenuates high frequencies. Unlike the high-frequency accentuator, the low pass filter decreases the magnitude of the high frequencies, attenuating abrupt variations.

\begin{figure}[htbp]
\centerline{\includegraphics[width=11cm,height=7cm,keepaspectratio]{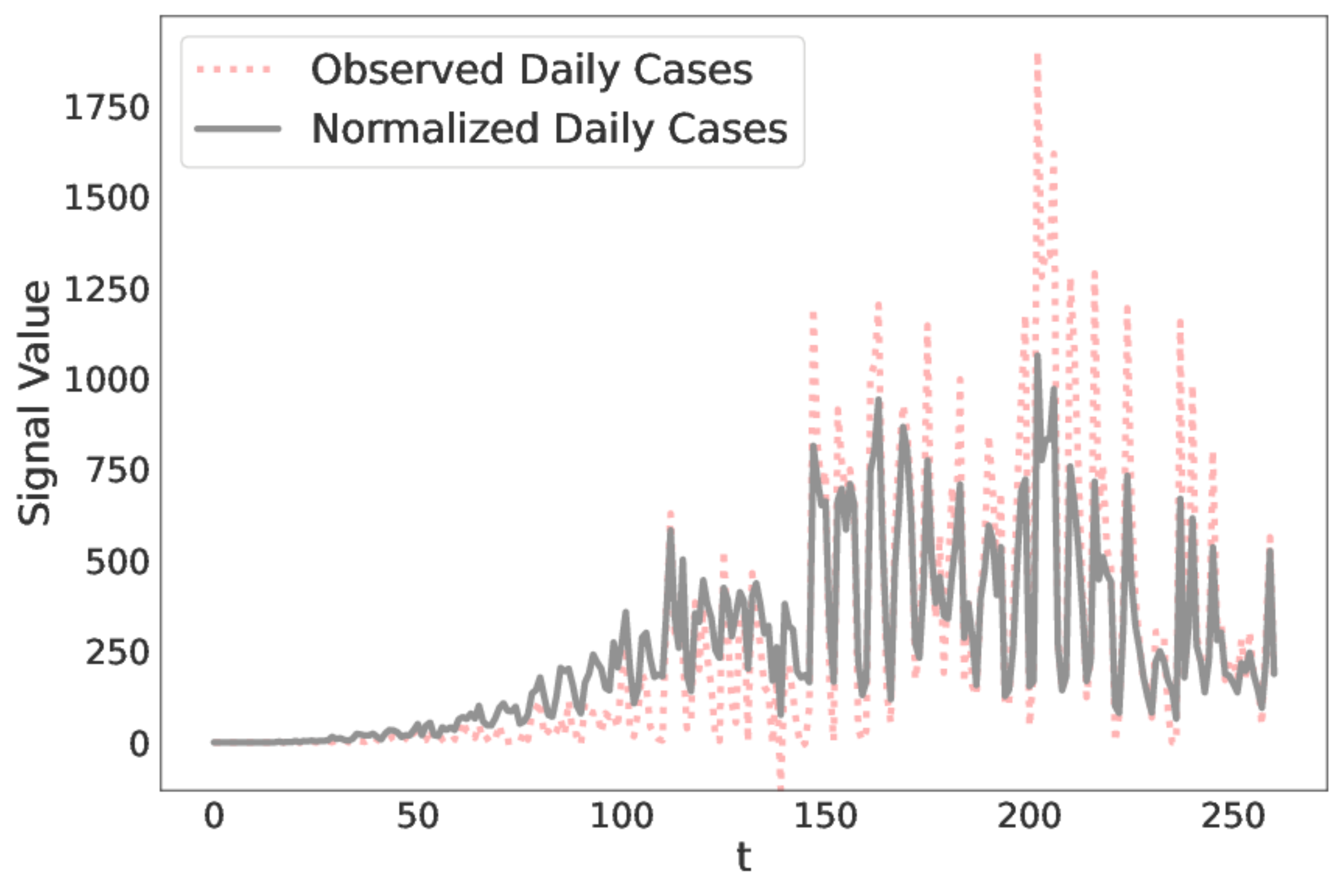}}
\caption{Observed Daily Cases versus Normalized Daily Cases}
\label{fig:normalized_signal}
\end{figure}

While the accentuator was applied to the signal that carried the daily variation in the number of cases, the low-pass filter was applied to the signal formed by the number of cases on each day, that is, the $C$ signal, where $c_i^t$ is the number of cases in city $i$ at day $t$, to generate a filtered signal $\hat{C}$. Figure \ref{fig:normalized_signal} compares an original signal and a filtered signal. It is possible to note that, in general, the signal oscillation is mitigated, ensuring a more reliable signal.

By applying both the ARIMAX and the EEMD-ARIMAX methods to the normalized data, we obtained the results shown in Table \ref{tab:resul-eemd-norm}, which are presented as in Table \ref{tab:resul-eemd}. The average RMSE for the forecasting considering only the ARIMAX method was 142.981 with a standard deviation of 122.703. The average RMSE for the EEMD-ARIMAX method was 107.664 with a standard deviation of 99.917. Therefore, EEMD-ARIMAX was 24.70\% better than ARIMAX.

\begin{table}[!htpb]
    \centering
    \caption{Results achieved by ARIMAX and EEMD-ARIMAX using normalized data.}
    \label{tab:resul-eemd-norm}
    \scriptsize{
\begin{tabular}{|c|l|c|c|c|c|c|c|}
  \hline
  \multirow{2}{*}{Region} & \multirow{2}{*}{City-Federative unit}  & \multicolumn{3}{c|}{ARIMAX} & \multicolumn{3}{c|}{EEMD-ARIMAX}\\ \cline{3-8}
  & & ME & RMSE & MAE & ME & RMSE & MAE \\ \hline
  \multirow{7}{*}{North} & Belém-PA (4,1,1) & 3.496 & 88.611 & 64.383 & 0.823 & 55.300 & 40.659\\
  & Boa Vista-RR (1,0,1) & 7.276 & 228.034 & 115.729 & -15.266 & 165.307 & 100.733\\
  & Macapá-AP (2,1,3) & 2.783 & 129.141 & 56.598 & -0.702 & 79.429 & 39.340 \\
  & Manaus-AM (4,1,1) & 0.796 & 18.994 & 12.861 & 0.005 & 12.406 & 8.771 \\
  & Palmas-TO (3,1,2) & 1.301 & 49.561 & 35.493 & 1.416 & 35.471 & 26.719 \\
  & Porto Velho-RO (2,1,3) & -0.193 & 151.779 & 87.690 & -0.024 & 131.221 & 19.299 \\
  & Rio Branco-AC (2,0,2) & 0.287 & 44.530 & 29.891 & -1.546 & 28.245 & 17.304 \\ \hline
  \multirow{9}{*}{Northeast} & Aracaju-SE (0,1,1) & 2.672 & 103.176 & 65.999 & -2.111 & 75.438 & 48.003\\
  & Fortaleza-CE (1,0,1) & 7.836 & 179.518 & 103.165 & 0.758 & 122.469 & 76.730 \\
  & João Pessoa-PB (0,1,4) & 2.929 & 83.694 & 57.921 & 0.827 & 45.635 & 32.355 \\
  & Maceió-AL (0,1,5) & 1.645 & 65.674 & 43.626 & -1.429 & 42.589 & 29.490 \\
  & Natal-RN (0,1,3) & 2.701 & 155.989 & 83.457 & -12.897 & 130.216 & 72.329 \\
  & Recife-PE (0,1,5) & 2.483 & 105.509 & 63.195 & -4.679 & 80.054 & 50.157 \\
  & Salvador-BA (4,1,1) & 3.053 & 195.621 & 121.726 & -7.470 & 155.119 & 102.551\\
  & São Luis-MA (0,1,4) & 1.747 & 42.258 & 31.661 & 1.173 & 25.417 & 18.514 \\
  & Teresina-PI (3,1,2) & 2.676 & 59.647 & 42.921 & 2.135 & 47.579 & 35.588 \\ \hline
  \multirow{4}{*}{Midwest} & Brasilia-DF (3,1,2) & 4.041 & 135.460 & 87.941 & -1.909 & 91.309 & 57.297\\
  & Campo Grande-MS (3,1,2) & 3.944 & 90.317 & 55.276 & -0.974 & 62.449 & 40.751\\
  & Cuiabá-MT (2,1,3) & 1.410 & 54.245 & 37.052 & -0.109 & 35.409 & 24.553 \\
  & Goiânia-GO (3,1,2) & 3.249 & 136.126 & 88.814 & -11.139 & 100.058 & 73.351 \\ \hline
  \multirow{4}{*}{Southeast} & Belo Horizonte-MG (3,1,2) & 4.806 & 156.051 & 109.193 & 1.246 & 114.420 & 81.809\\
  & Rio de Janeiro-RJ (0,1,5) & 8.852 & 292.669 & 192.758 & -22.889 & 230.824 & 159.742 \\
  & São Paulo-SP (0,1,5) & 12.976 & 628.305 & 420.853 & -57.823 & 497.835 & 314.477 \\
  & Vitória-ES (3,1,2) & 1.569 & 56.394 & 41.890 & 3.439 & 44.391 & 32.097 \\ \hline
  \multirow{4}{*}{South} & Curitiba-PR (2,1,3) & 2.342 & 99.194 & 61.001 & -7.094 & 80.935 & 51.362 \\
  & Florianópolis-SC (0,1,3) & 18.719 & 184.337 & 76.301 & -8.676 & 144.855 & 68.046 \\
  & Porto Alegre-RS (0,1,1) & 25.068 & 325.657 & 129.257 & -2.233 & 272.542 & 150.659 \\
  \hline
\end{tabular}
}
\end{table}

\begin{figure}[!htp]
    \centering
    \includegraphics[scale=0.8]{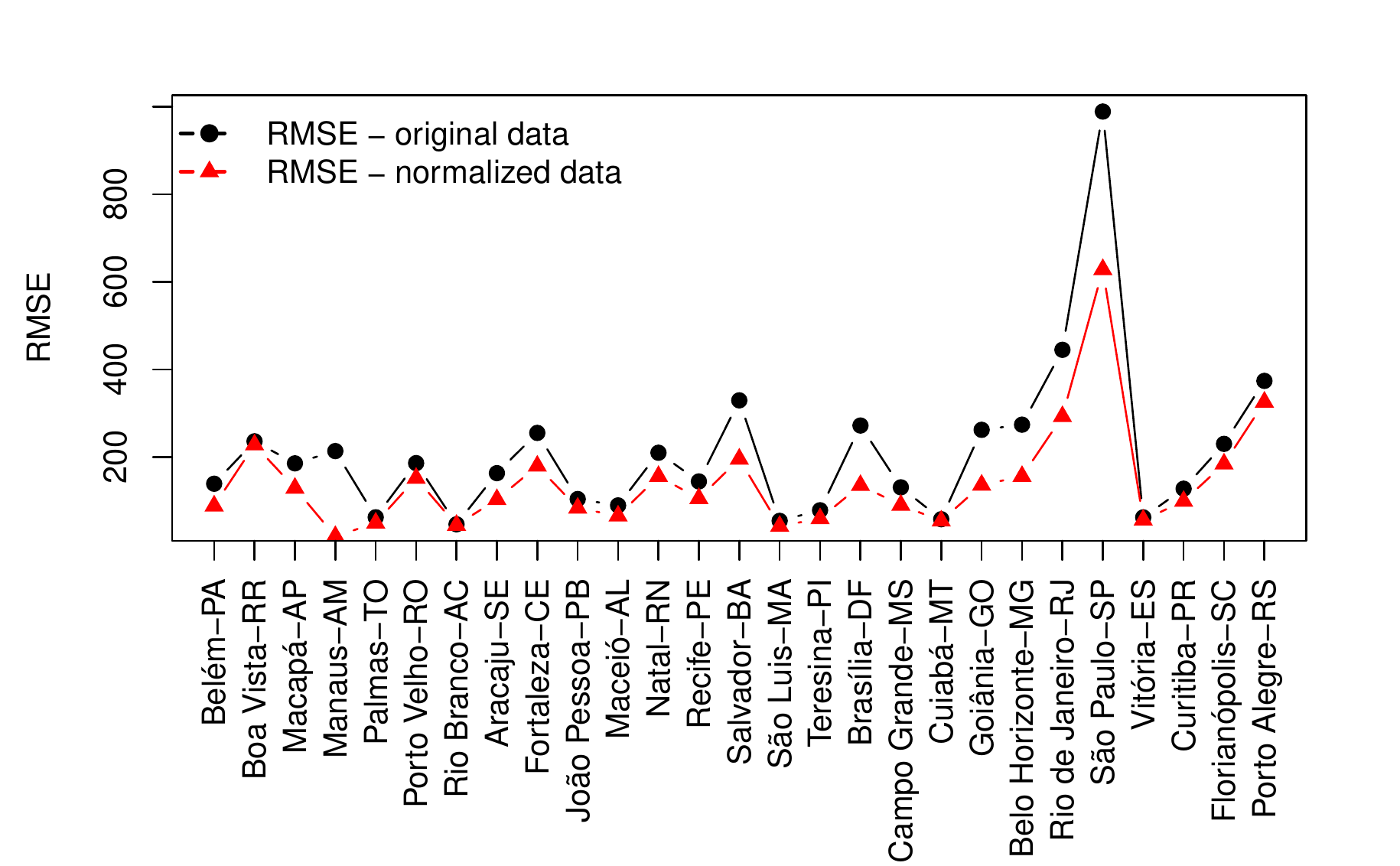}
    \caption{RMSE of original data versus RMSE of normalized data.}
    \label{fig:RMSE-compare}
\end{figure}

There was an improvement of 30.69\% in the prediction by EEMD-ARIMAX  when normalized data were used. Figure \ref{fig:RMSE-compare} shows all the RMSEs obtained by the EEMD-ARIMAX method using non-normalized (black) and normalized (red) data. 

\section{Final Remarks and Future Works} 
\label{sec:final}

As stated by \cite{Fildes2008},  contributions to forecasting are normally achieved by developing new methods that establish a connection between their effectiveness and the context they are applied to. The contributions offered by this paper meet this purpose, since it is case-oriented and we examine the system as a whole, identifying patterns in the time series as well as anomalies to draw conclusions about the correlation between meteorological and mobility variables. The novel method is an EEMD-ARIMAX hybrid, which uses an intelligent strategy to detect anomalies in data after the method has provided a forecast.

The analysis of the original data indicated that the correlation between the number of COVID-19 cases and the meteorological/human mobility variables depends on the region the Brazilian city under study is located. The prediction methods ARIMAX and EEMD-ARIMAX achieved an average square error of 211.99 and 155.33, respectively. These results indicate that the decomposition method improved the prediction of COVID-19 cases.


Because some data anomalies were observed, e.g., as very high peaks and negative numbers of cases, we proceeded with an anomaly study to normalize the data. When the ARIMAX and EEMD-ARIMAX methods were applied to the normalized data, an average quadratic error equal to 142.98 and 99.92 was found, respectively, confirming the positive effect of the data decomposition in the prediction of COVID-19 cases. Therefore, anomaly detection played a key role in effectively fitting the COVID-19 curve as it repaired the data deficiencies found in the vast majority of real-world applications. 

Future studies may involve the use of other prediction methods, including deep learning strategies. We also suggest the use of optimization algorithms, such as nature-inspired metaheuristics  \citep{kar2016bio, Abualigah2021} and the sine and cosine algorithm  \citep{mirjalili2016sca}, to either  identify approximations of the local maxima and minima or to optimize the decision on which points to interpolate in the EEMD-ARIMAX. Optimization algorithms can also be used to find the best values of $p$, $d$, and $q$ in the ARIMAX model ($p, d, q, n$) in each extracted IMF, or to determine a linear regression model as described by \cite{MAKADE2020}, which used particle swarm optimization for this task.
Another future work direction would be to determine anomalies in the extracted IMFs instead of in the original time series.


\section*{Acknowledgments}

This study was financed in part by the \textit{Coordenação de Aperfeiçoamento de Pessoal de Nível Superior - Brasil} (CAPES) - Finance Code 001 -- \textit{Bolsista} CAPES/BRASIL (Grant: 88887.507037/2020-00); \textit{Fundação de Amparo à Pesquisa do Estado de São Paulo} (FAPESP) (Proc. 17/24185-0; 13/07375-0) and \textit{Conselho Nacional de Desenvolvimento Científico e Tecnológico} (CNPq) (306036/2018-5).

The authors are also grateful to \textit{Centro de Previsão de Tempo e Estudos Climáticos} for providing the meteorological dataset and  the anonymous reviewers for their valuable comments and suggestions, which improved the quality of this paper.

\bibliographystyle{elsarticle-harv}
\bibliography{mybibfile}

{\color{black}{
\appendix
\section{List of symbols}
\label{sec:appendix}

This appendix contains a list of symbols used throughout the paper and their descriptions.
}}

\begin{table}[!htpb]
    \centering
            \caption{Part 1 of the list of symbols and notations used in this paper.}    \begin{tabular}{ll} 
    \hline
       Symbol & \multicolumn{1}{c}{Description} \\ \hline
        $X_t$, $W_t$ & time series \\
        $m$ & number of ensembles in EEMD \\
        $s$ & number of IMFs to be extracted from $X_t$ or $W_t$\\
        $k$ & variable that specifies an ensemble in a given iteration of EEMD\\
        $n$ & number of meteorological and human mobility variables\\
        $Y_1, \ldots, Y_n$ & meteorological and human mobility variables\\
        $Z_t$ & time series obtained from $W_t$\\
        $\sigma_{original}$ & standard deviation of $X_t$\\
        $\sigma_{noise}$ & standard deviation of $Z_t$\\
$\mu$ & a relatively small number which relates  $\sigma_ {noise}$ and $\sigma_{original}$\\
        $e^k_{max}$ & upper envelope of $X_t$\\
        $e^k_{min}$ & lower envelope of $X_t$\\
        $m^k_t$ & time series which correspond to average between $e^k_{max}$ and
        $e^k_{min}$\\
        $d_t^k$ & time series obtained by the operation $d_t^k = Z_t - m_t^k$\\
        IMF$^k_j$ & $j$-th IMF of ensemble $k$\\
        IMF$^j_{X_t}$ & $j$-th IMF obtained  from the time series $X_t$\\
        IMF$^j_{Y_t}$ & $j$-th IMF obtained  from the time series $Y_t$\\
        RES$_{Y_j}$ & residual values found by applying EEMD to $Y_j$\\
        $\hat{\mbox{IMF}}^{j}$ & $j$-th IMF of the estimated time series \\
        $\hat{\mbox{Res}}$ & estimated residual values\\
        $\hat{X}_t$ & time series $\hat{X}_t = \hat{\mbox{IMF}}^{1} + \ldots + \hat{\mbox{IMF}}^{s} + \hat{\mbox{Res}}$\\
        $\rho$ & Spearman correlation coefficient\\
           \hline      \end{tabular}
    \end{table}{}

        \begin{table}[!htpb]
    \centering
    \caption{Part 2 of the list of symbols and notations used in this paper.}
    \label{table:symbols}
    \begin{tabular}{ll}
    \hline
           Symbol & \multicolumn{1}{c}{Description} \\ \hline
    $p$ & number of autoregressive terms in  ARIMAX \\
$d$ & number of nonseasonal differences needed for stationarity in  ARIMAX \\
$q$ & number of lagged forecast errors in  ARIMAX \\
$\eta$ & constant of the ARIMAX\\
$\phi_i$ & $i$-th element of parameter $\phi$ in ARIMAX, for $i=1,\ldots, p$\\
$\theta_j$ & $j$-th element of parameter $\theta$ in ARIMAX, for $j=1, \ldots q$ \\
$\zeta_l$ & $l$-th element of parameter $\zeta$ in ARIMAX, for $l=1,\ldots,n$\\
$e_{t-j}$ & error terms of the ARIMAX, for $j=1,\ldots,q$\\
    $nc$& number of days in the dataset\\
     $c_i^t$ & observed number of COVID-19 cases on day $t$ in city $i$\\
        $\hat{c}_i^t$ & predicted number of COVID-19 cases on day $t$ in city $i$\\
        $e_i^t$ & error in the prediction of the number of COVID-19 cases on day $t$ in city $i$\\
        $s_i^t$ & absolute value of the difference between 1 and $\frac{c_i^t}{c_i^{(t-1)}}$\\
            $S^t$ & set $S^t=\{s_i^t, \forall i\}$ \\
        $\hat{S}^t$ & spectrum of $S^t$\\
        $\lambda_t$ & eigenvalues of graph Laplacian\\
        $R^t$ & time series obtained by applying the inverse Fourier transform in $\hat{R}^t$\\
        $r_i^t$ & $i$-th element of vector $R^t$\\ \hline
    \end{tabular}
\end{table}{}

\newpage

\section{Descriptive statistics of the data}\label{appendix-data}

This section presents a brief discussion about the data employed in the prediction analysis.

Tables \ref{tab:average-weather} and  \ref{tab:average-mobility} summarize, in terms of average values (Mean) and standard deviations (SD), the sample data over the considered period. The results were divided according to the five regions of Brazil: North, South, Midwest, Northeast and Southeast.

Considering each region, the cities with the highest average daily number of COVID-19 cases are Manaus (North region), Salvador (Northeast region), Brasília (Midwest region), São Paulo (Southeast region) and Curitiba (South region). These are the largest cities in each region,  except Curitiba \citep{ibge}.

Regarding meteorological variables, the North region had the highest average rainfall, while the Midwest region had the lowest average rainfall. The South region had the lowest averages in terms of maximum and minimum temperatures.  The Midwest region had the least average values of humidity. 

\begin{figure} 
\centering
 \subfloat[Retail and recreation variable.]{ 
    \includegraphics[scale=0.5]{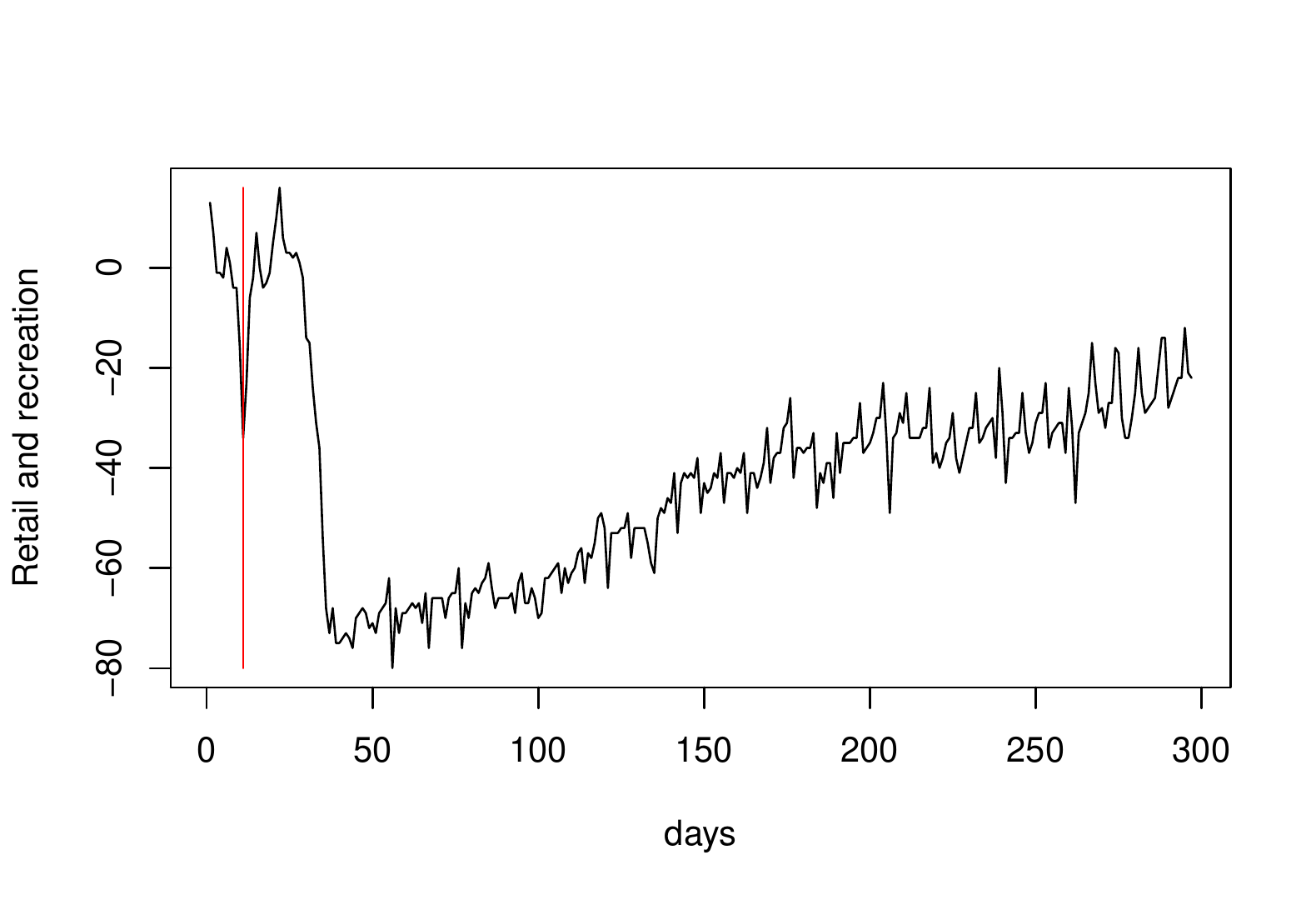}
  } 
  \subfloat[Grocery and pharmacy variable.]{ 
    \includegraphics[scale=0.5]{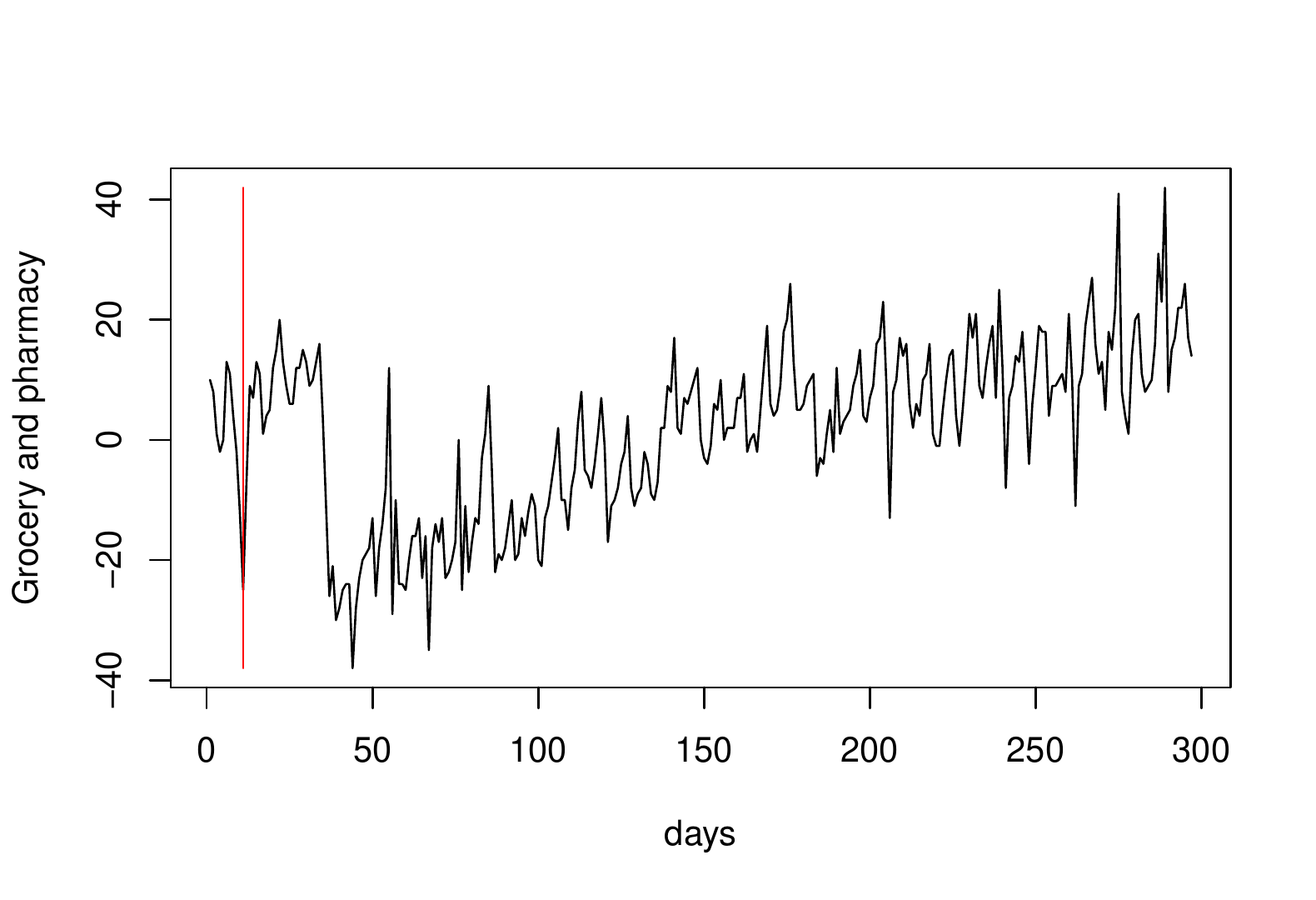}
  } 
  \\ 
  \subfloat[Parks variable.]{ 
     \includegraphics[scale=0.5]{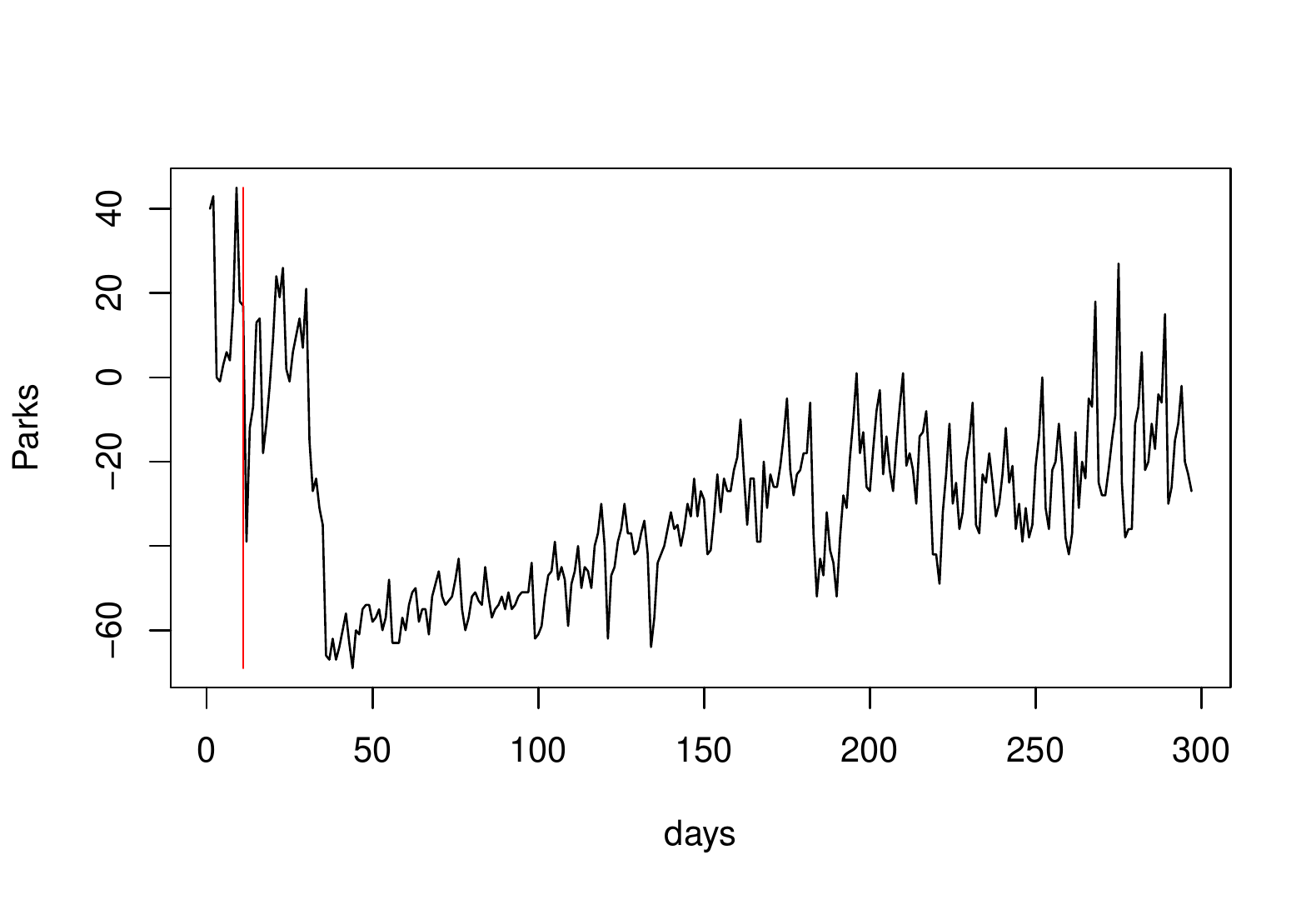}
  } 
    \subfloat[Transit station variable.]{ 
    \includegraphics[scale=0.5]{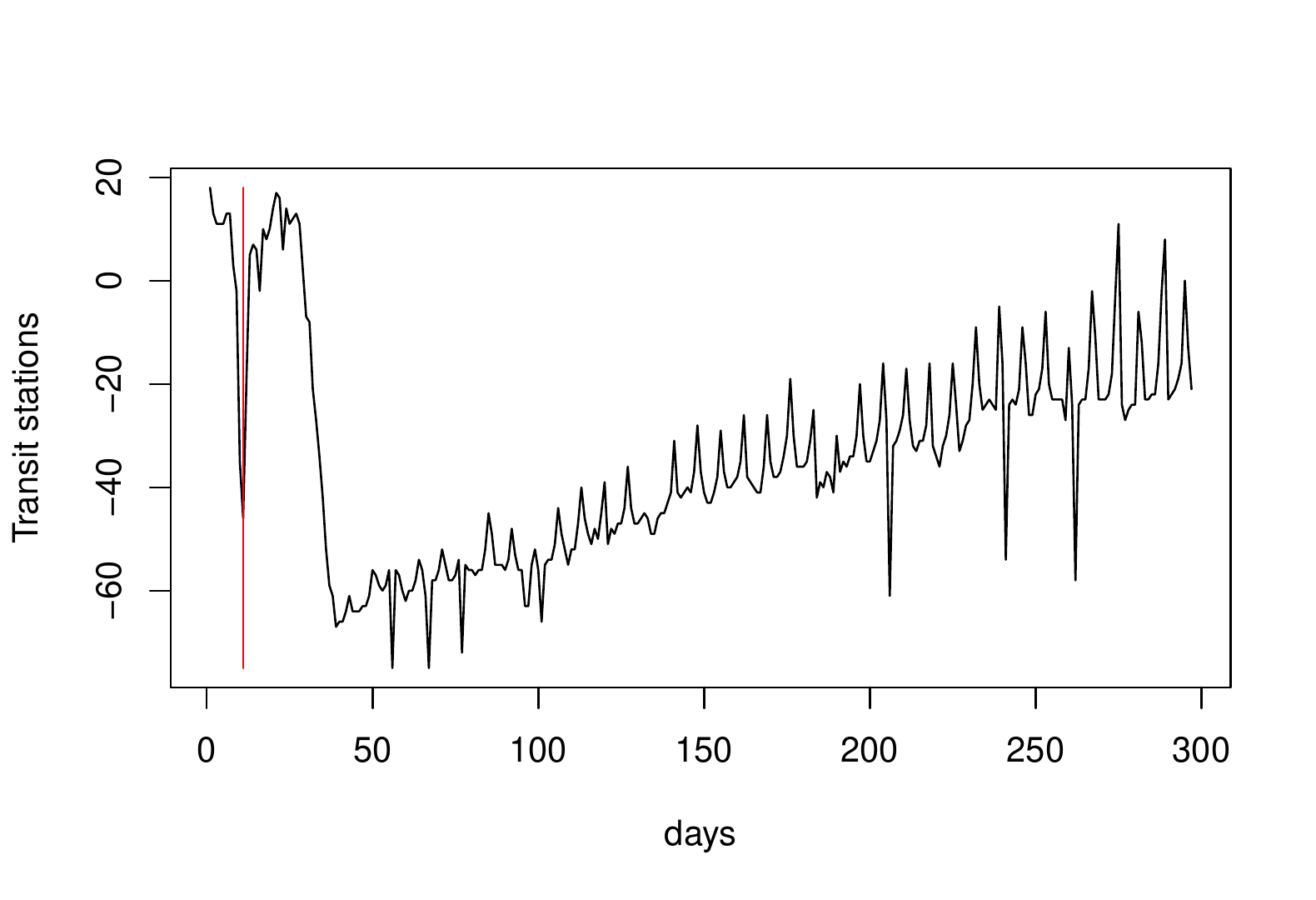}
  }
    \\ 
  \subfloat[Workplace variable.]{ 
     \includegraphics[scale=0.5]{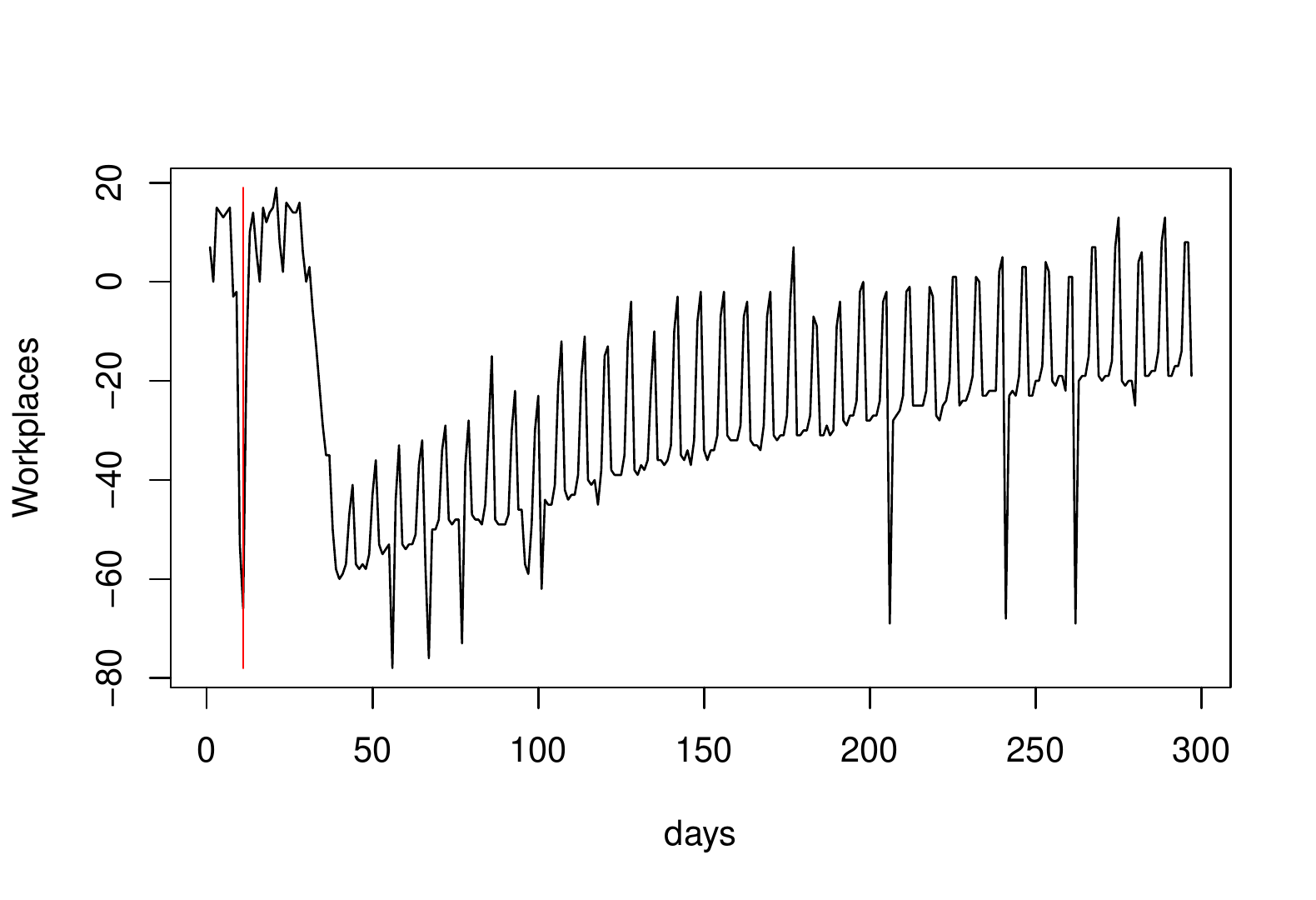}
  } 
    \subfloat[Residential variable]{ 
    \includegraphics[scale=0.5]{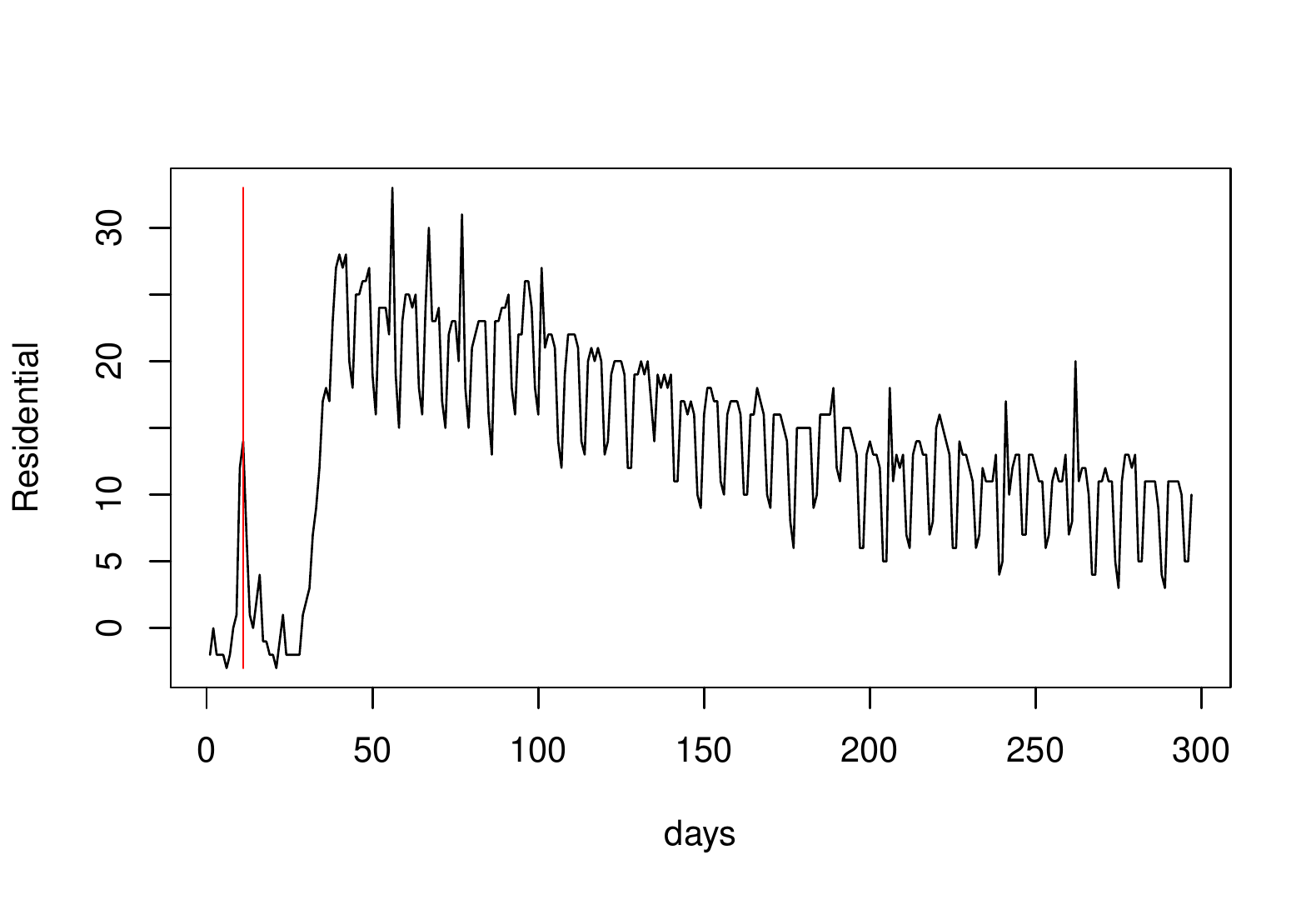}
  }
    
  \caption{Human mobility numbers in São Paulo during 297 days starting at February 15, 2020, and ending on December 2, 2020. The first recorded COVID-19 case in São Paulo was on February 25, 2020.}\label{figmobil}
\end{figure}

The behavior of the Brazilian population changed after the first confirmed COVID-19 cases. This can be attested by the human mobility data. Figure \ref{figmobil} displays the human mobility variables in São Paulo. 

\begin{figure}
    \centering
    \includegraphics[scale=0.6]{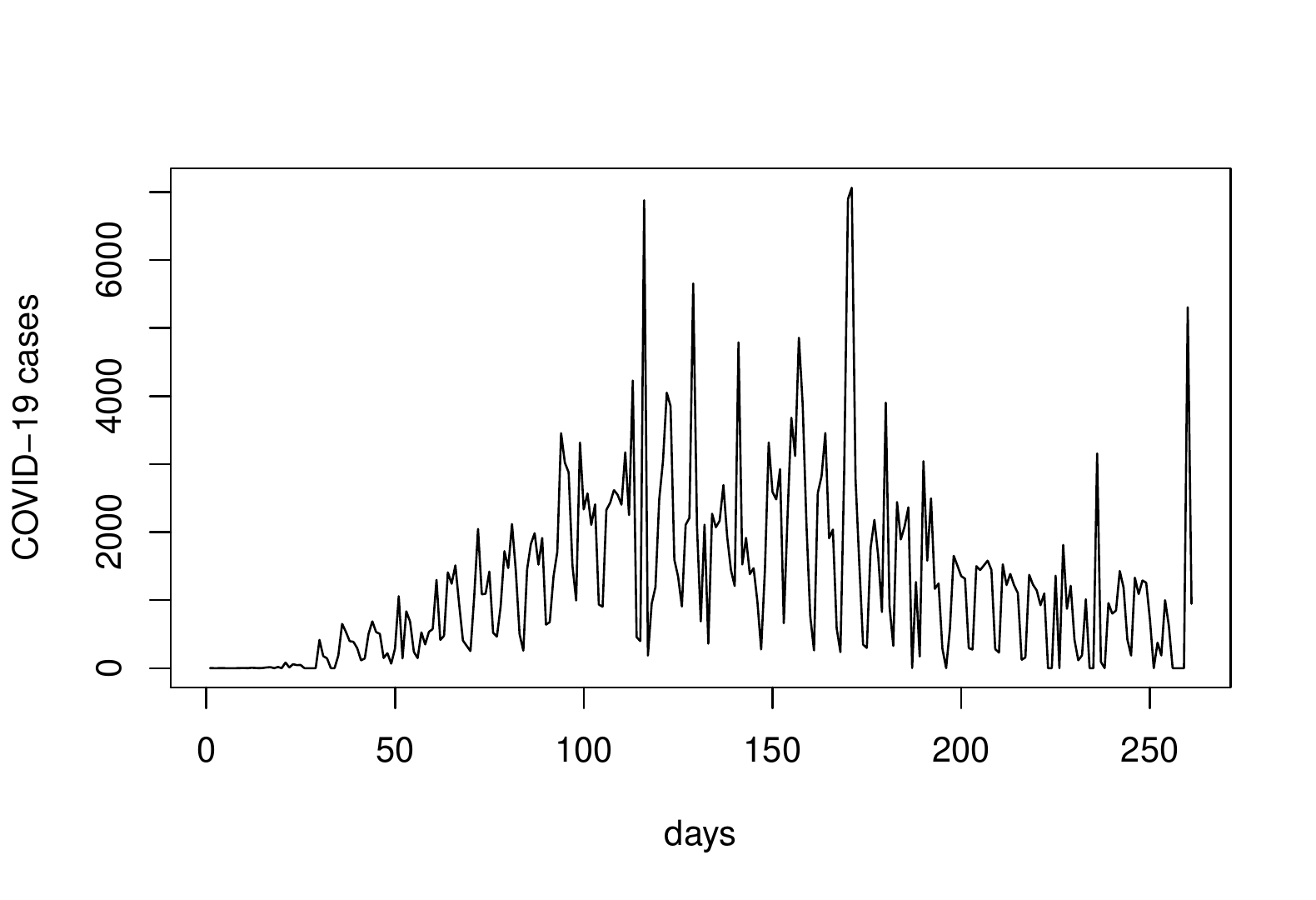}
    \caption{The number of COVID-19 cases in São Paulo in 261 days starting at February 25, 2020.}
    \label{fig:casessp}
\end{figure}

The red vertical line points to the day when the first case was found in the city. A few days after the first case was confirmed, the population started to change mobility patterns. For example, grocery consumers were visiting stores less often; the number of park visits by people has reduced;  transit stations became less busy; and so on.   But we can see that, on average, 50 days after the sudden change in human mobility patterns, the population slowly started to return to their old mobility trends. As a consequence, the number of COVID-19 cases increased, as we can see in Figure \ref{fig:casessp}.

 Table~\ref{tab:average-mobility} shows the mobility data for the 27 capitals of Brazil.  One can notice an abrupt change in the mobility behaviour within the 50 days after the first COVID-19 case was reported  in São Paulo. This information is clear by the average values of mobility trends for retail and recreational, parks, transit stations and work. Again, they are negative in relation to the baseline, and  the average values of the residential data are positive.

The mobility data for parks and transit stations related to the cities Rio Branco, Macapá, Palmas, Porto Velho, and Boa Vista were incomplete. These data comprised a limited sequence of days in the middle of the corresponding series. To overcome this limitation, we generated the missing data using the ARIMA method, with the training data being the data until the last day before the missing data sequence. The missing data are deliberate because it was not possible to guarantee anonymity, not meeting the minimum standards of quality and privacy \citep{google}.

\begin{table}[!htpb]
    \centering
      \caption{Average Daily meteorological outcomes and number of COVID-19 cases in the 27 capitals of Brazil in 2020.}
    \label{tab:average-weather}  \tiny{
\begin{tabular}{|c|l|c|c|c|c|c|c|c|c|c|c|c|}
  \hline
  \multirow{2}{*}{Region} & \multirow{2}{*}{City-Federative unit} & \multirow{2}{*}{First case} & \multicolumn{2}{c|}{Daily cases}  & \multicolumn{2}{c|}{Rain (mm)} & \multicolumn{2}{c|}{Max Temp (ºC)} & \multicolumn{2}{c|}{Min Temp (ºC)} & \multicolumn{2}{c|}{Hum (\%)}\\ \cline{4-13}
  & &  &  Mean & SD  & Mean & SD  & Mean & SD & Mean & SD  & Mean & SD \\\hline
   \multirow{7}{*}{North} & Belém-PA & Mar. 18 & 197.115 & 163.009 & 6.492 & 8.713 & 32.591 & 1.220 & 24.332 & 0.462 & 59.919 & 10.053 \\
  & Boa Vista-RR & Mar. 21 & 193.299 & 262.747 & 5.078 & 7.657 & 33.063 & 1.961 & 24.375 & 0.880 & 53.000 & 9.937 \\
  & Macapá-AP & Mar. 20 & 89.681 & 203.286 & 6.281 & 7.543 & 31.980 & 1.677 & 24.012 & 0.604 & 59.450 & 11.207 \\
  & Manaus-AM & Mar. 13 & 275.665 & 247.215 & 6.410 & 7.896 & 32.311 & 1.950 & 23.721 & 0.784 & 59.995 & 11.206 \\
  & Palmas-TO & Mar. 03 & 77.244 & 97.651 & 2.119 & 5.899 & 33.996 & 2.528 & 22.209 & 1.953 & 39.237 & 17.909 \\
  & Porto Velho-RO & Mar. 21 & 147.697 & 208.449 &  3.749 &  6.423 & 32.428 &  2.376 & 22.135 &  1.715 & 55.524 & 13.013 \\
  & Rio Branco-AC & Mar. 17 & 53.766 & 62.227 & 2.380 & 4.276 & 31.317 & 2.943 & 20.317 & 2.171 & 55.371 & 13.951 \\ \hline
  \multirow{9}{*}{Northeast} & Aracaju-SE & Mar. 14 & 164.038 & 214.724 & 1.967 & 3.317 & 29.294 & 1.824 & 22.250 & 1.475 & 28.462 & 19.018\\
  & Fortaleza-CE & Mar. 16 & 244.119 & 290.521 & 3.311 & 7.183 & 31.981 & 1.201 & 23.357 & 0.937 & 58.372 & 15.969\\
  & João Pessoa-PB & Mar. 18 & 144.842 & 140.762 & 2.779 & 5.967 & 28.802 & 1.693 & 21.486 & 1.363 & 63.376 & 10.272\\
  & Maceió-AL & Mar. 08 & 122.709 &136.231 &  5.456 &  9.884 & 29.239 &  2.082 & 21.443 &  1.566 & 62.794 &  9.238 \\
  & Natal-RN & Mar. 03 & 111.988 & 233.551 & 4.619 & 9.039 & 30.218 & 1.286 & 23.232 & 1.176 & 61.497 & 9.626 \\
  & Recife-PE & Mar. 12 & 156.083 & 173.076 & 2.260 & 4.368 & 29.335 & 1.958 & 20.974 & 1.456 & 61.335 & 10.460 \\
  & Salvador-BA & Mar. 03 & 389.050 & 428.000 & 5.575 & 8.374 & 28.175 & 1.906 & 21.435 & 1.533 & 66.525 & 9.684 \\
  & São Luis-MA & Mar. 03 & 104.190 & 76.485 & 7.966 & 9.269 & 33.299 & 1.720 & 23.829 & 0.791 & 60.032 & 15.095 \\
  & Teresina-PI & Mar. 19 & 171.391 & 138.570 & 2.627 & 7.227 & 34.424 & 2.411 & 22.657 & 1.625 & 42.158 & 18.229 \\ \hline
  \multirow{4}{*}{Midwest} & Brasília-DF & Mar. 07 & 779.633 & 657.652 & 2.375 & 5.481 & 27.984 & 2.863 & 16.345 & 2.612 & 39.216 & 16.106 \\
  & Campo Grande-MS & Mar. 14 & 160.004 & 187.732 & 1.750 & 5.499 & 30.653 & 4.819 & 17.163 & 3.834 & 35.860 & 12.122\\
  & Cuiabá-MT & Mar. 20 & 60.142 & 109.988 & 1.308 & 3.943 & 34.861 & 4.163 & 20.596 & 3.417 & 36.021 & 18.261\\
  & Goiânia-GO & Mar. 12 & 283.125 & 365.502 & 1.329 & 3.574 & 31.603 & 3.214 & 17.988 & 2.861 & 35.294 & 15.746\\ \hline
  \multirow{4}{*}{Southeast} & Belo Horizonte-MG & Mar. 16 & 210.390 & 332.629 & 1.351 & 3.995 & 25.840 & 3.350 & 12.242 & 3.817 & 45.616 & 13.695\\
  & Rio de Janeiro-RJ & Mar. 06 & 503.545 & 516.836 & 1.854 & 5.997 & 27.892 & 3.521 & 18.761 & 2.485 & 55.643 & 13.905 \\
  & São Paulo-SP & Feb. 25 & 1271.117 & 1278.987 & 2.964 & 7.599 & 25.624 & 3.750 & 15.645 & 2.769 & 57.710 & 13.350 \\
  & Vitória-ES & Mar. 19 & 92.176 & 76.349 & 2.360 & 6.893 & 26.274 & 2.906 & 17.717 & 2.126 & 44.898 & 8.728 \\ \hline
  \multirow{4}{*}{South} & Curitiba-PR & Mar. 12 & 171.408 & 209.541 & 2.245 & 5.928 & 23.207 & 4.778 & 11.358 & 3.694 & 54.833 & 17.165\\
  & Florianópolis-SC & Mar. 12 & 101.121 & 283.230 & 2.537 & 6.708 & 22.200 & 3.791 & 14.001 & 3.442 & 59.544 & 14.037\\
  & Porto Alegre-RS & Mar. 11 & 182.946 & 403.069 & 4.030 & 10.551 & 23.528 & 5.564 & 12.644 & 4.368 & 53.136 & 14.700 \\
  \hline
\end{tabular}
}

\end{table}

\begin{table}[!htpb]
    \centering
    \caption{Average Daily Mobility outcomes in the 27 capitals of Brazil in 2020.}
    \label{tab:average-mobility}
    \tiny{
\begin{tabular}{|c|l|c|c|c|c|c|c|c|c|c|c|c|c|}
  \hline
  \multirow{2}{*}{Region} & \multirow{2}{*}{City-Federative unit} & \multicolumn{2}{c|}{RR} & \multicolumn{2}{c|}{GP} & \multicolumn{2}{c|}{PA} & \multicolumn{2}{c|}{TS} & \multicolumn{2}{c|}{WO} & \multicolumn{2}{c|}{RE}\\ \cline{3-14}
  & &  Mean & SD  & Mean & SD & Mean & SD & Mean & SD  & Mean & SD & Mean & SD \\\hline

   \multirow{7}{*}{North} & Belém-PA & -35.312 & 25.832 & 4.949 & 21.322 & -13.462 & 41.875 & -32.188 & 24.972 & -21.145 & 19.478 & 12.034 & 5.534 \\
  & Boa Vista-RR & -31.186 & 21.595 & 8.182 & 16.354 & -36.738 & 24.833 & -43.054 & 21.553 & -11.242 & 16.423 & 11.390 & 3.688\\
  & Macapá-AP & -42.397 & 24.631 & 8.931 & 24.195 &-27.592 & 31.360 &-47.847 & 23.750 &-24.284 & 22.120 & 11.987 & 7.568\\
  & Manaus-AM & -23.004 & 25.252 & 16.410 & 20.614 & -24.385 & 27.292 & -9.732 & 24.354 &-11.941 & 20.574 &  9.251 &  5.030 \\
  & Palmas-TO & -35.517 & 17.736 & -2.090 & 14.930 & -41.091 & 25.342 & -45.692 & 15.887 & -17.876 & 17.140 & 12.697 &  4.885\\
  & Porto Velho-RO & -29.745 & 20.042 &  6.087 & 15.260 &-29.370 & 22.445 &-46.111 & 19.859 &-14.788 & 15.953 & 12.303 &  3.601 \\
  & Rio Branco-AC & -35.617 & 21.537 & 6.472 & 15.572 &-33.934 & 14.490 &-28.929 & 63.994 &-14.957 & 16.856 & 11.804 &  3.668 \\ \hline
  \multirow{9}{*}{Northeast} &  Aracaju-SE & -46.437 & 18.266 & -6.336 & 17.536 & -56.685 & 20.277 & -54.538 & 19.647 & -30.080 & 17.081 & 15.000 & 4.343\\
  & Fortaleza-CE & -45.432 & 24.277 & -4.360 & 17.288 & -62.331 & 19.439 & -36.432 & 23.416 & -29.174 & 20.000 & 14.182 & 6.150\\
  & João Pessoa-PB & -51.675 & 22.633 & -5.308 & 17.577 &-57.333 & 24.585 &-62.936 & 27.449 &-26.303 & 18.695 & 15.192 &  5.459\\
  & Maceió-AL & -42.037 & 21.166 & -7.828 & 16.143 & -38.832 & 21.254 &-31.180 & 18.528 &-23.820 & 17.033 & 12.307 & 5.150 \\
  & Natal-RN & -48.429 & 20.224 & -9.037 & 17.077 & -50.954 & 20.587 & -27.471 & 24.412 & -25.692 & 16.102 & 14.550 & 4.710 \\
  & Recife-PE & -48.929 & 21.464 & -3.467 & 21.369 & -44.417 & 19.026 & -31.562 & 24.025 & -28.167 & 19.216 & 14.371 & 5.480 \\
  & Salvador-BA & -54.418 & 16.492 & -5.414 & 19.258 & -59.180 & 14.703 & -41.184 & 17.735 & -30.661 & 16.285 & 17.607 & 5.021 \\
  & São Luis-MA & -31.474 & 27.618 & 11.358 & 23.003 & -21.694 & 29.205 & -40.272 & 21.138 & -20.672 & 20.521 & 12.263 & 5.741 \\
  & Teresina-PI & -54.567 & 19.814 & -12.811 & 27.881 & -38.712 & 17.869 & -62.210 & 15.384 & -30.399 & 17.221 & 18.502 & 5.268 \\ \hline
  \multirow{4}{*}{Midwest} & Brasília-DF & -36.727 & 17.673 & 7.796 & 13.966 & -30.976 & 23.291 & -26.543 & 16.725 & -22.788 & 18.711 & 14.759 & 5.558\\
  & Campo Grande-MS & -27.508 & 16.548 & 4.185 & 13.095 & -29.798 & 17.011 & -32.571 & 14.316 & -10.849 & 15.021 & 10.853 & 3.451\\
  & Cuiabá-MT & -41.134 & 18.657 & -0.522 & 12.057 & -53.534 & 12.493 & -29.556 & 17.200 & -19.806 & 16.210 & 13.207 & 4.201\\
  & Goiânia-GO & -39.908 & 17.393 & -0.362 & 13.111 & -26.267 & 17.847 & -26.750 & 14.500 & -20.613 & 16.128 & 13.562 & 5.294 \\ \hline
  \multirow{4}{*}{Southeast} & Belo Horizonte-MG & -47.415 & 14.166 & 4.148 & 14.837 & -44.076 & 14.669 & -29.508 & 14.944 & -25.538 & 16.407 & 14.521 & 4.824\\
  & Rio de Janeiro-RJ & -45.077 & 19.047 & -4.496 & 14.365 & -48.171 & 23.398 & -40.508 & 18.095 & -27.805 & 18.166 & 13.780 & 6.015 \\
  & São Paulo-SP & -44.402 & 19.667 & -0.363 & 13.463 & -33.719 & 19.750 & -37.387 & 19.178 & -27.207 & 20.239 & 14.891 & 7.061 \\
  & Vitória-ES & -52.575 & 14.905 & -5.888 & 17.864 & -53.996 & 14.497 & -52.712 & 15.100 & -28.112 & 15.954 & 15.914 & 4.505 \\ \hline
  \multirow{4}{*}{South} & Curitiba-PR & -41.554 & 16.657 & -5.158 & 17.210 & -28.533 & 16.907 & -34.696 & 15.779 & -24.579 & 16.085 & 14.421 & 5.087\\
  & Florianópolis-SC & -57.192 & 13.327 & -31.829 & 13.217 & -73.671 & 11.086 & -71.725 & 16.036 & -33.600 & 15.449 & 17.012 & 5.506\\
  & Porto Alegre-RS & -49.398 & 15.710 & -5.241 & 13.805 & -38.000 & 18.273 & -41.813 & 14.907 & -29.606 & 16.680 & 17.739 & 5.349 \\
  \hline
\end{tabular}
}
\end{table}


\section{Spearman correlation coefficient and metrics}\label{appendix-spearman}

We used the Spearman correlation coefficient to find the strength of the pairwise relationship between the data variables. Spearman is a well-known nonparametric measure to assess the rank correlation between a pair of variables by a monotonic function.   COVID-19 number of cases per day is the dependent variable, whereas the meteorological and mobility information is modeled as independent variables. The generalized expression for the Spearman rank correlation is given by Equation \eqref{eq:spearman}.

\begin{equation}
    \rho = 1- \displaystyle\frac{6}{n_s^3-n_s} \displaystyle\sum_{i=1}^{n_s} D_i^2, \label{eq:spearman}
\end{equation}

\noindent where $\rho$ is the rank correlation, $D_i$ is the pairwise difference between the ranks of the samples, and $n_s$ is the number of samples. Our interpretation to the Spearman correlation coefficient absolute value considers 0.0 to 0.3 a  negligible correlation between the variables; 0.3 to 0.5 is a low correlation;  0.5 to 0.7 is a moderate correlation; 0.7 to 0.9 is considered a high correlation; and 0.9 to 1 indicates a very high correlation. If $\rho$ is positive, then the variables are directly proportional,  otherwise, they are inversely proportional. 

To evaluate the performance of fitting models we used widely employed measures known as Mean Error (ME), Root Mean Square Error (RMSE), and Mean Absolute Error (MAE). Their respective formulas are given by Equations \eqref{eq:me}, \eqref{eq:rmse}, and \eqref{eq:mae}:

\begin{equation}
    \mbox{ME} = \displaystyle\frac{1}{n_s} \displaystyle\sum_{i=1}^{n_s} (predict_i - observed_i), \label{eq:me}
\end{equation}

\begin{equation}
    \mbox{RMSE} = \sqrt{\displaystyle\frac{1}{n_s} \displaystyle\sum_{i=1}^{n_s} (predict_i - observed_i)^2}, \label{eq:rmse}
\end{equation}

\begin{equation}
    \mbox{MAE} = \displaystyle\frac{1}{n_s} \displaystyle\sum_{i=1}^{n_s} |predict_i - observed_i|, \label{eq:mae}
\end{equation}

\noindent where $predict_i$ are the predicted values for the model, and $observed_i$ are the original values of the time series, for all $i=1,\ldots,n_s$.

\section{Prediction analysis} \label{appendix-graphics}

This section shows an illustrative analysis of the results achieved by EEMD-ARIMAX on the data without the normalization by the anomaly detection strategy.

\begin{figure} 
  \subfloat[Belém-PA]{ 
    \includegraphics[trim = 0mm 0mm 0mm 14mm,clip,scale=0.45]{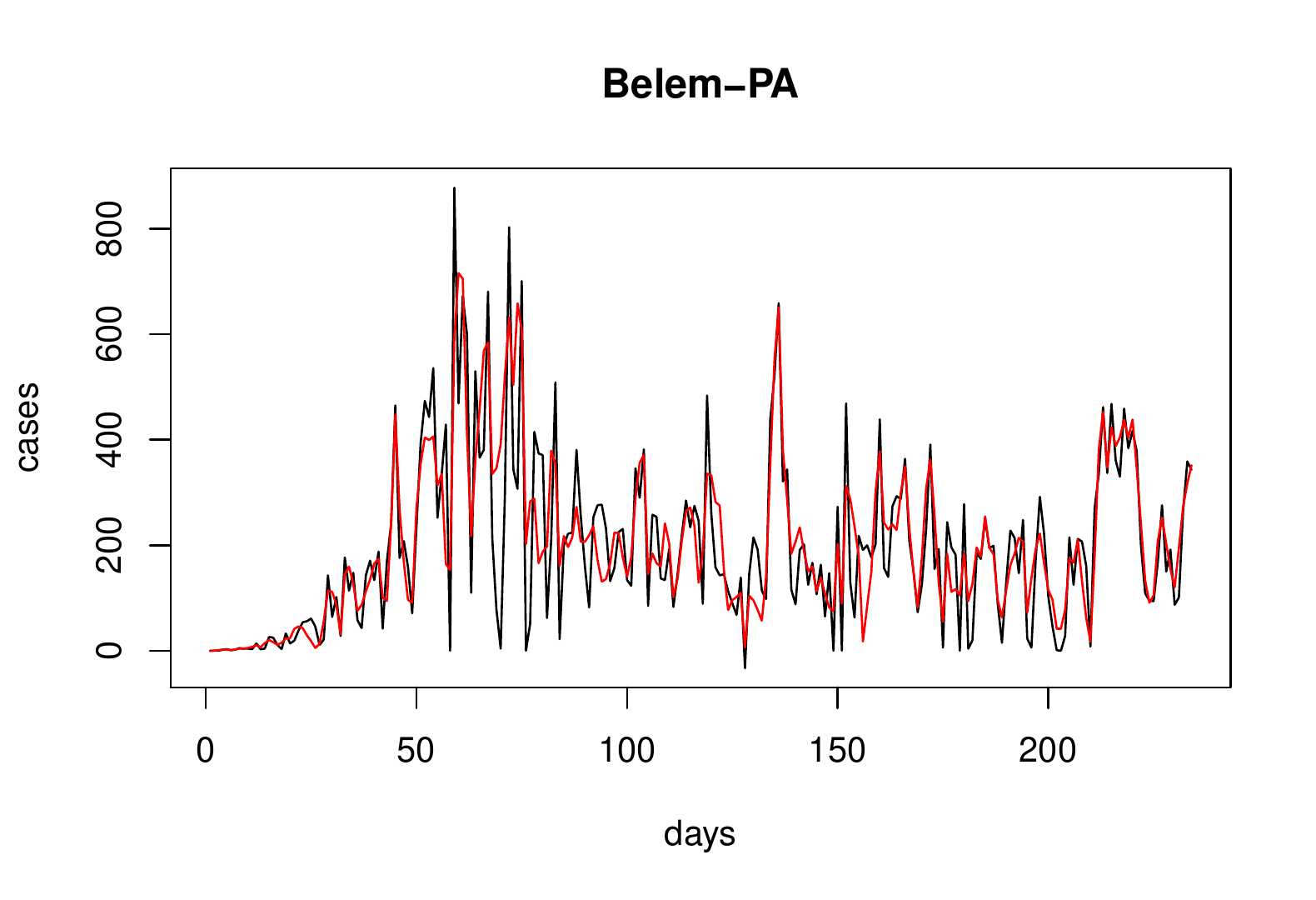}
  } 
  \subfloat[Boa Vista-RR]{ 
    \includegraphics[trim = 0mm 0mm 0mm 14mm,clip,scale=0.45]{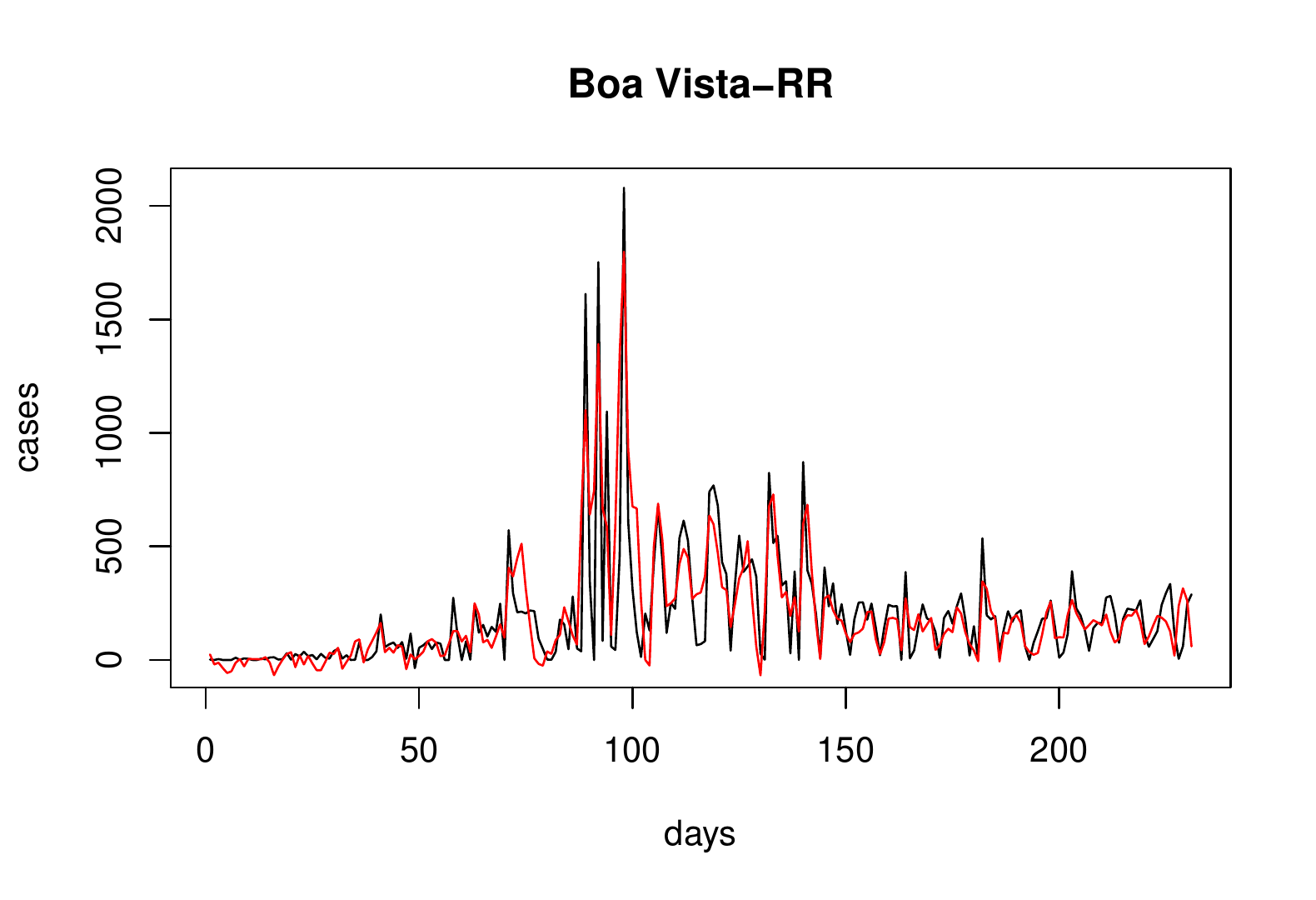}
  } 
  \\ 
  \subfloat[Macapá-AP]{ 
     \includegraphics[trim = 0mm 0mm 0mm 14mm,clip,scale=0.45]{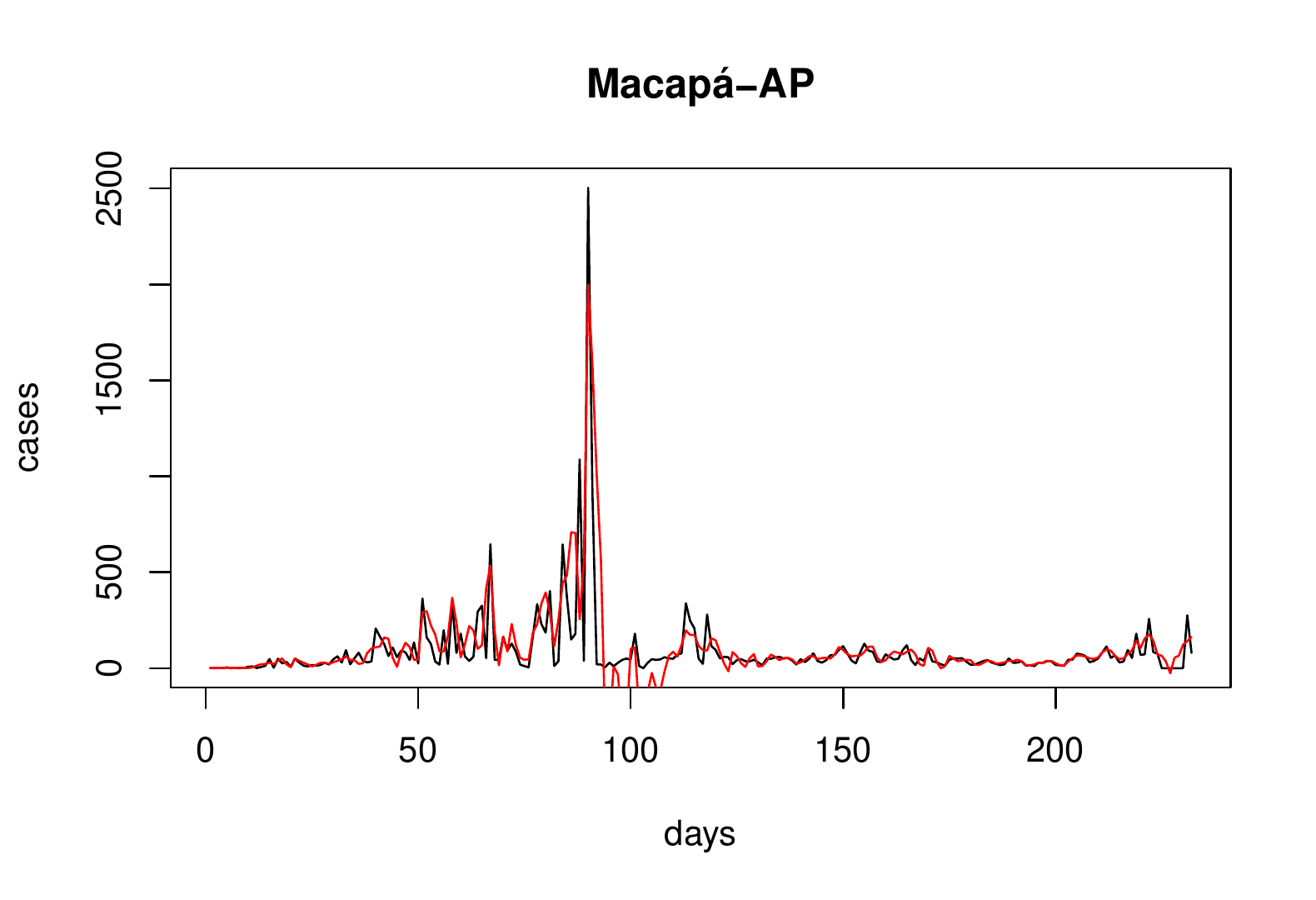}
  } 
    \subfloat[Manaus-AM]{ 
    \includegraphics[trim = 0mm 0mm 0mm 14mm,clip,scale=0.45]{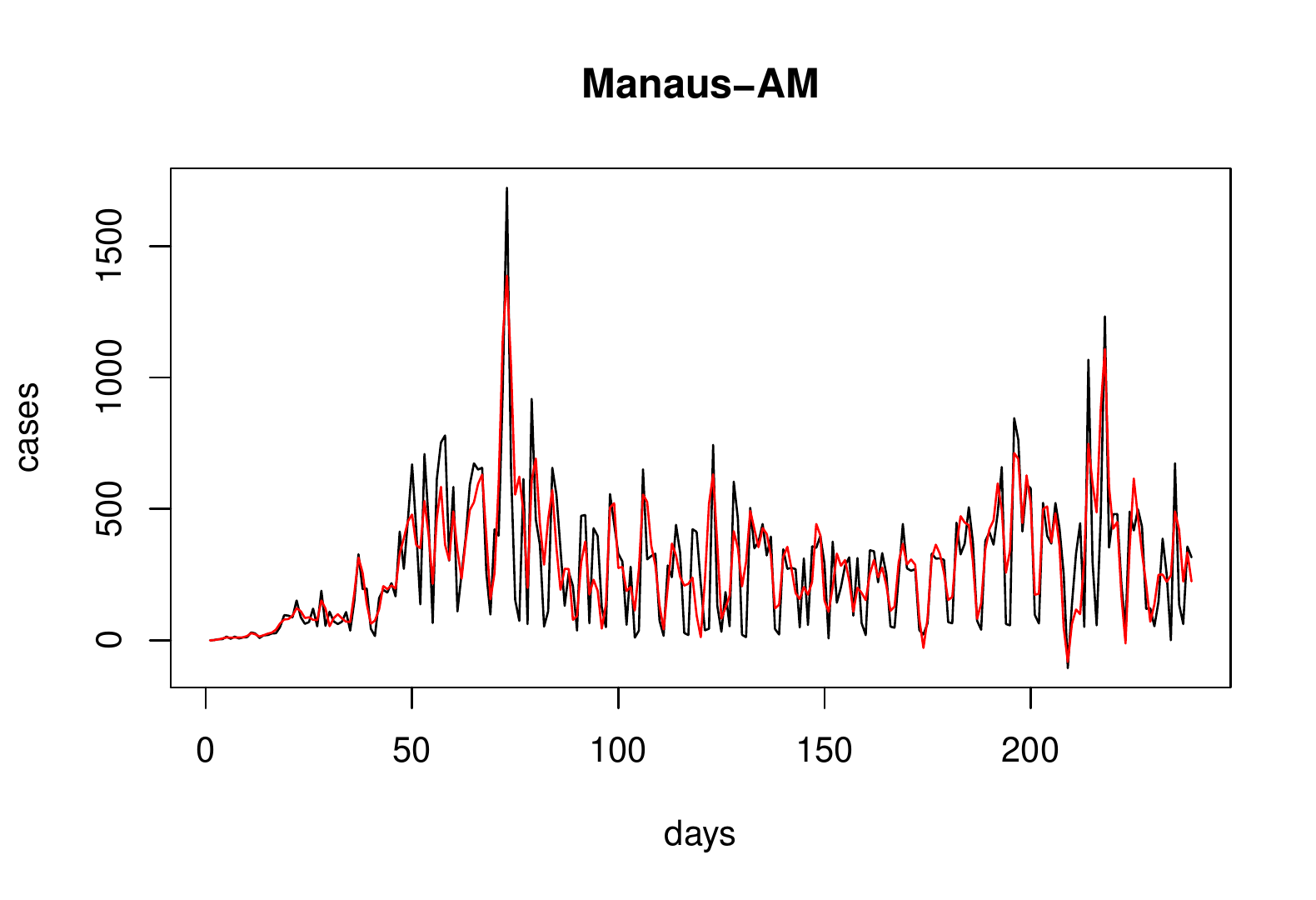}
  }
    \\ 
  \subfloat[Palmas-TO]{ 
     \includegraphics[trim = 0mm 0mm 0mm 14mm,clip,scale=0.45]{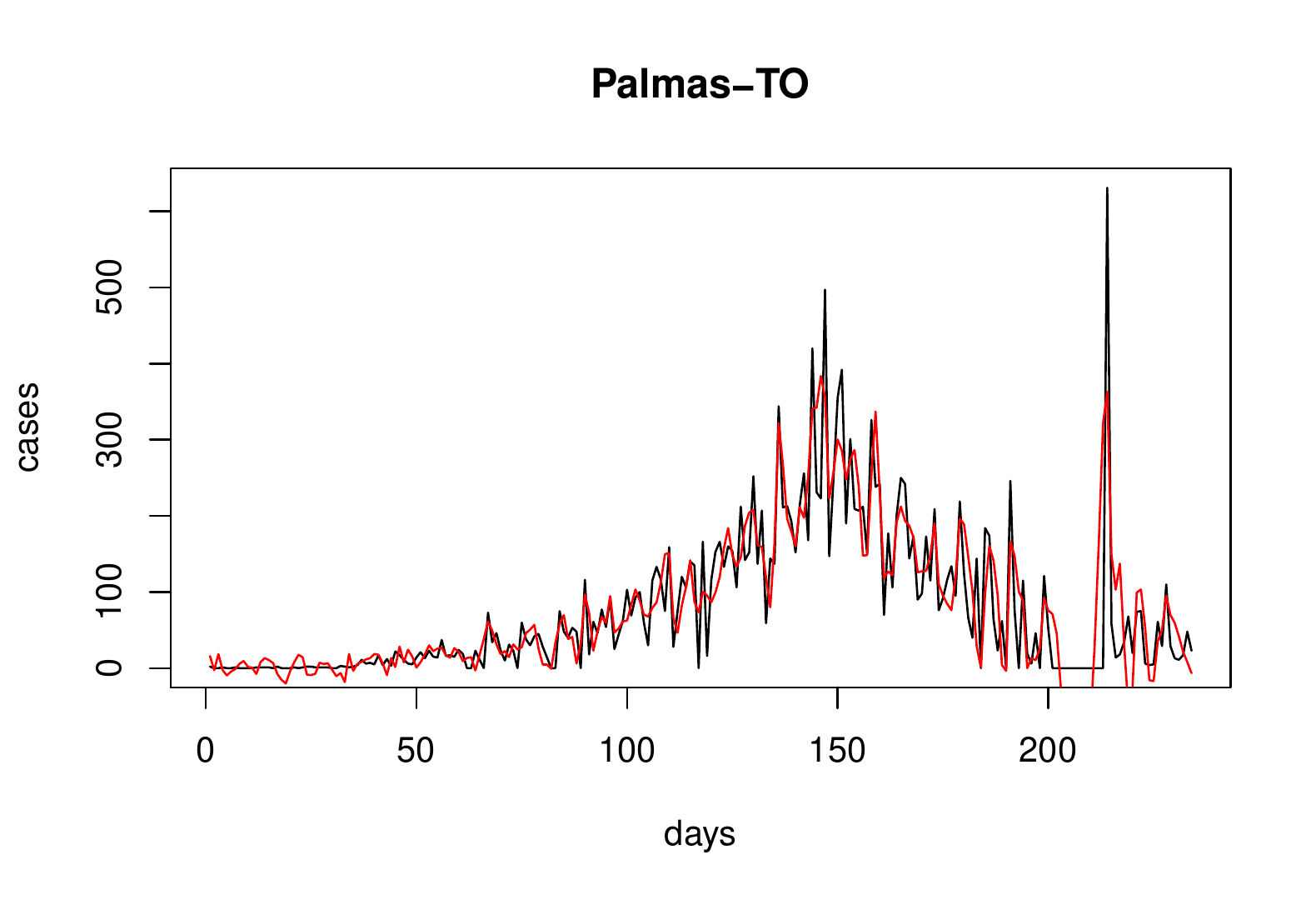}
  } 
    \subfloat[Porto Velho-RO]{ 
    \includegraphics[trim = 0mm 0mm 0mm 14mm,clip,scale=0.45]{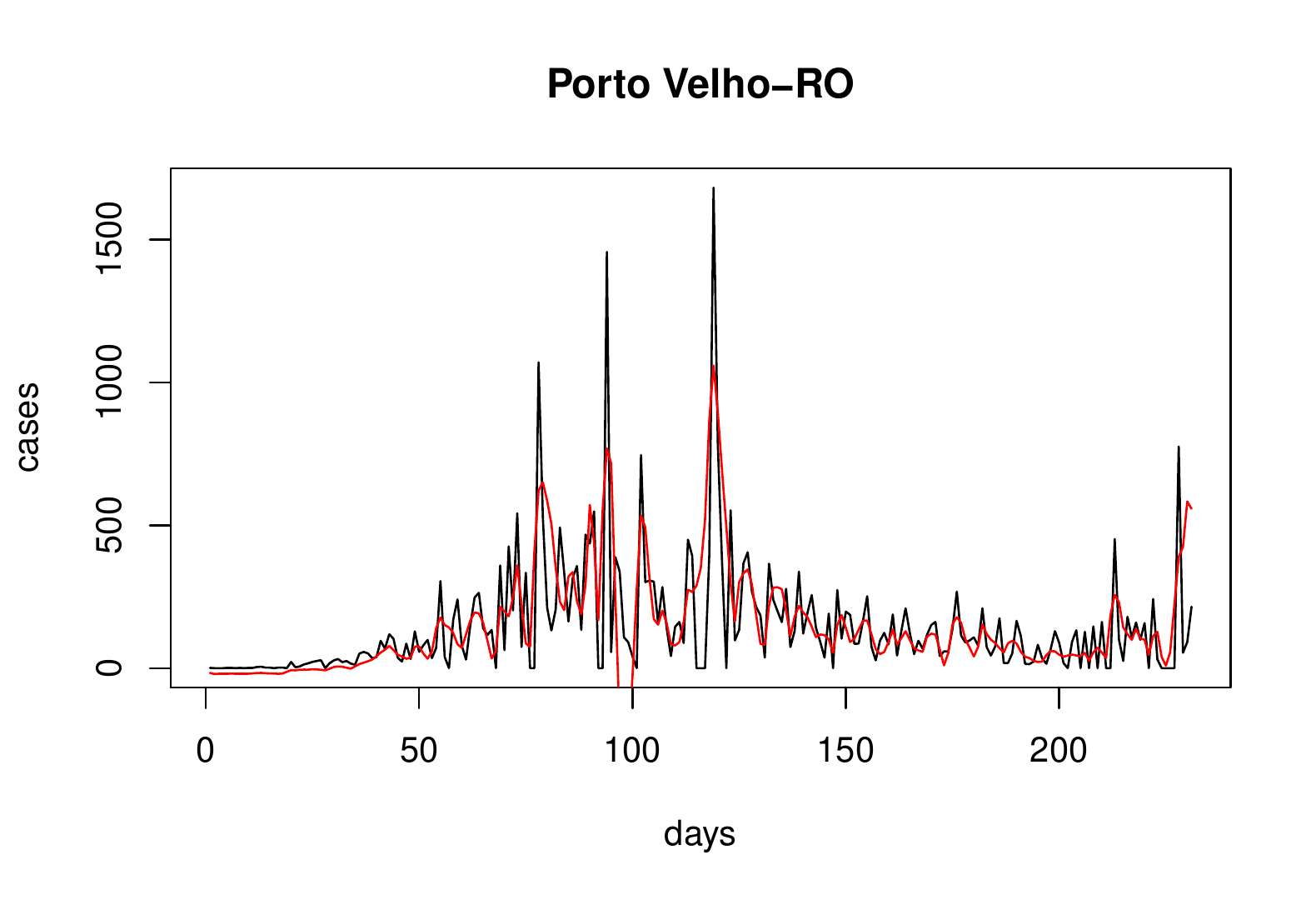}
  }
  \caption{Predicted (red) and observed (black) models to the North region of Brazil. } \label{fig1}
\end{figure} 

\begin{figure} 
   \subfloat[Rio Branco-AC]{ 
     \includegraphics[trim = 0mm 0mm 0mm 14mm,clip,scale=0.45]{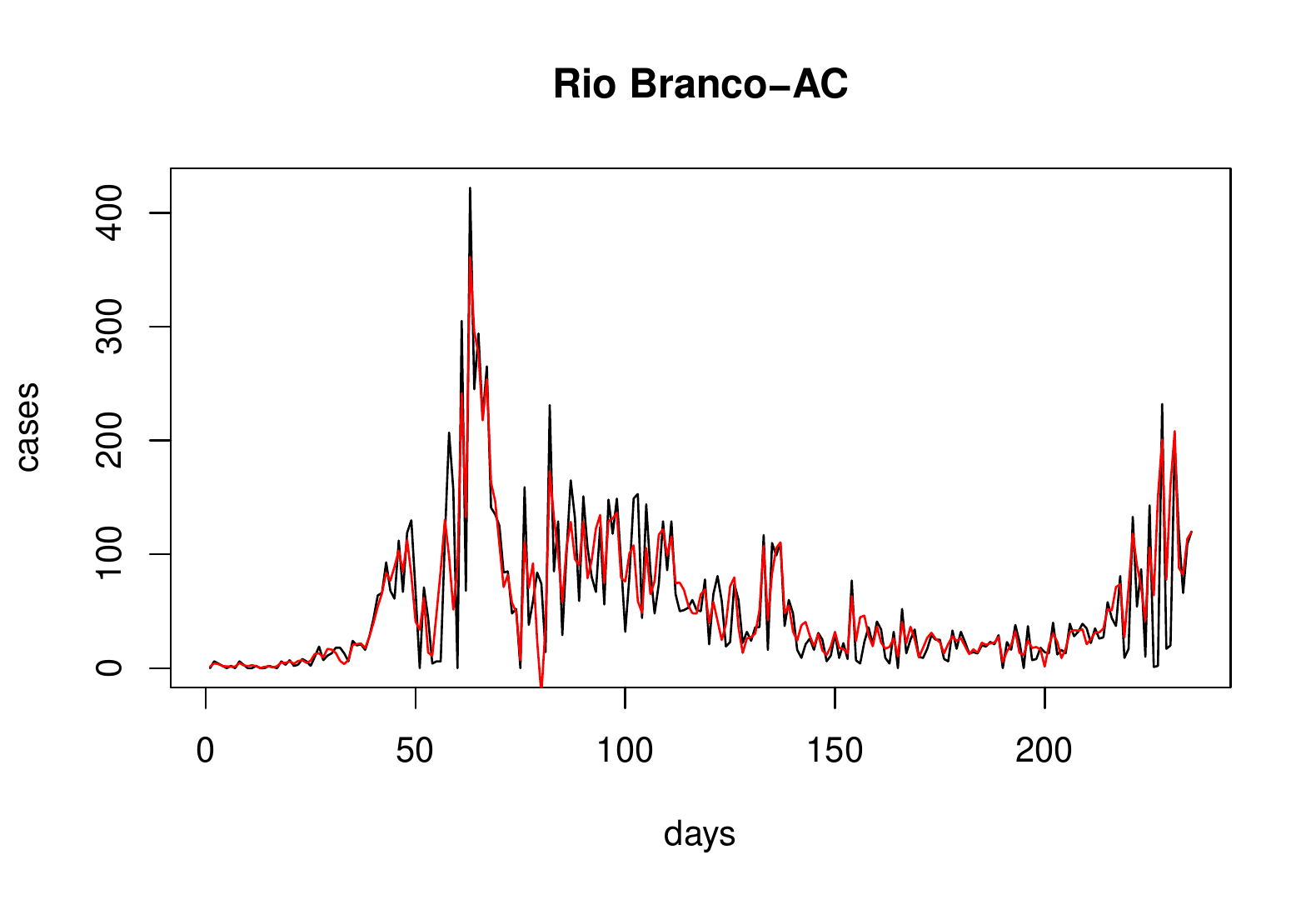}
  }
  \caption{Predicted (red) and observed (black) models to the North region of Brazil (continuation). } \label{fig1b}
\end{figure}

\begin{figure} 
  \subfloat[Aracaju-SE]{ 
    \includegraphics[trim = 0mm 0mm 0mm 14mm,clip,scale=0.45]{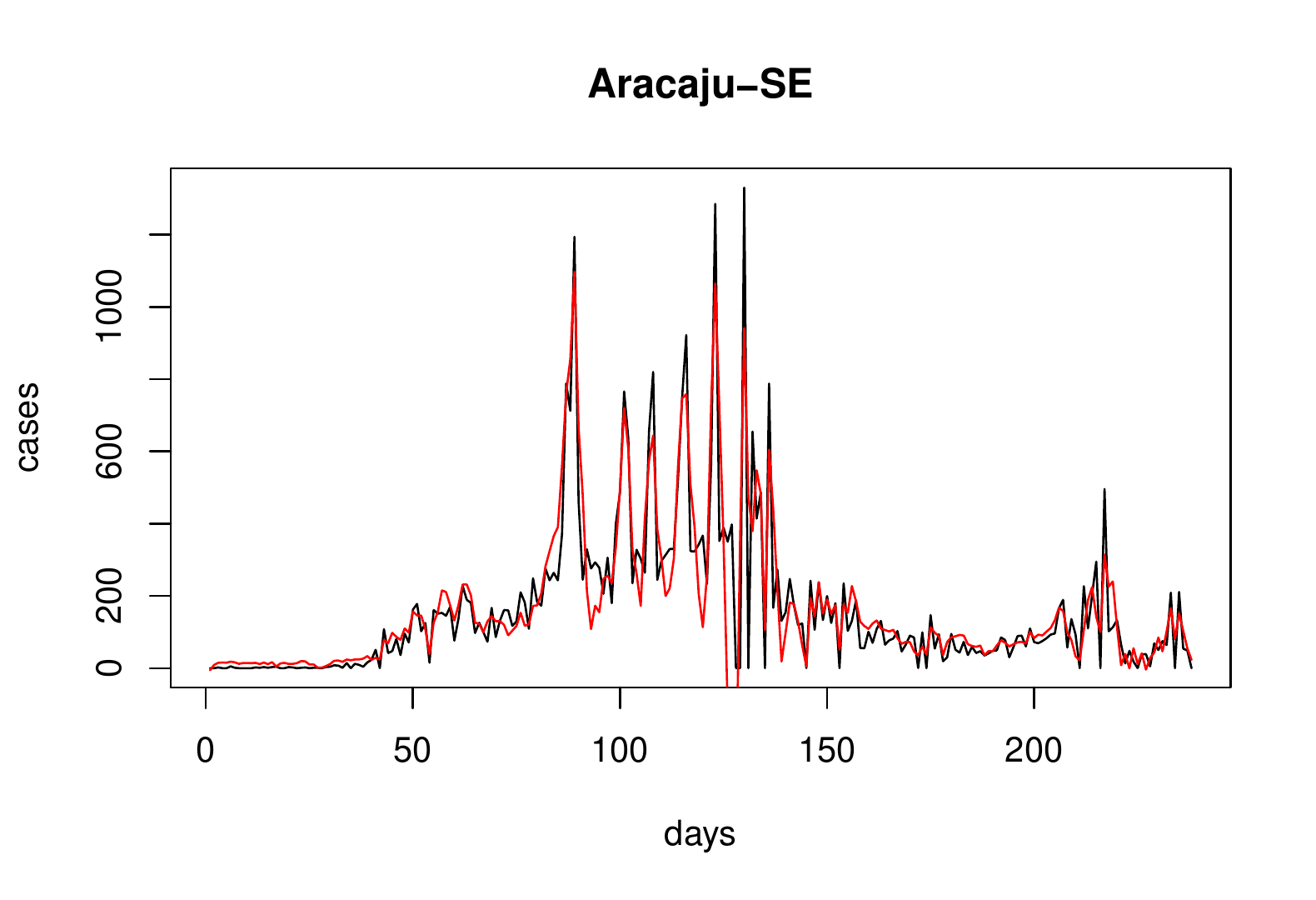}
  } 
  \subfloat[Fortaleza-CE]{ 
    \includegraphics[trim = 0mm 0mm 0mm 14mm,clip,scale=0.45]{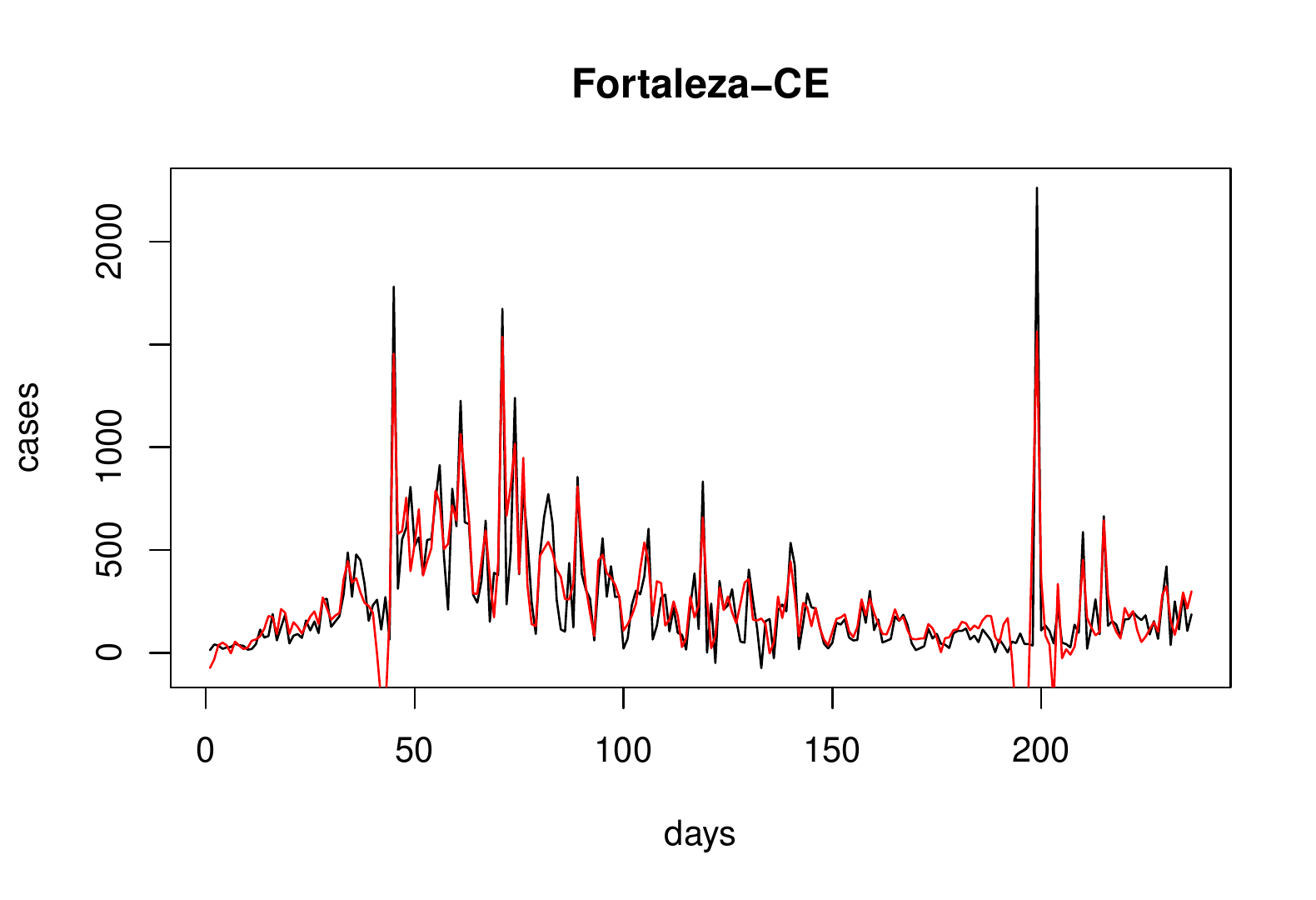}
  } 
  \\ 
  \subfloat[João Pessoa-PB]{ 
     \includegraphics[trim = 0mm 0mm 0mm 14mm,clip,scale=0.45]{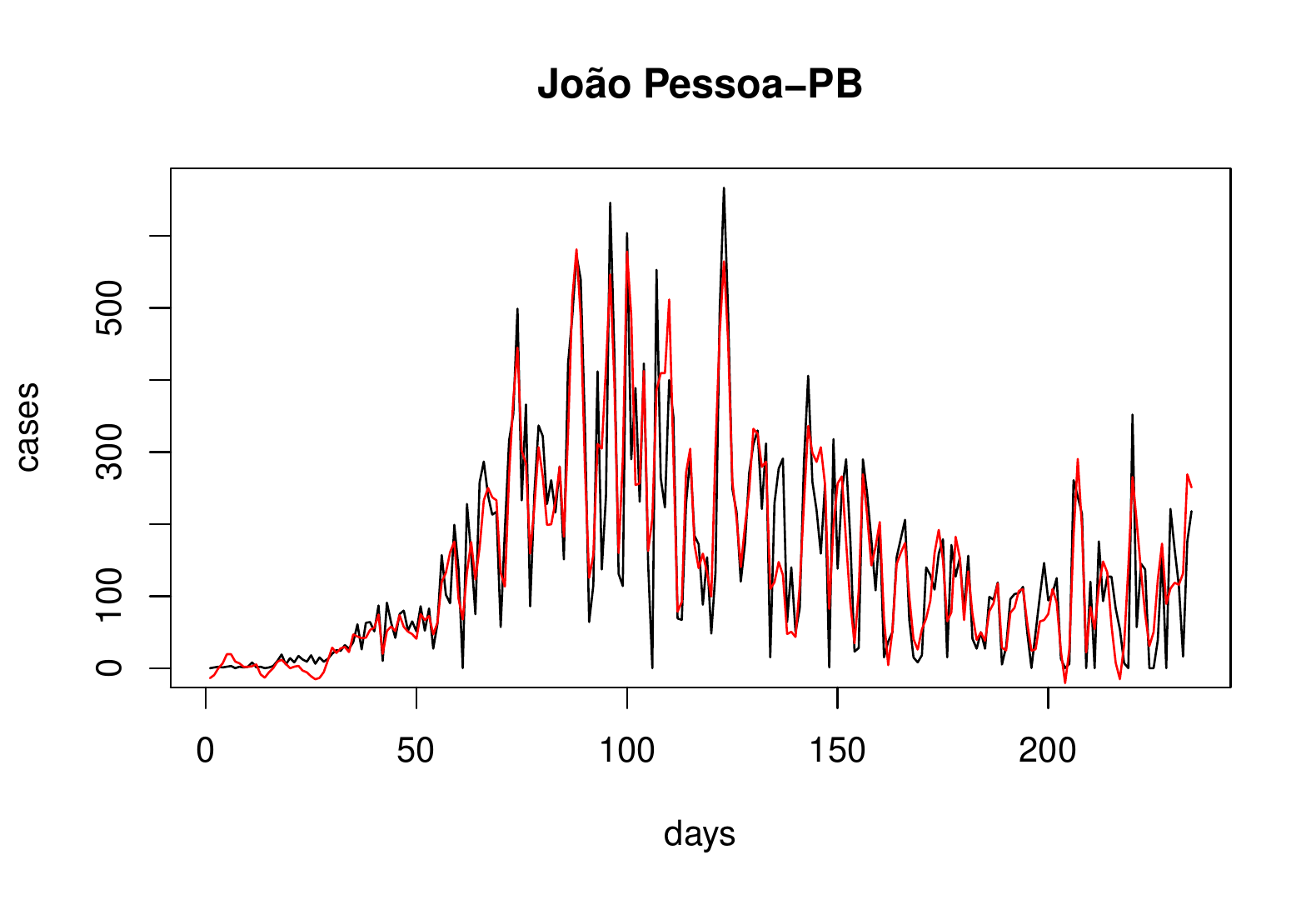}
  } 
    \subfloat[Maceió-AL]{ 
    \includegraphics[trim = 0mm 0mm 0mm 14mm,clip,scale=0.45]{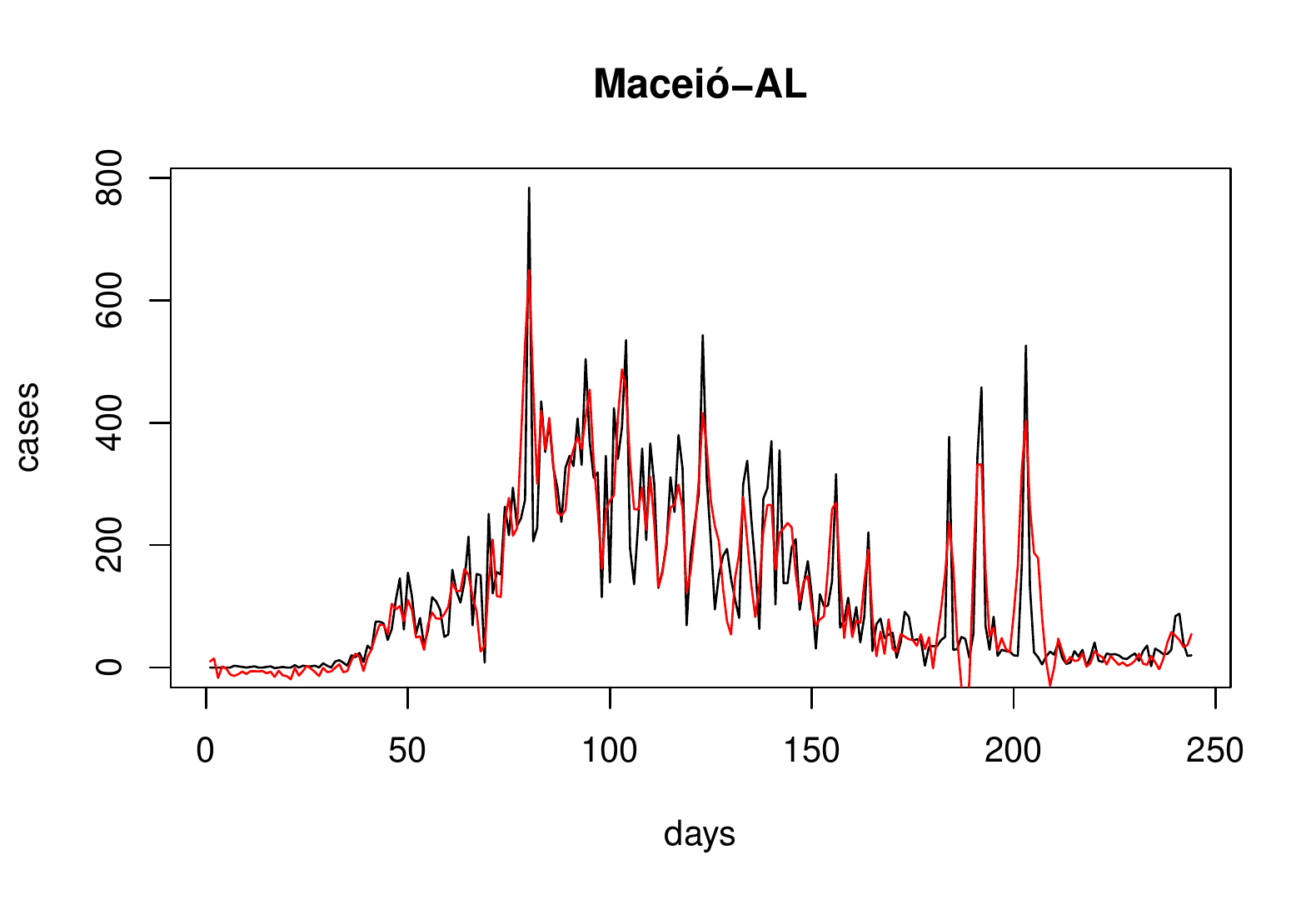}
  }
    \\ 
  \subfloat[Natal-RN]{ 
     \includegraphics[trim = 0mm 0mm 0mm 14mm,clip,scale=0.45]{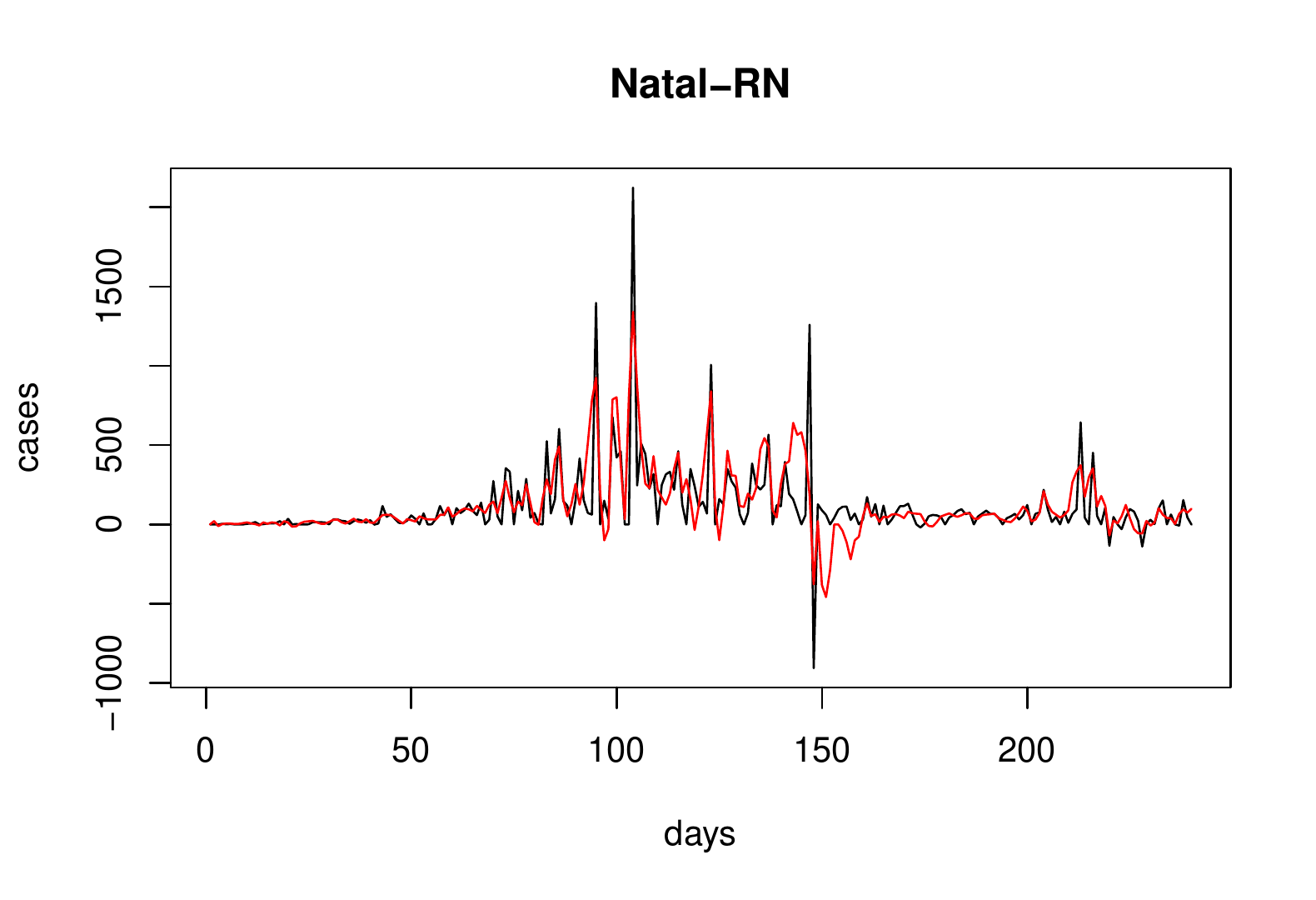}
  } 
    \subfloat[Recife-PE]{ 
    \includegraphics[trim = 0mm 0mm 0mm 14mm,clip,scale=0.45]{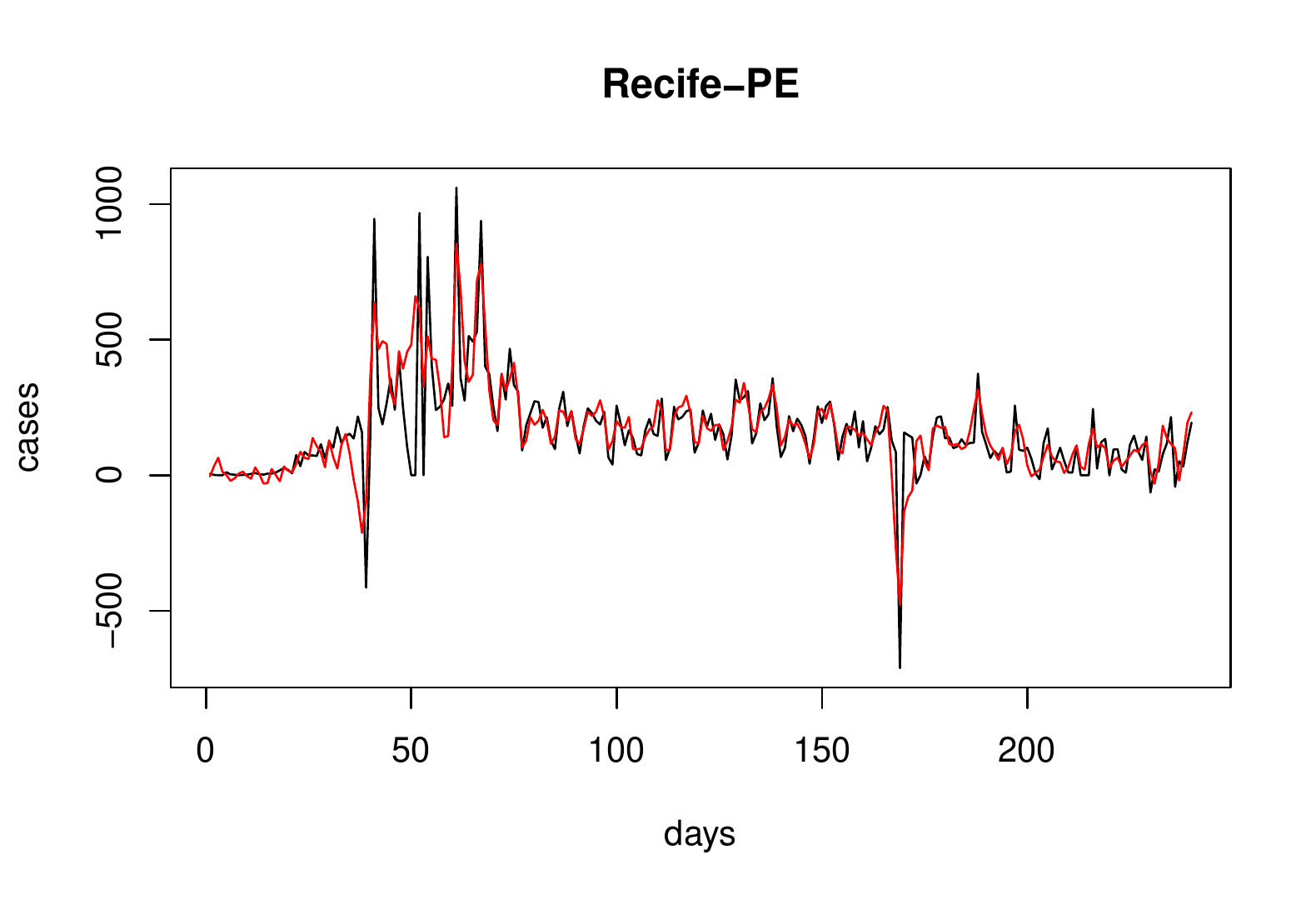}
  }
      \caption{Predicted (red) and observed (black) models to the Northeast region of Brazil. }\label{fig2}
\end{figure} 

\begin{figure} 
  \subfloat[Salvador-BA]{ 
     \includegraphics[trim = 0mm 0mm 0mm 14mm,clip,scale=0.45]{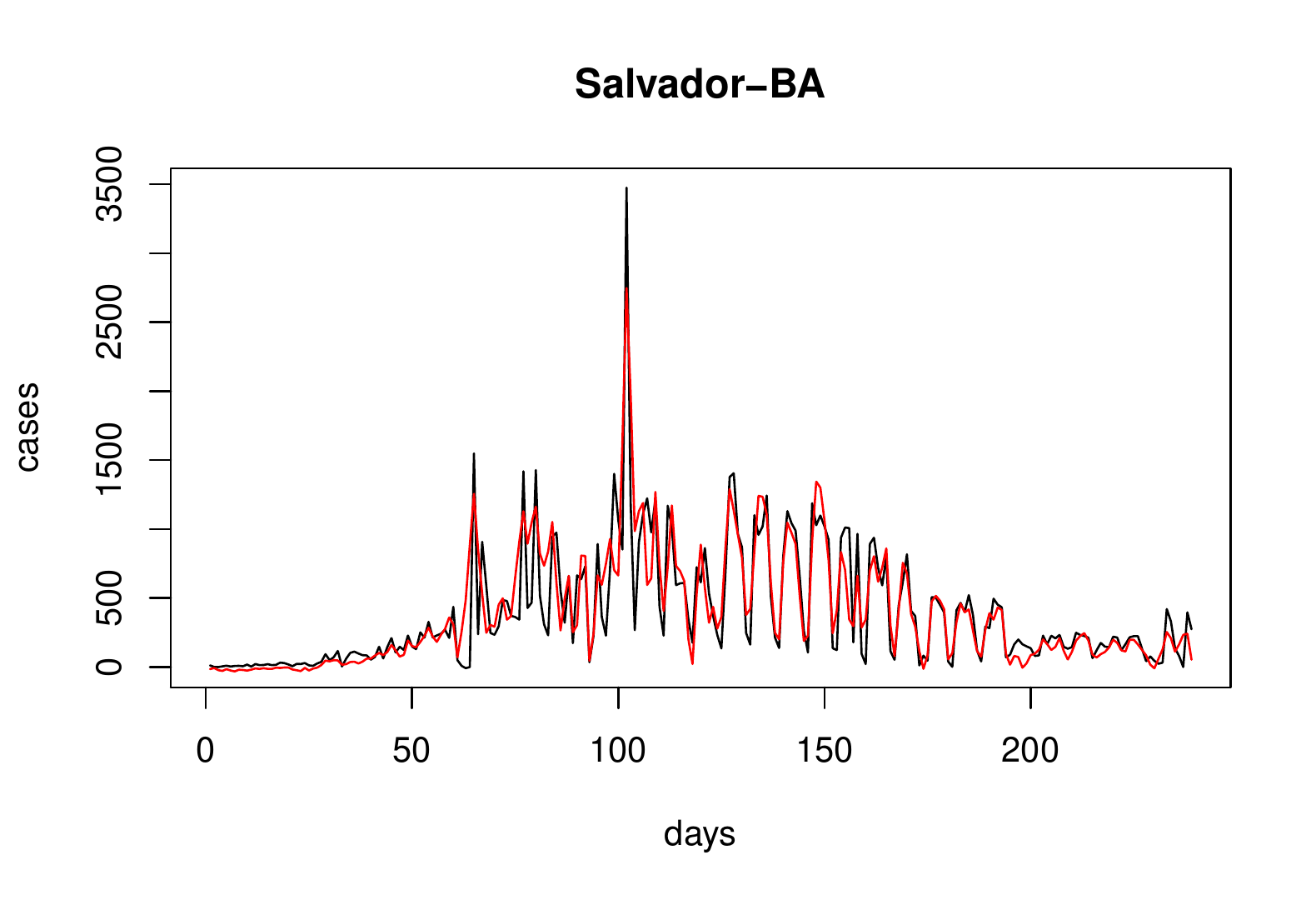}
  } 
\subfloat[São Luis-MA]{ 
     \includegraphics[trim = 0mm 0mm 0mm 14mm,clip,scale=0.45]{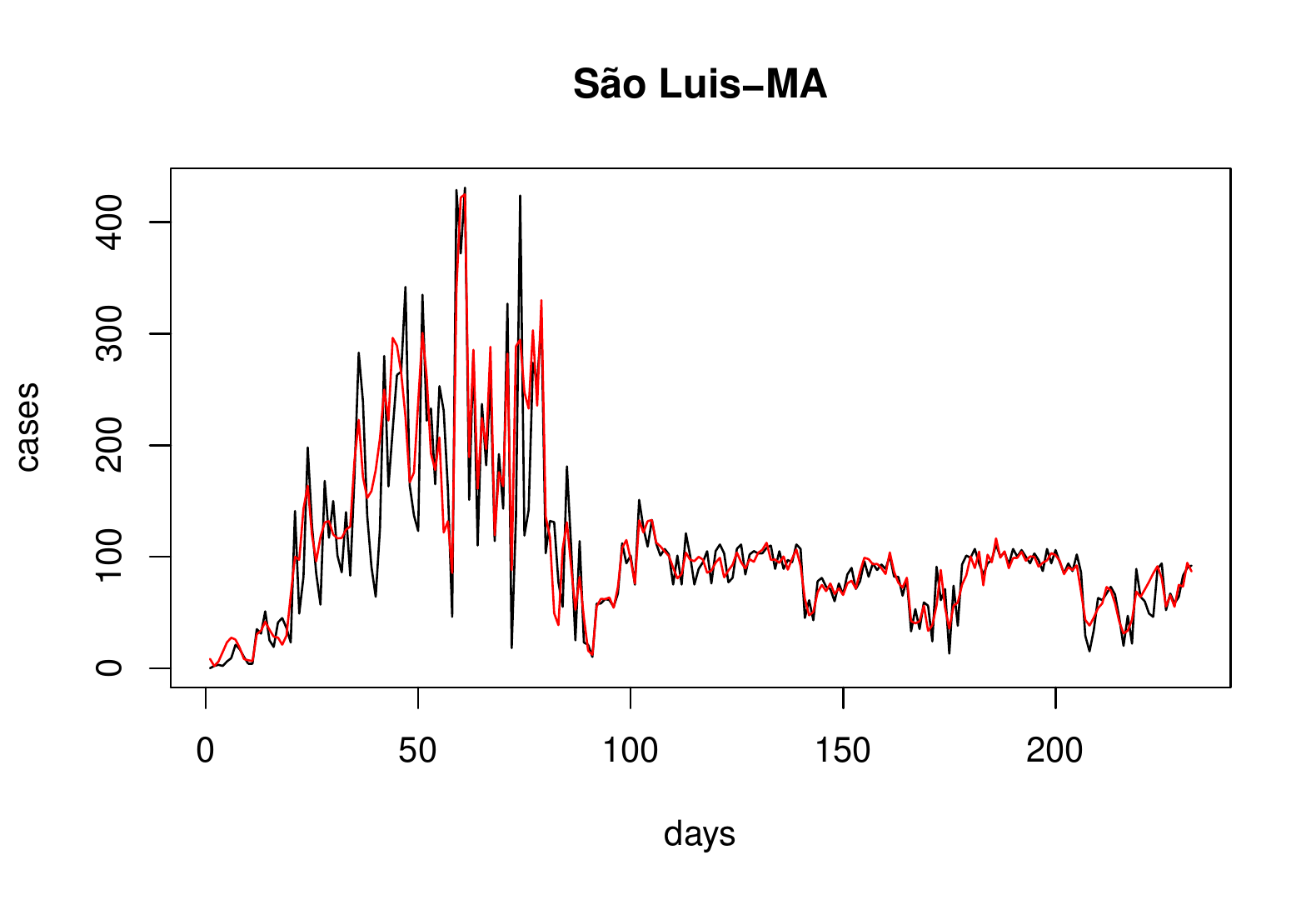}
  } 
  \\ 
  \subfloat[Teresina-PI]{ 
     \includegraphics[trim = 0mm 0mm 0mm 14mm,clip,scale=0.45]{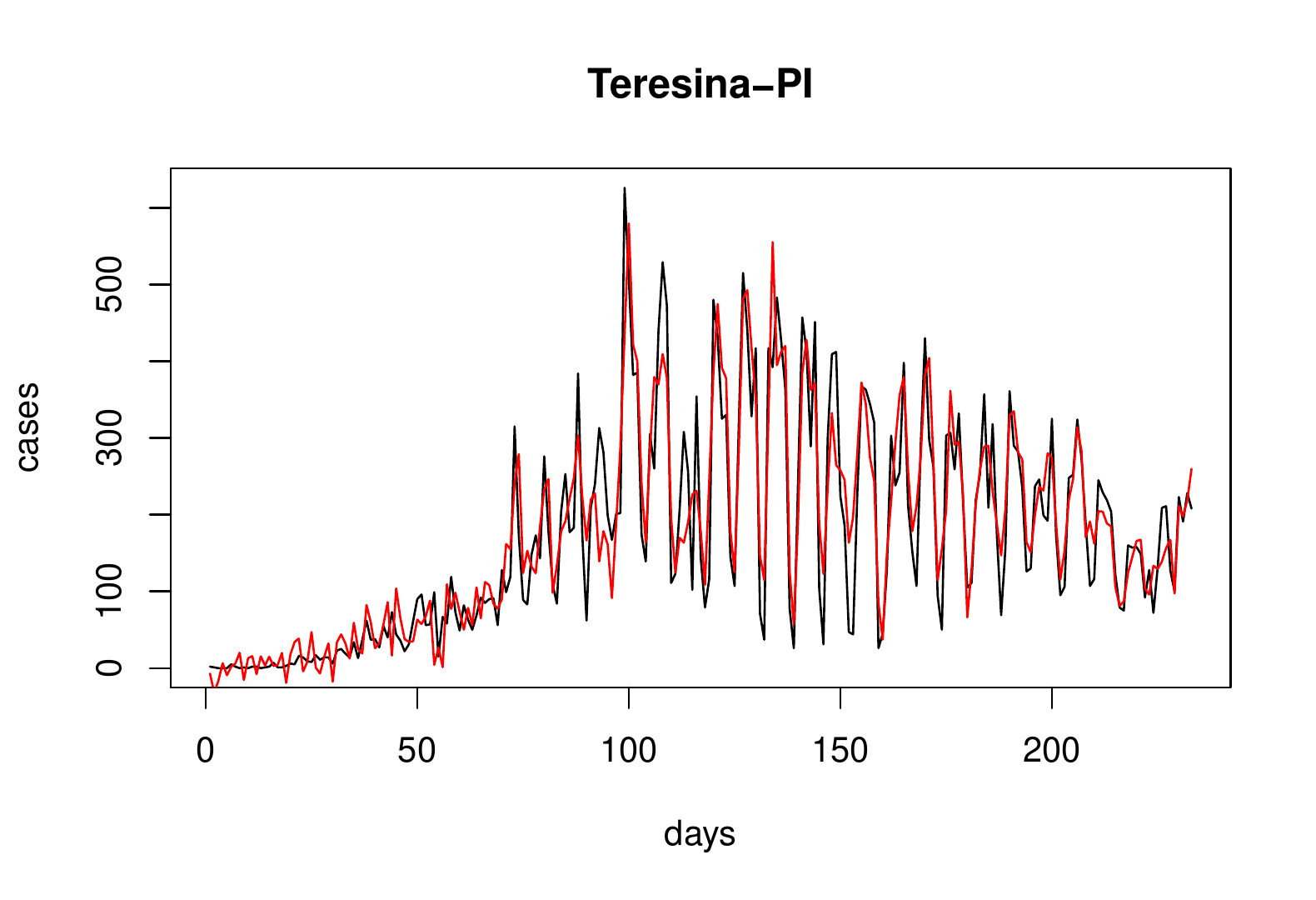}
  } 
  \caption{Predicted (red) and observed (black) models to the Northeast region of Brazil (continuation). }\label{fig2b}
\end{figure}

\begin{figure} 
  \subfloat[Brasília-DF]{ 
    \includegraphics[trim = 0mm 0mm 0mm 14mm,clip,scale=0.45]{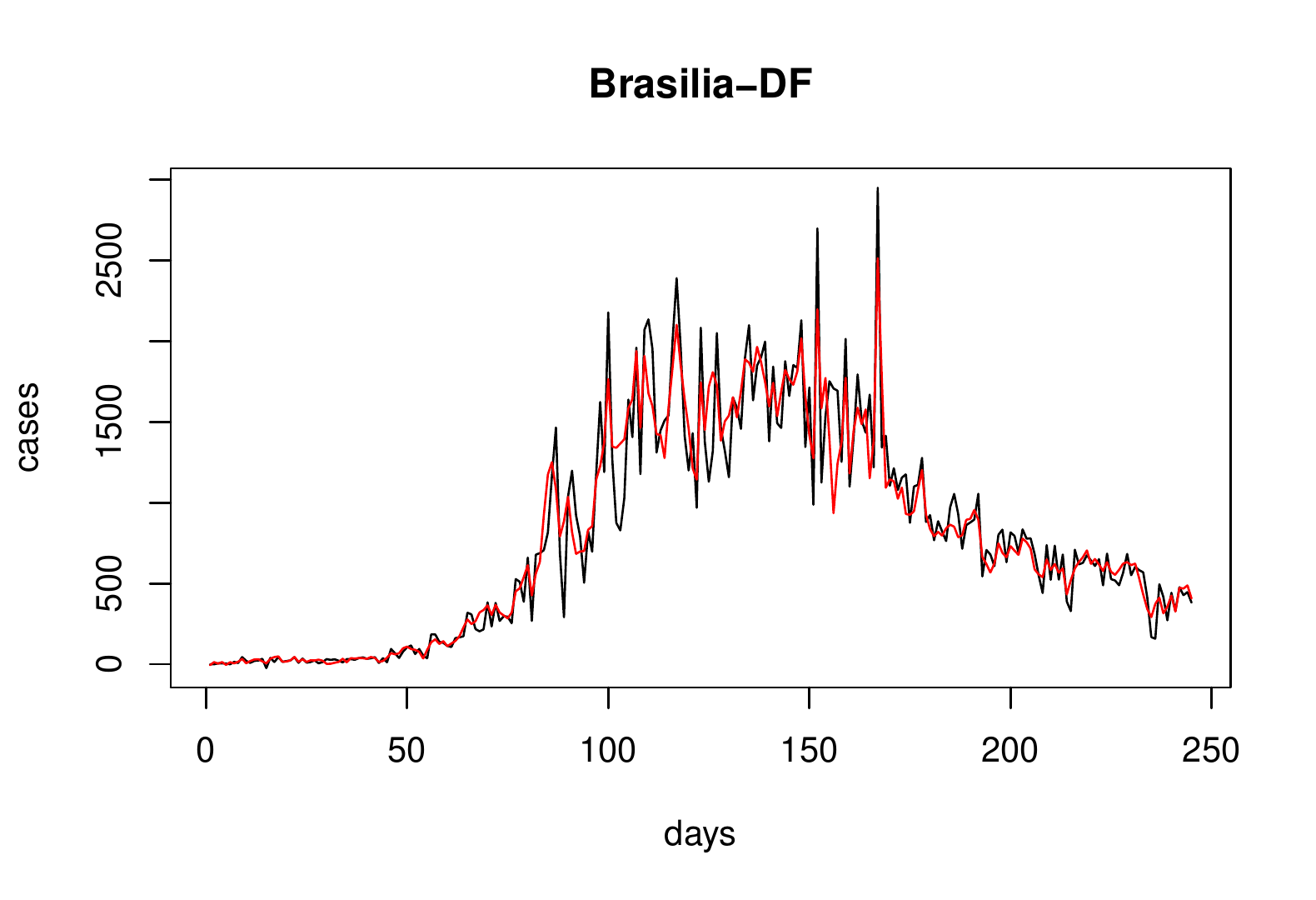}
  } 
  \subfloat[Campo Grande-MS]{ 
    \includegraphics[trim = 0mm 0mm 0mm 14mm,clip,scale=0.45]{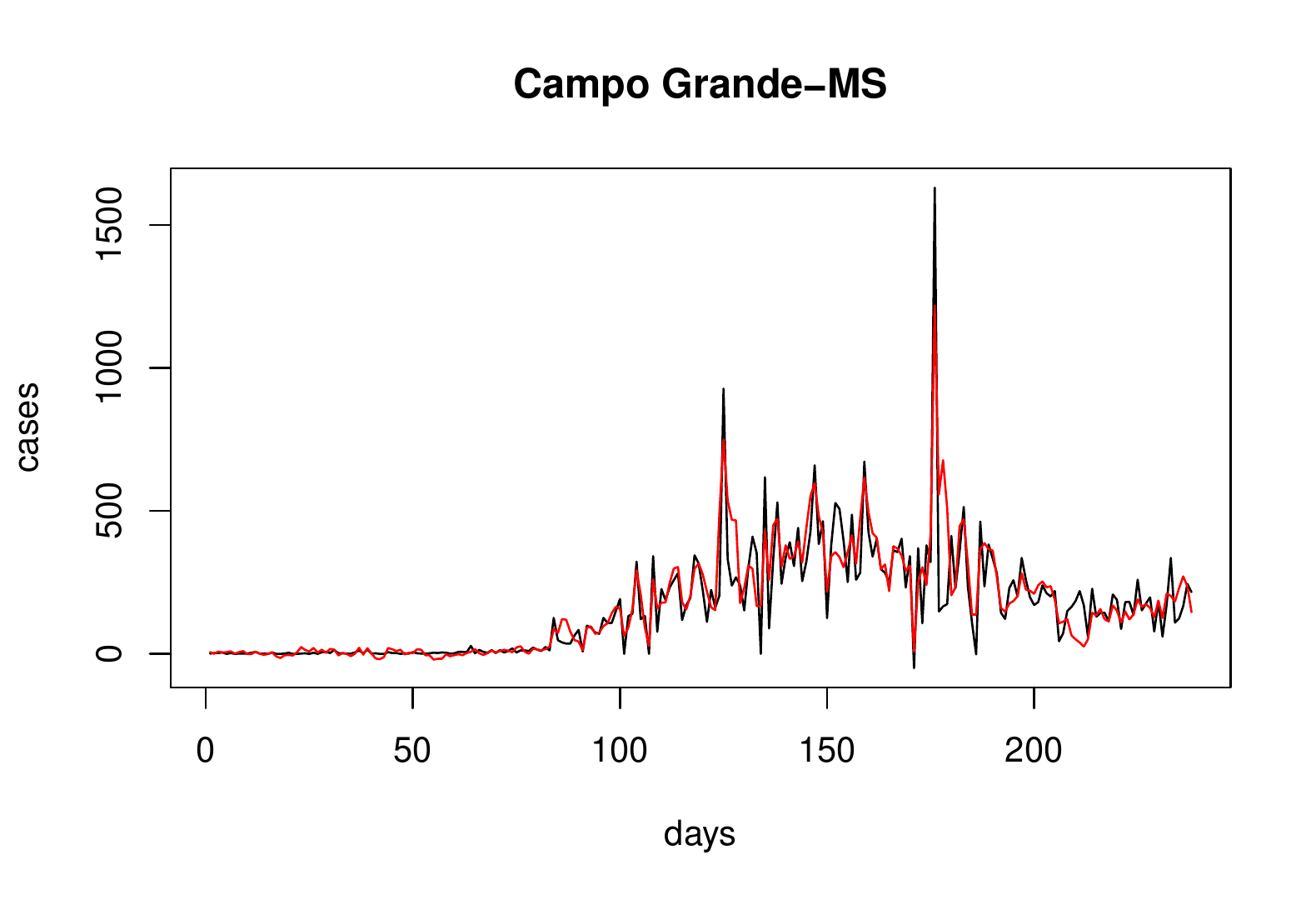}
  } 
  \\ 
  \subfloat[Cuiabá-MT]{ 
     \includegraphics[trim = 0mm 0mm 0mm 14mm,clip,scale=0.45]{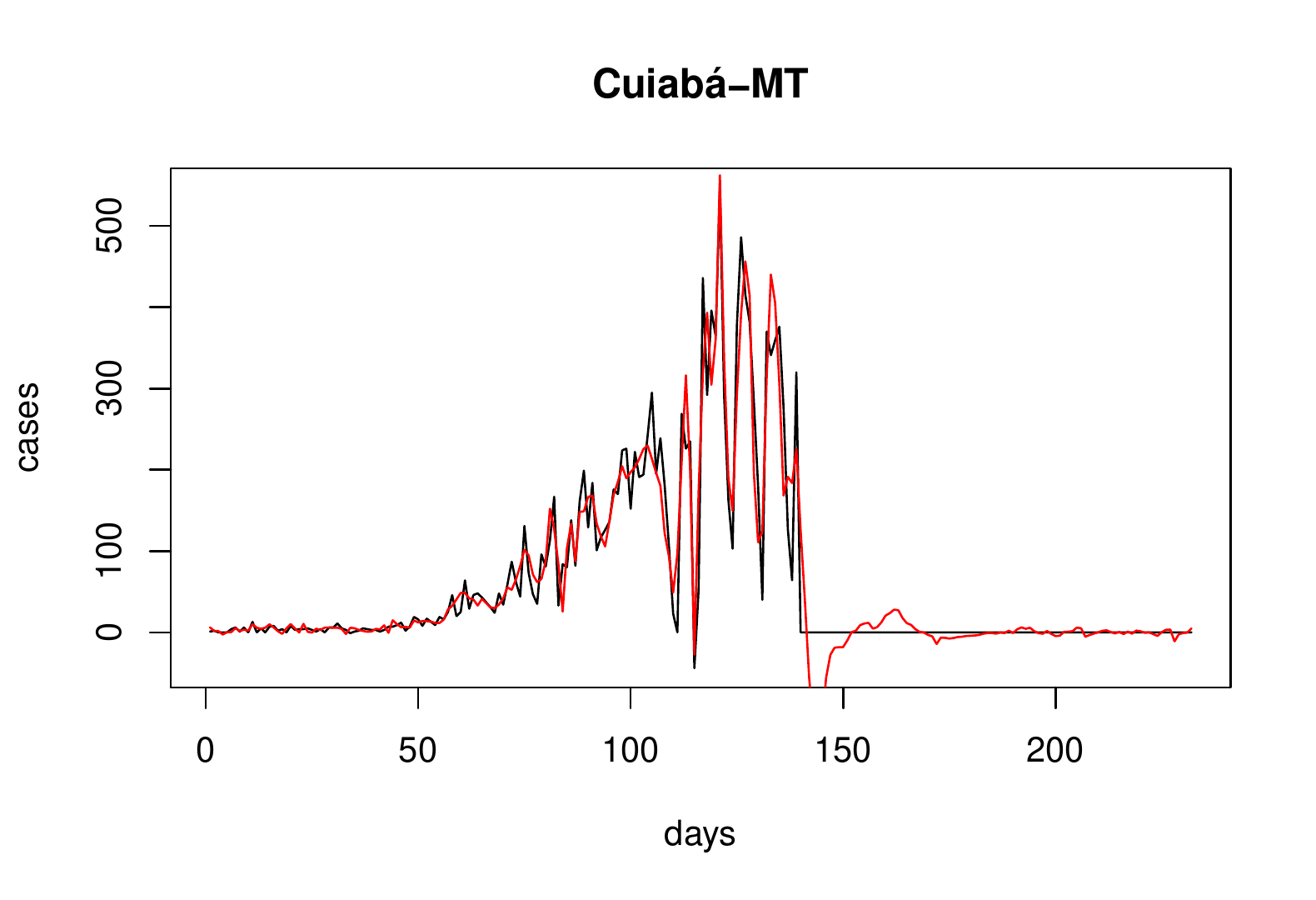}
  } 
    \subfloat[Goiânia-GO]{ 
    \includegraphics[trim = 0mm 0mm 0mm 14mm,clip,scale=0.45]{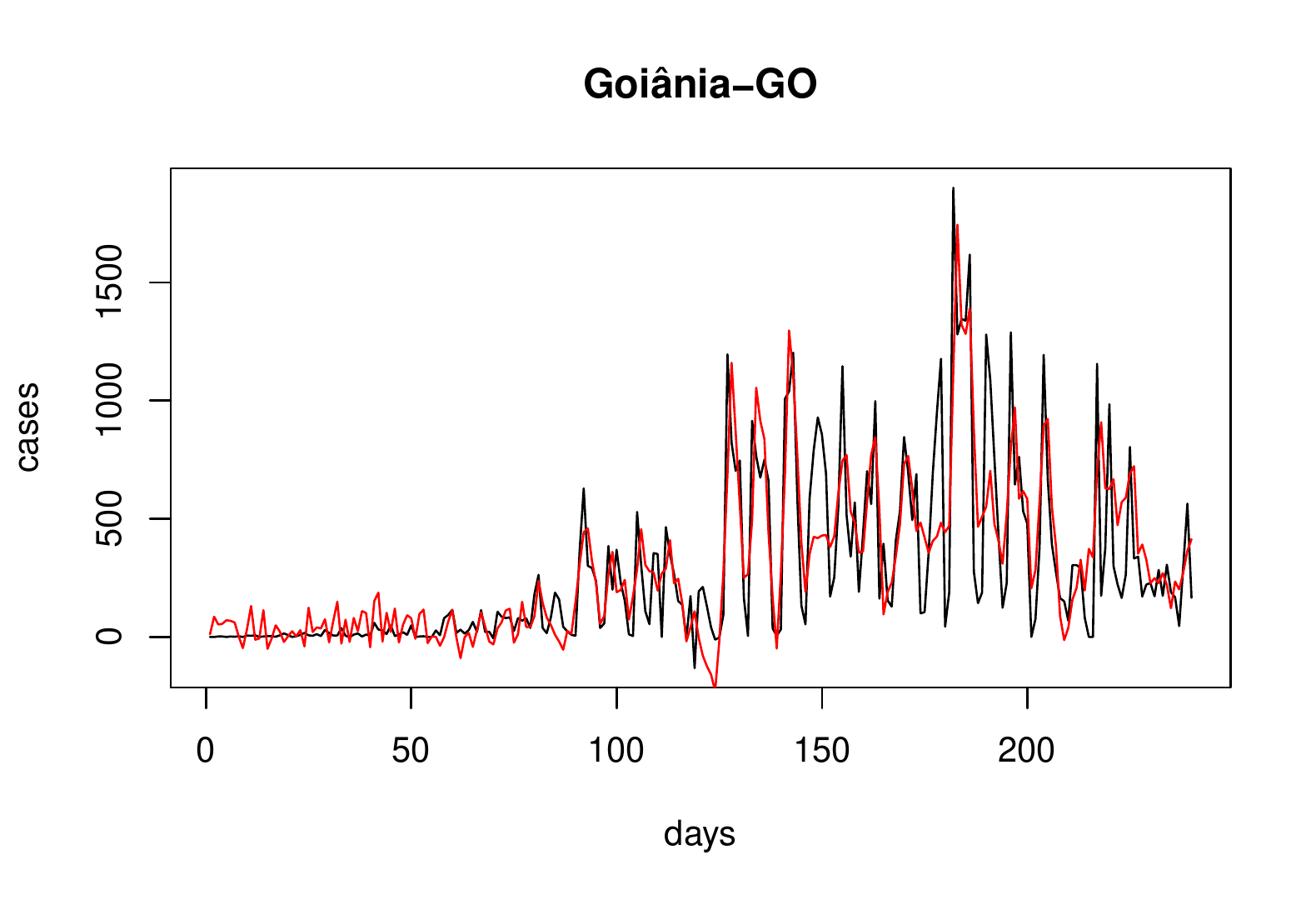}
  }
   \caption{Predicted (red) and observed (black) models to the Midwest region of Brazil. } \label{fig3}
\end{figure} 

\begin{figure} 
  \subfloat[Belo Horizonte-MG]{ 
    \includegraphics[trim = 0mm 0mm 0mm 14mm,clip,scale=0.45]{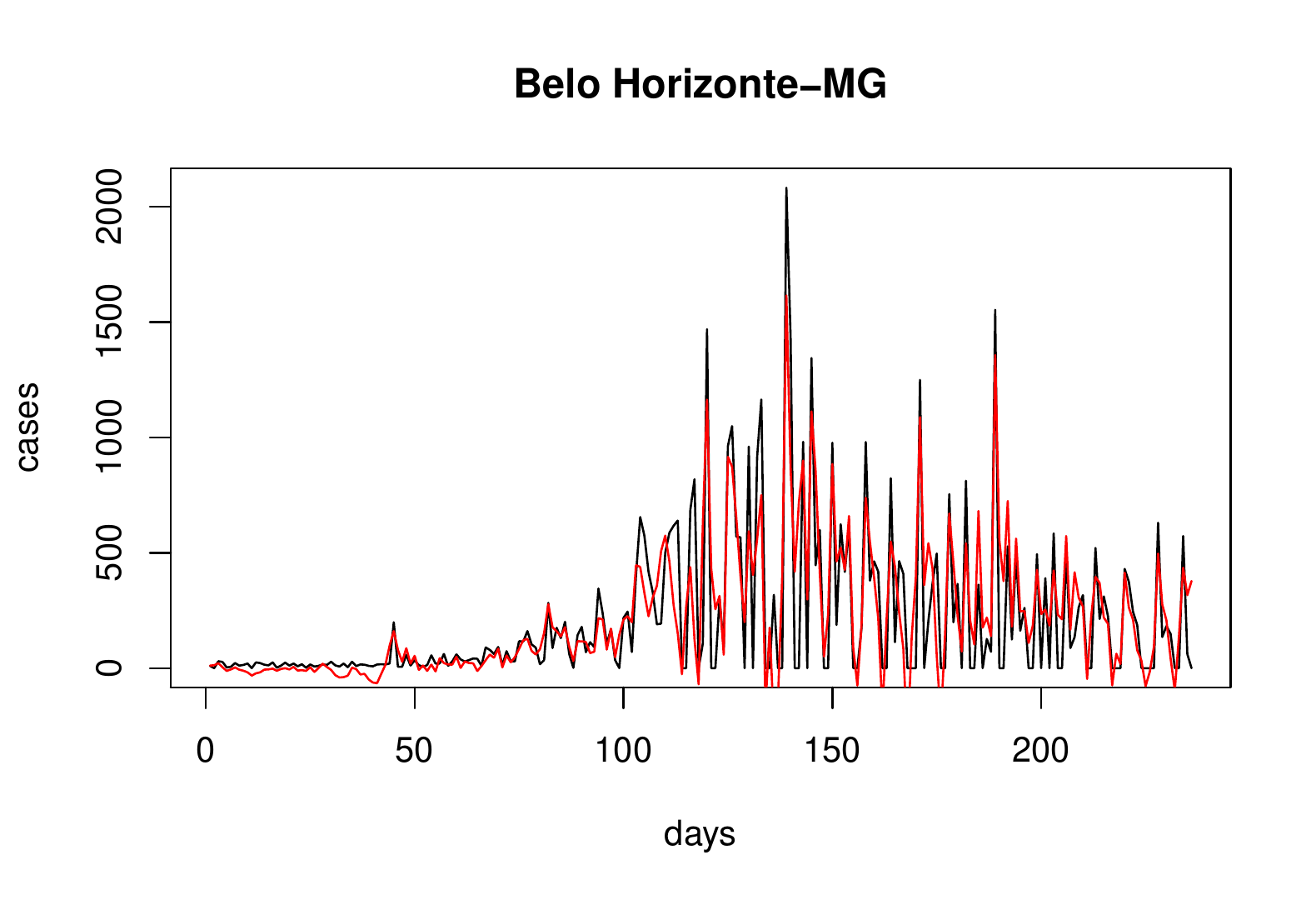}
  } 
  \subfloat[Rio de Janeiro-RJ]{ 
    \includegraphics[trim = 0mm 0mm 0mm 14mm,clip,scale=0.45]{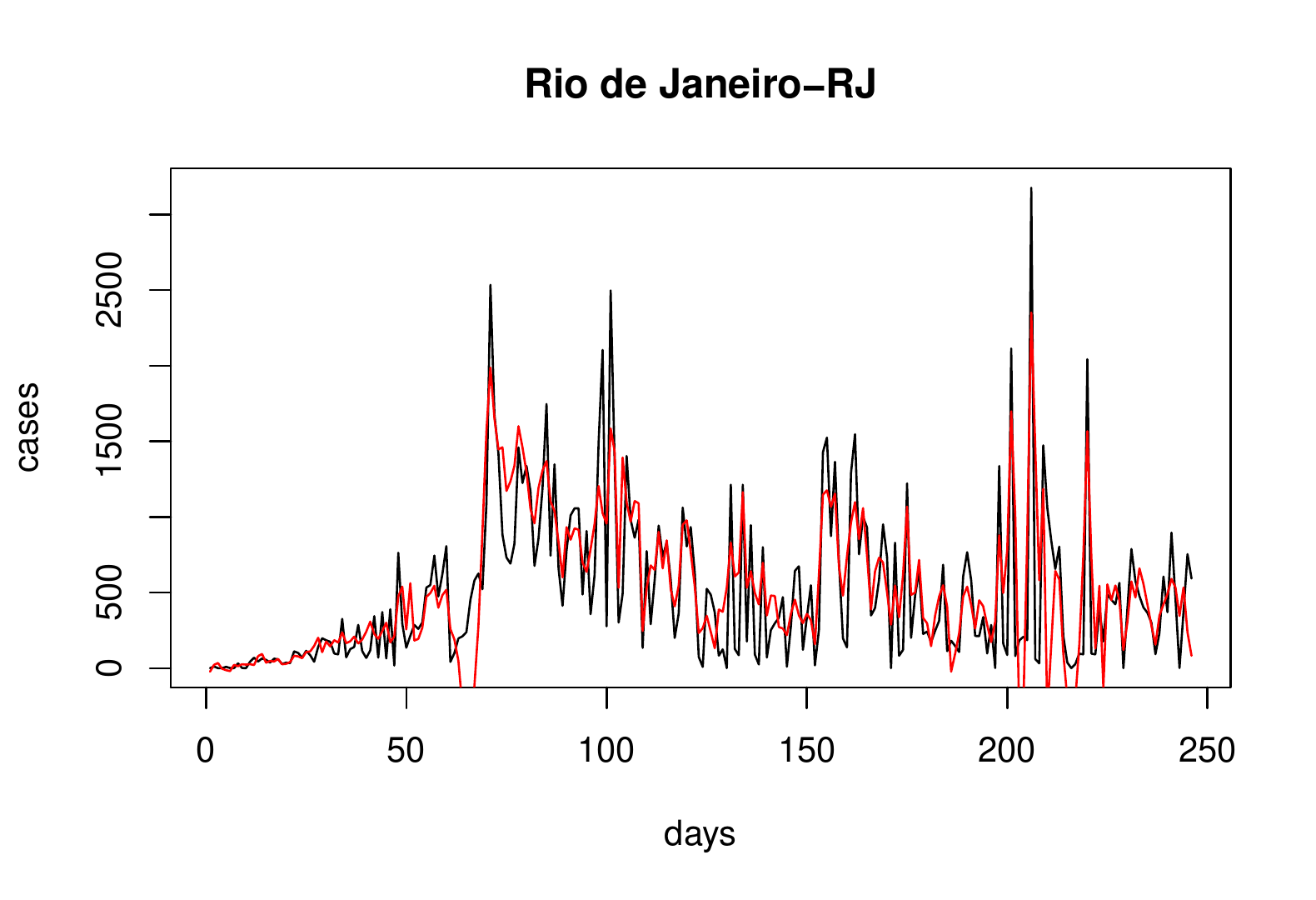}
  } 
  \\ 
  \subfloat[São Paulo-SP]{ 
     \includegraphics[trim = 0mm 0mm 0mm 14mm,clip,scale=0.45]{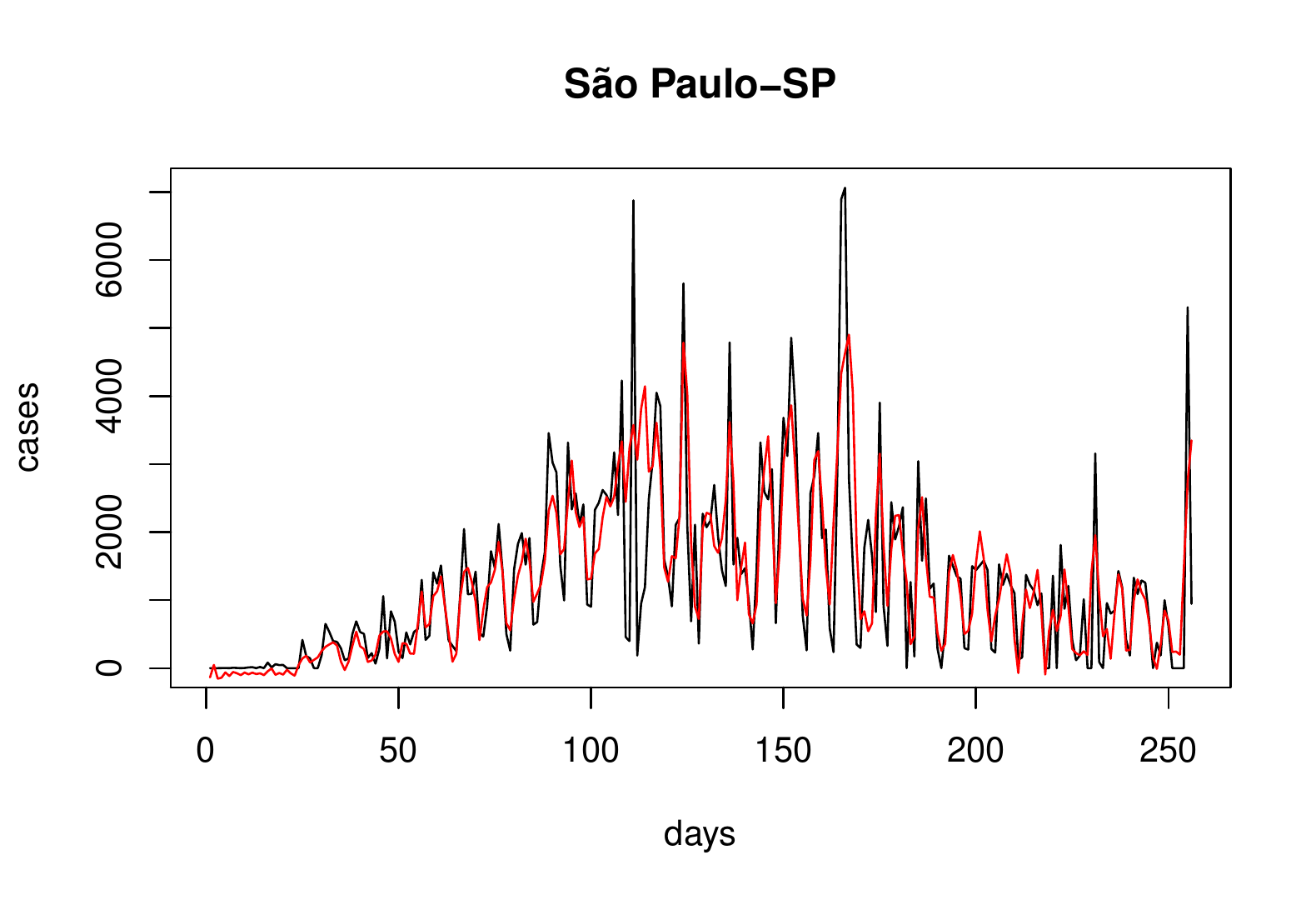}
  } 
    \subfloat[Vitória-ES]{ 
    \includegraphics[trim = 0mm 0mm 0mm 14mm,clip,scale=0.45]{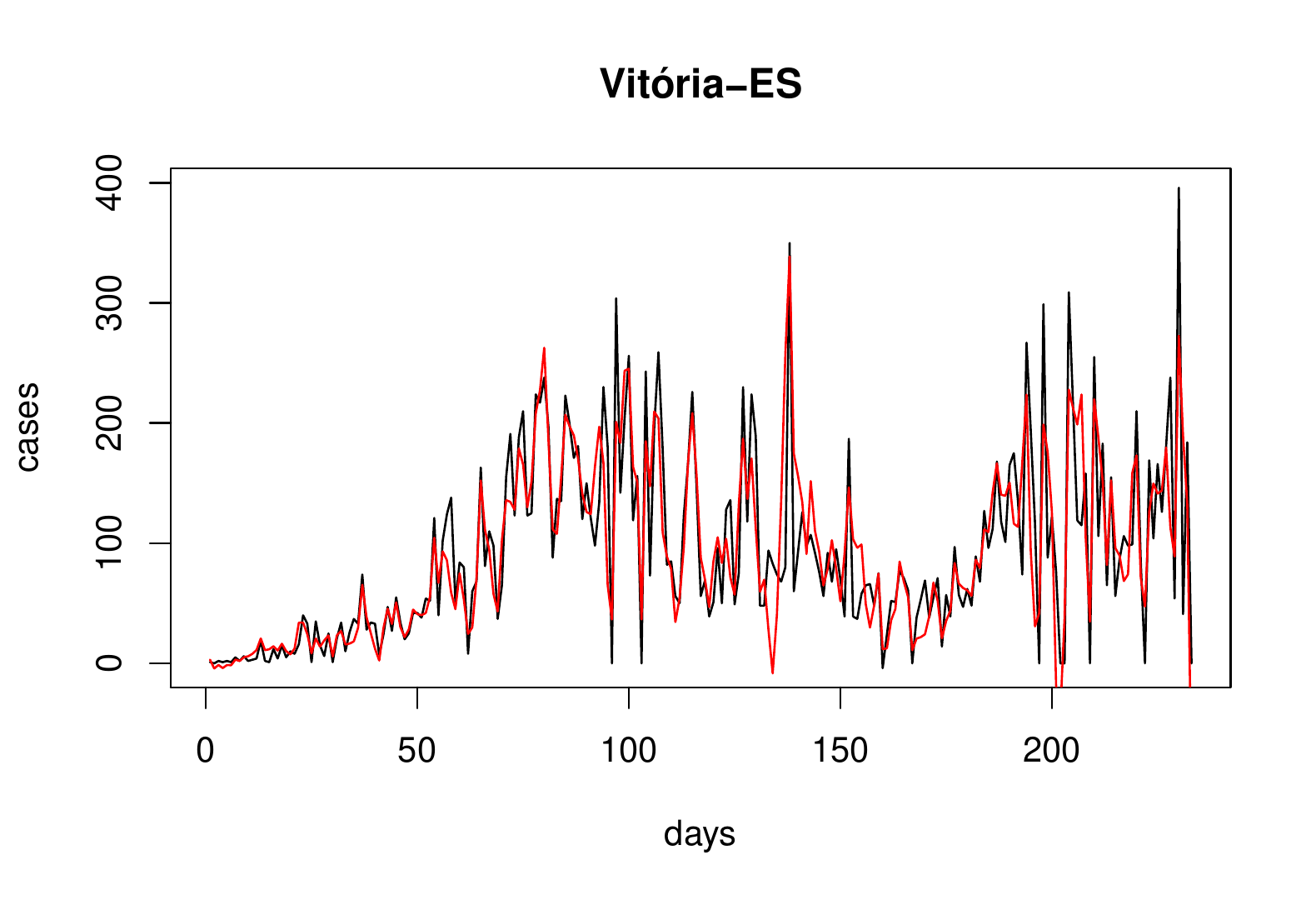}
  }
   \caption{Predicted (red) and observed (black) models to the Southeast region of Brazil. } \label{fig4}
\end{figure}

\begin{figure} 
  \subfloat[Curitiba-PR]{ 
    \includegraphics[trim = 0mm 0mm 0mm 14mm,clip,scale=0.45]{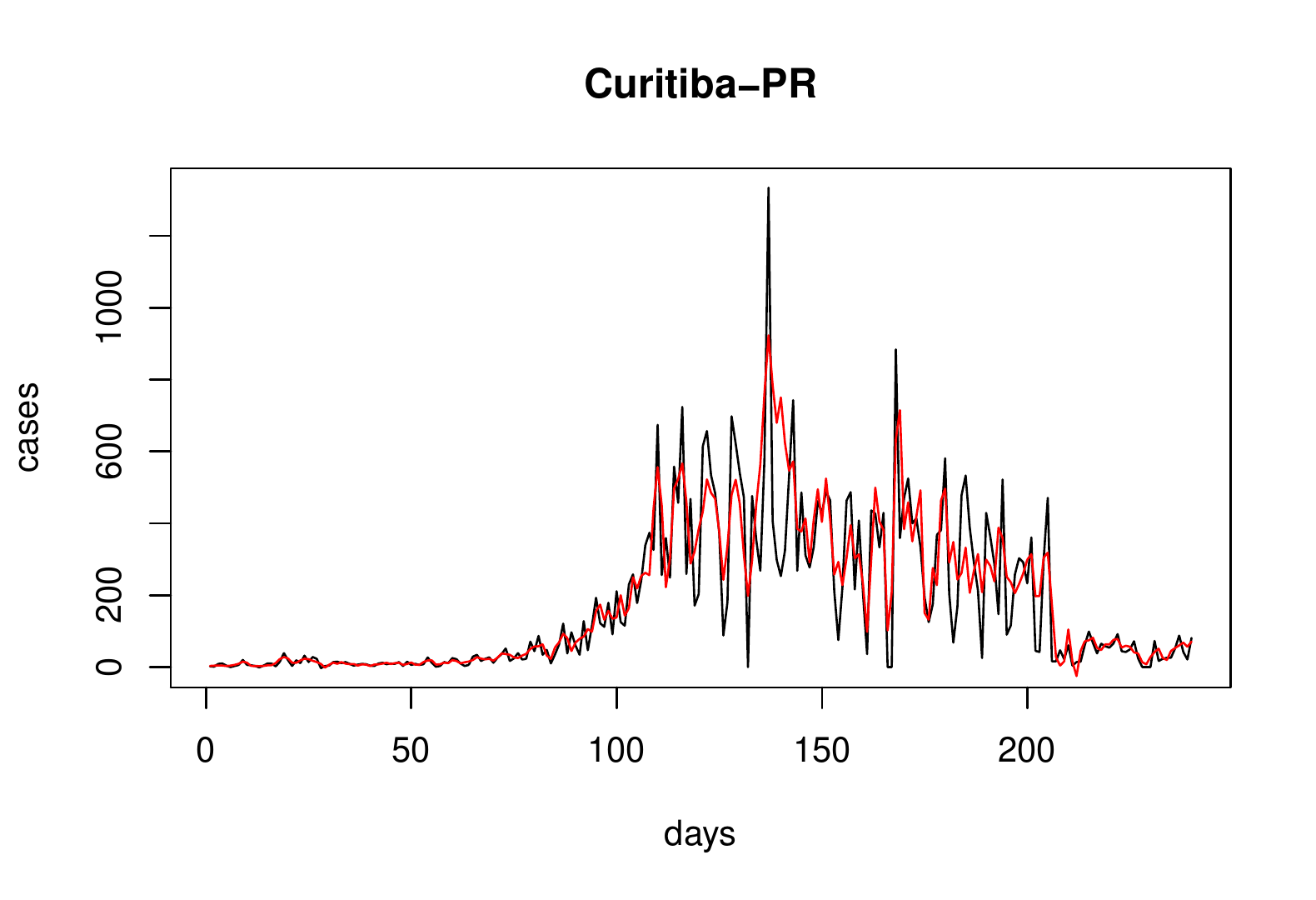}
  } 
  \subfloat[Florianópolis-SC]{ 
    \includegraphics[trim = 0mm 0mm 0mm 14mm,clip,scale=0.45]{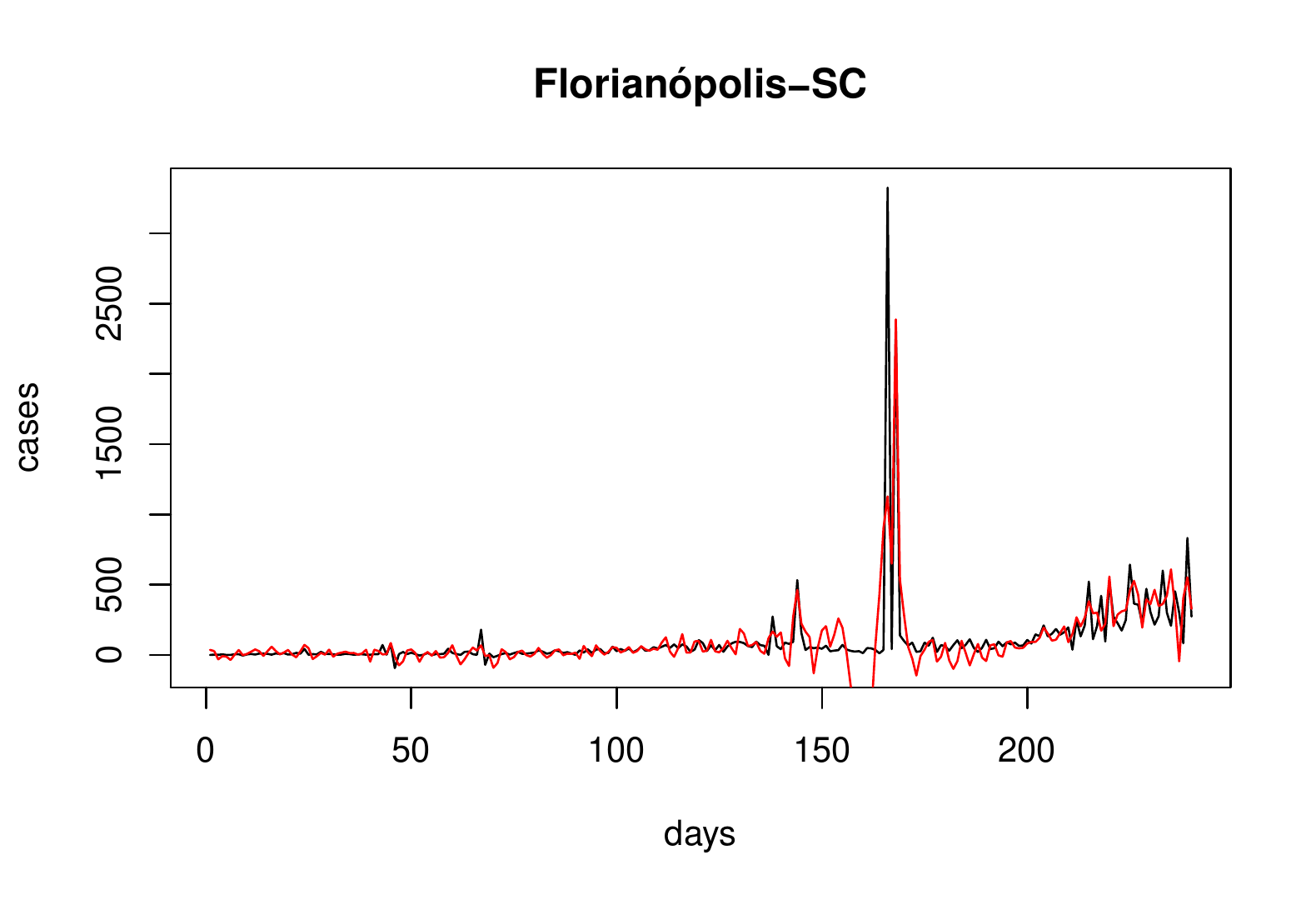}
  } 
  \\ 
  \subfloat[Porto Alegre-RS]{ 
     \includegraphics[trim = 0mm 0mm 0mm 14mm,clip,scale=0.45]{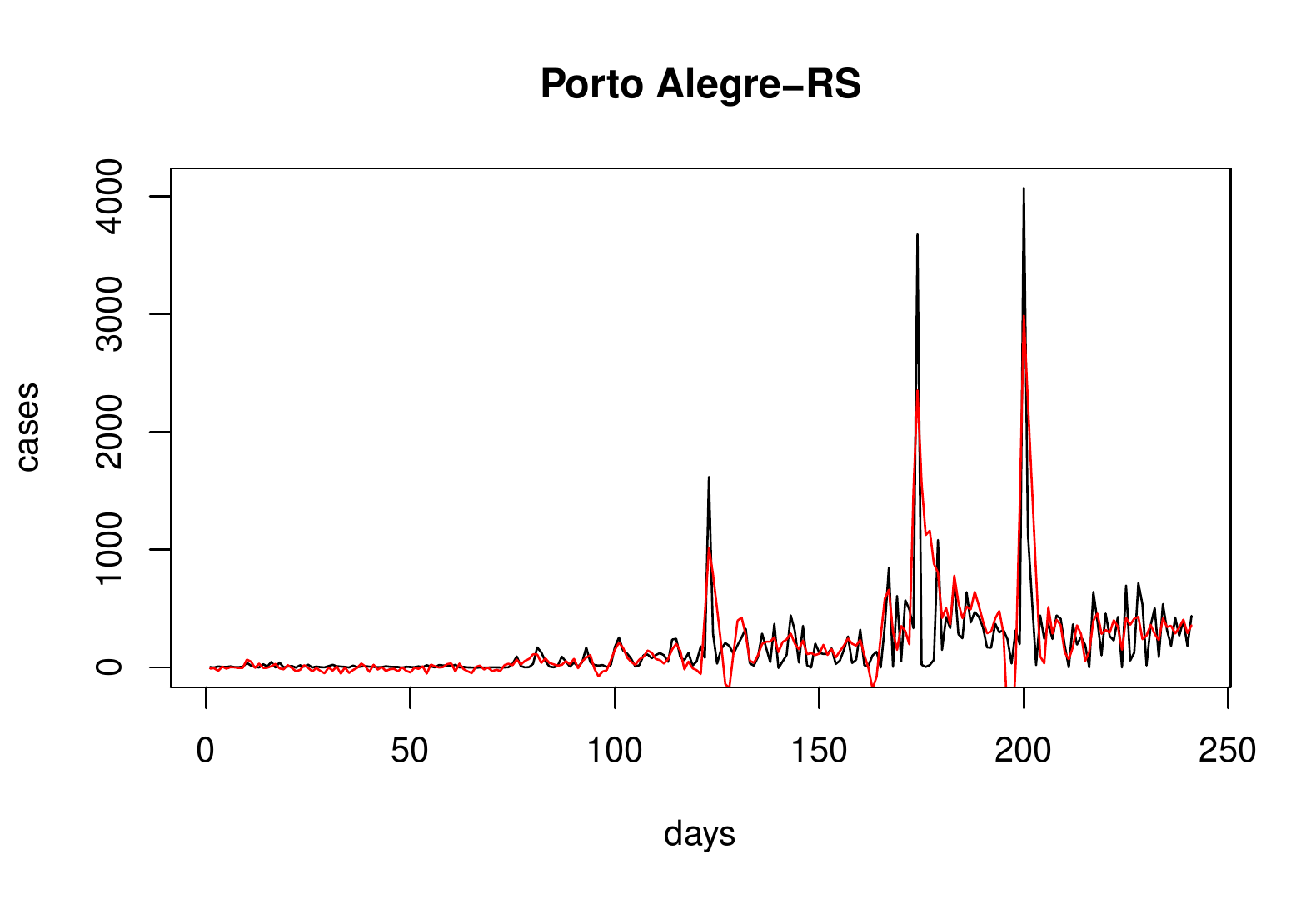}
  } 
   \caption{Predicted (red) and observed (black) models to the South region of Brazil. } \label{fig5}
\end{figure} 

Figures~\ref{fig1} to \ref{fig5} present the original daily number of COVID-19 cases and the daily predicted number of COVID-19 cases by the EEMD-ARIMAX method in each of the Brazilian capitals. The x-axis corresponds to the days of the analyzed period and the y-axis refers to the number of daily COVID-19 cases.

On the one hand, Porto Alegre-RS, São Paulo-SP and Rio de Janeiro-RJ were the cities with the worst prediction results by the EEMD-ARIMAX method, reaching RMSE values of 282.510, 775.817, and 318.717, respectively. On the other, Cuiabá-MT, Palmas-TO, Rio Branco-AC, and São Luis-MA were the cities that had the best prediction results by the EEMD-ARIMAX method, reaching RMSEs of 35.695, 53.240, 30.006, and 32.342, respectively.

\end{document}